\pdfoutput=1
\documentclass[10pt,letterpaper]{article}
\usepackage[letterpaper,textwidth=5.5in,textheight=9in,centering]{geometry}
\usepackage{times}
\usepackage[utf8]{inputenc}
\usepackage[T1]{fontenc}
\usepackage[round]{natbib}
\usepackage[affil-it]{authblk}
\usepackage{url,booktabs,array,graphicx,placeins,needspace,makecell}
\newsavebox{\fitwidthbox}
\newcommand{\fitwidth}[1]{\sbox{\fitwidthbox}{#1}\ifdim\wd\fitwidthbox>\linewidth\resizebox{\linewidth}{!}{\usebox{\fitwidthbox}}\else\usebox{\fitwidthbox}\fi}
\usepackage{enumitem}
\usepackage{etoc}
\usepackage{amsfonts,amsmath,nicefrac,microtype,xcolor}
\usepackage{float}
\floatstyle{ruled}
\newfloat{algorithm}{tbp}{loa}
\floatname{algorithm}{Algorithm}
\usepackage{hyperref}
\hypersetup{colorlinks=true,linkcolor=blue!50!black,citecolor=blue!50!black,urlcolor=blue!50!black,
  pdftitle={PhysioTRACE: Provenance-Aware Stress Tests for Physiological Foundation Models},
  pdfauthor={Ayana Mussabayeva, Anuar Aimoldin, Olivier Oullier, Xue Liu, Kun Zhang}}
\colorlet{revisiongreen}{black}
\colorlet{revisionblue}{black}
\colorlet{revisionred}{black}
\DeclareRobustCommand{\rev}[1]{#1}
\DeclareRobustCommand{\postrev}[1]{{\color{revisiongreen}#1}}

\newcommand{\repourl}{https://github.com/AyanaMussabayeva/physio-trace}

\title{\bfseries PhysioTRACE: Provenance-Aware Stress Tests\\for Physiological Foundation Models}
\author[1,$\ast$]{Ayana Mussabayeva}
\author[1,$\ast$]{Anuar Aimoldin}
\author[1]{Olivier Oullier}
\author[1,2]{Xue Liu}
\author[1,3]{Kun Zhang}
\affil[1]{Mohamed bin Zayed University of Artificial Intelligence (MBZUAI), Abu Dhabi, UAE}
\affil[2]{McGill University, Montreal, Canada}
\affil[3]{Carnegie Mellon University, Pittsburgh, USA}
\affil[$\ast$]{Equal contribution}
\affil[ ]{\vspace{2pt}{\fontsize{7.6}{9}\selectfont\texttt{\{ayana.mussabayeva,\,anuar.aimoldin,\,olivier.oullier,\,steve.liu,\,kun.zhang\}@mbzuai.ac.ae}}}
\date{}
\begin{document}
\maketitle

\begin{abstract}
Physiological foundation models encode how a signal was recorded alongside the physiology it reflects. When recording conditions are associated with diagnosis, this acquisition provenance can become a shortcut, yet the two measurements usually offered as evidence, shifted transfer and provenance decodability, do not show whether a predictor uses it. We introduce PhysioTRACE, a four-axis behavioral audit for frozen encoders that separates what a probe can decode from what a fixed task head relies on. Recover scores how decodable provenance is; Stress reverses only the provenance--target association on the same held-out records; Intervene removes a train-localized provenance component; and Verify certifies that removal only if it beats matched random projections within a declared utility margin. Each audit thus ends in one of three verdicts: no reliance, or reliance with the remedy certified or refused. Across electroencephalography (EEG) and electrocardiography (ECG), five training objectives, and five frozen foundation models, the relation between Recover's calibrated score (Linear Provenance Information) and out-of-distribution utility changes sign between datasets, so neither predicts the other and neither can stand in for a reliance test. On paired EEG views where the shortcut is known by construction, the audit detects it (the head fitted under the association loses about 0.2 AUROC when the association is reversed, while a control head fitted without it is unaffected), localizes a rank-two component whose removal returns that head to control-level behavior without measurable loss of ordinary utility, and certifies the remedy for both encoder objectives tested. On real ECG device metadata it returns all three verdicts: it certifies a remedy that removes 91\% of one model's excess vulnerability, finds no reliance where device and diagnosis are barely associated, and refuses the remedy for a second model whose localized direction also carries task signal. A claim that a physiological model is robust to how its inputs were recorded therefore needs a behavioral test, and PhysioTRACE turns it into a test that can pass, fail, or refuse a remedy. We release the protocol, splits, probes, and machine-readable results.
\end{abstract}

\etocdepthtag.toc{main}
\section{Introduction}
\label{sec:intro}
Physiological foundation models (PFMs) promise reusable encoders for electroencephalography (EEG), electrocardiography (ECG), and wearable signals, trained with contrastive, masked-reconstruction, token-prediction, and hybrid objectives \citep{Kostas2021,Yang2023,Wang2024,Jiang2024,Li2025ECGFounder,Liu2024MERL}. Their appeal is that a frozen encoder captures physiology and a light task head can be deployed wherever that physiology is recorded. This holds only if the head reads physiology from the embedding rather than the circumstances of recording, and current evaluation does not check it.

\begin{figure}[t]
\centering
\includegraphics[width=0.7\linewidth]{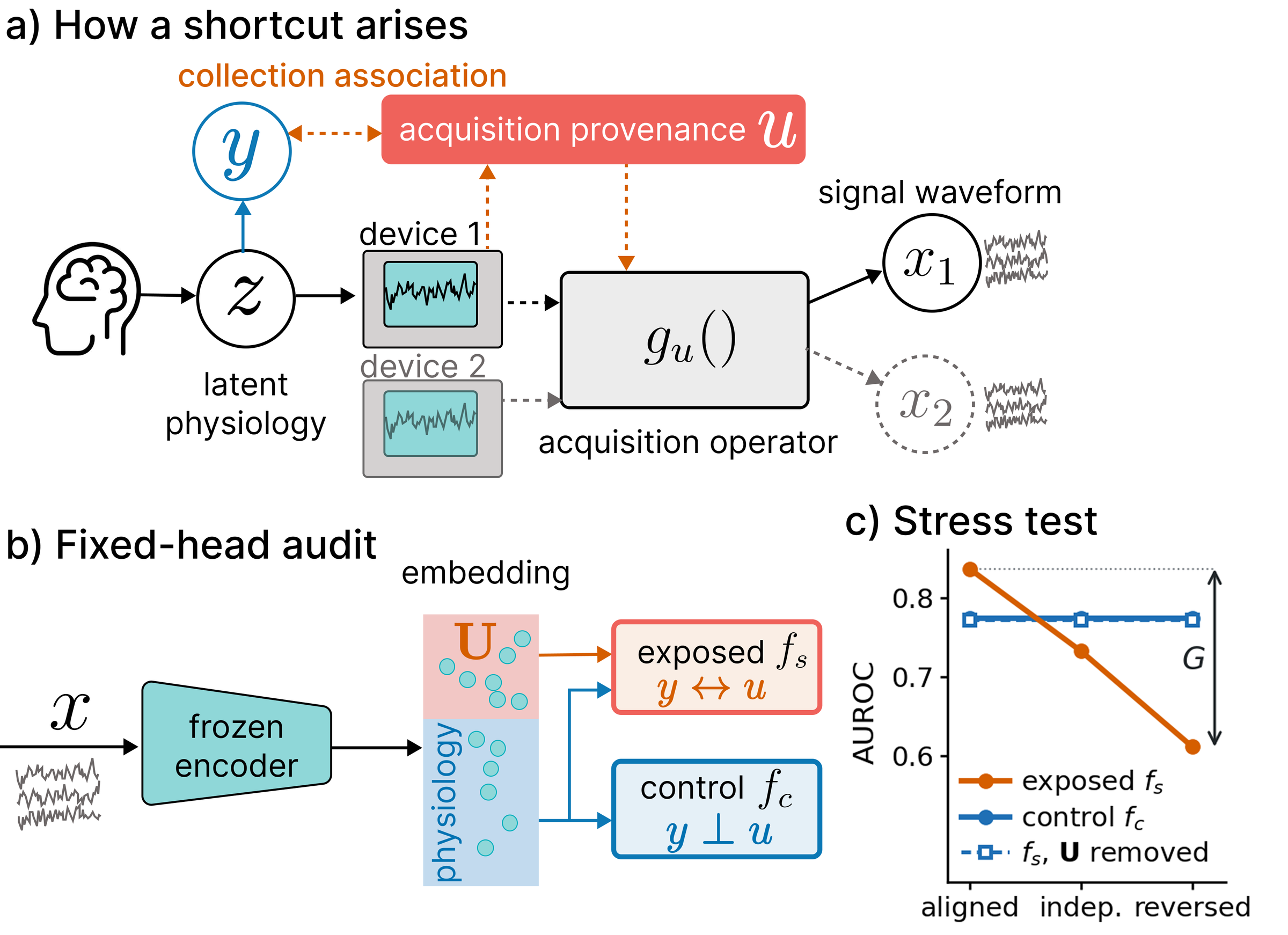}
\caption{Acquisition provenance as a shortcut. \textbf{(a)} Provenance $u$, such as device or reference, shapes the recorded waveform $x$ of latent physiology $z$ and is associated with the target $y$ only through data collection. \textbf{(b)} A frozen encoder retains a low-rank provenance component $\mathbf U$; the exposed head $f_s$ is fitted where $y$ and $u$ are associated, the control head $f_c$ with $y\perp u$. \textbf{(c)} NMT stress test (Appendix Table~\ref{tab:nmt-regime-recon}): reversing only the $y$--$u$ association lowers $f_s$ by the excess vulnerability $G$, while $f_c$, and $f_s$ with $\mathbf U$ removed, are unaffected.}
\label{fig:acquisition-leakage}
\end{figure}

This premise is fragile because recording leaves its own trace in the signal. Acquisition changes the waveform without changing the physiology it reflects: EEG reference and montage alone alter amplitudes, topographies, and spectral estimates \citep{Nunez2006,Kayser2010,Yao2001,Dong2019}, and an encoder retains such provenance because it helps reconstruct the input or separate training sources (Figure~\ref{fig:acquisition-leakage}(a)). Provenance is not biological evidence about the target, yet in a collected dataset it is rarely independent of it, because sites and datasets differ in whom they recruit \citep{Castro2020}: site, device, and preprocessing co-vary with diagnosis, age, and case mix. A task head can then substitute collection-specific signal for physiology and fail when collection practices change, even if aggregate performance looked strong \citep{Geirhos2020}. Hospital cues create the same kind of shortcut in clinical imaging \citep{Zech2018,Roberts2021}.

In practice, the risk of such shortcuts is assessed with two measurements, shifted transfer and provenance probing, but neither shows whether the head actually relies on provenance. Benchmarks span tasks, subjects, and datasets \citep{Jayaram2018,Wagner2020,Strodthoff2021,Kuruppu2025} but fold reference, montage, device, site, and preprocessing into domain labels, so provenance enters evaluation only through these two numbers. Shifted transfer asks whether utility survives a change of acquisition; a provenance probe asks whether acquisition can be decoded from the embedding. Transfer can fail for reasons unrelated to provenance, and a head can ignore information that a probe recovers easily, so accessibility is not reliance. Reliance is a property of the head's decisions, and it can only be established by changing what the head could rely on and watching what it does.

We therefore propose PhysioTRACE (Physiological Transfer Robustness and Acquisition Consistency Evaluation), a behavioral audit for frozen encoders. It changes only the provenance--target association on the same held-out records, with the task heads fixed, and compares an exposed head fitted under the association with a control head fitted without it. How much more the exposed head degrades is its excess vulnerability. It then asks whether removing a train-localized provenance component, a candidate remedy, reverses that failure without costing ordinary utility. That last condition is not a formality. Since provenance co-varies with the participants whose signals were recorded, a remedy that removes whatever separates sources would, by default, also remove target-relevant population signal. PhysioTRACE therefore localizes the component conditionally on the target and certifies its removal only if it beats matched random controls and stays within a declared utility margin. Section~\ref{sec:protocol} formalizes the audit in four steps: Recover, Stress, Intervene, and Verify.

We begin by comparing shifted transfer and provenance probing directly, and find that they do not tell a consistent story. We measure shifted transfer as out-of-distribution (OOD) utility. To rule out a weak probe, we score provenance probing with Linear Provenance Information (LPI), a calibrated, identity-aware estimator that we introduce as the Recover step of the audit. Within each benchmark, we rank the five shared-backbone objectives and that modality's frozen PFMs by both. On ERP/P300 the relation is negative: the encoders that expose the most provenance also transfer worst, as if accessibility could predict transfer failure. On PTB-XL it reverses: the encoder with the least provenance information transfers worst, and the one with the most transfers well. HMC sleep staging again decouples the two. Since whether one measurement predicts the other depends on the benchmark, neither can validate the other, and neither tells whether a task head actually relies on provenance; that question needs a behavioral test (Section~\ref{sec:broad-screen}).

To validate the audit itself, we next apply it where the shortcut is known by construction. In paired NMT EEG, a linear head fitted while reference co-occurs with diagnosis loses $0.20$--$0.22$ AUROC when only that association is reversed on the same records and encoder. A control head fitted without the association is unaffected, for both encoder objectives. Removing a rank-two subspace, localized from training records without diagnosis labels, restores control-level behavior and leaves in-domain and cross-reference AUROC unchanged. A model can therefore pass in-domain and shifted-transfer evaluations and still carry a deployable shortcut (Section~\ref{sec:nmt-audit}).

Finally, we test the audit where provenance and population are entangled, as in most collected data. On PTB-XL, device also indexes recording era, workflow, and case mix, and there are neither paired views nor a designed association; the same audit returns each of its three verdicts. It certifies a rank-one remedy that removes 91\% of ECGFounder's excess vulnerability, finds no reliance on a device pair with almost no natural association, and refuses to certify the remedy for CLEF-Small, whose localized direction also carries task signal, the case in which removing provenance by default would have cost $0.039$ AUROC (Section~\ref{sec:ecg-audit}).

Our main contributions are:
\begin{enumerate}[leftmargin=*,nosep,topsep=2pt,itemsep=2pt]
\item \textbf{A behavioral audit of provenance reliance.} PhysioTRACE tests whether a fixed task head on a frozen encoder uses acquisition provenance and certifies a train-localized remedy only if it beats matched random controls within a declared utility margin (Section~\ref{sec:protocol}). We release it as a reusable code package (\url{\repourl}) with leakage-aware splits, calibrated probes, stress weights, localization operators, and machine-readable results.
\item \textbf{A calibrated accessibility measure, and evidence that it is not reliance.} Linear Provenance Information estimates linear usable information \citep{Xu2020Usable} about provenance per native recording unit with identity-aware uncertainty; resampling event-related potential (ERP) trials instead of subjects would overstate its precision 14--31-fold. The relation between LPI and out-of-distribution utility changes sign between datasets, so neither can stand in for the other or for a behavioral test (Sections~\ref{sec:lpi} and~\ref{sec:broad-screen}).
\item \textbf{An audit whose verdict can go either way.} A five-seed paired EEG audit with known ground truth and patient-disjoint observational audits of two frozen ECG PFMs return all three verdicts: three certified remedies, one refused remedy, and one audit without reliance (Sections~\ref{sec:nmt-audit} and~\ref{sec:ecg-audit}).
\end{enumerate}

\section{Related Work and Positioning}
\textbf{Probing and usable information.} A decodable property does not establish behavioral use: probe accuracy conflates information with probe capacity \citep{Hewitt2019Control}, which has motivated information-theoretic probe scores such as minimum description length \citep{Voita2020MDL} and predictive $\mathcal V$-information, the held-out log-score gain of a predictive family over a constant predictor \citep{Xu2020Usable}. LPI is a calibrated linear instance of the latter for acquisition provenance, estimated per native recording unit with identity-aware uncertainty. Like any probe score, it measures accessibility, not use; PhysioTRACE therefore treats it as one axis of four, never as a verdict.

\textbf{Spurious correlations and stress tests.} Group-robust training and evaluation assume known groups and report worst-group or shifted accuracy \citep{Sagawa2020GroupDRO,Koh2021}; nuisance-randomized reweighting \citep{Puli2022NuRD}, counterfactual-invariance stress tests \citep{Veitch2021Counterfactual}, and auxiliary-label regularization \citep{Makar2022Shortcut} target robustness to a known nuisance, and last-layer retraining shows that standard features often already support such robustness once the head is refitted \citep{Kirichenko2023DFR}. These methods change how a model is trained or which head is fitted. PhysioTRACE instead audits a fixed head on a frozen encoder: it changes only the provenance--target association at fixed marginals on the same held-out records and compares the exposed head with a control head fitted without the association, so that the excess vulnerability isolates reliance from ordinary distribution shift.

\begin{figure}[t]
\centering
\includegraphics[width=\linewidth]{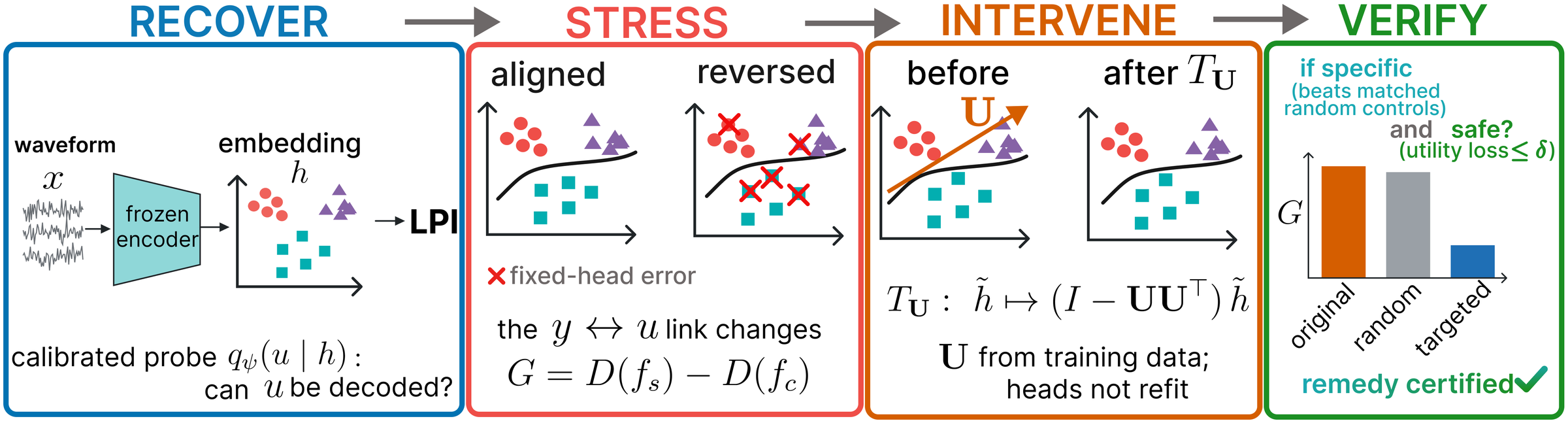}
\caption{The four PhysioTRACE axes. \textbf{Recover} measures how decodable provenance $u$ is from the frozen embedding $h$ with a calibrated probe, reported as LPI (Equation~\ref{eq:lpi}). \textbf{Stress} changes only the $y$--$u$ association on the same records and compares the fixed exposed and control heads through $G$ (Equations~\ref{eq:workshop-fixed-marginals}--\ref{eq:excess-vulnerability}). \textbf{Intervene} removes a train-localized component $\mathbf U$ with $T_{\mathbf U}$ (Equation~\ref{eq:workshop-intervention}); $\tilde h$ denotes train-whitened coordinates. \textbf{Verify} certifies the remedy only if it beats matched random controls and keeps the utility loss within $\delta$ (Section~\ref{sec:reliance-audit}). Colors and shapes denote provenance groups.}
\label{fig:four-axis-audit}
\end{figure}

\textbf{Concept erasure.} Iterative null-space projection \citep{Ravfogel2020INLP}, amnesic probing with random-direction controls \citep{Elazar2021Amnesic}, and closed-form least-squares erasure \citep{Belrose2023LEACE} remove a linearly decodable property and measure the consequence. Erasure presumes that the property should go. When sources differ in whom they recorded, that presumption also removes target-relevant population signal; erasure further does not establish that a head relies on the property, nor certify that removal preserves the task. PhysioTRACE uses a standard whitened projection, localized from training data by paired within-record contrasts or a target-conditional probe, and adds the decision rule: suppression is certified only if it beats matched random controls and keeps utility within a declared margin, and it can be refused (Appendix Table~\ref{tab:erasure-delta}).

\textbf{Physiological benchmarks and metadata.} MOABB, TUH EEG, PTB-XL, and recent PFM benchmarks evaluate cross-task and cross-dataset transfer \citep{Jayaram2018,Obeid2016,Wagner2020,Strodthoff2021,Xiong2025,Kastrati2025,Aristimunha2025,Wan2025} but usually fold acquisition into domain labels. Building on BIDS and EEG-BIDS \citep{Gorgolewski2016,Pernet2019} and dataset-documentation frameworks \citep{Bender2018,Gebru2021,Mitchell2019,Pushkarna2022}, PhysioTRACE turns reference, device, site, and preprocessing metadata into provenance groups with identity-disjoint splits (models covered: Appendix Tables~\ref{tab:app-pfm-data-coverage} and~\ref{tab:app-pfm-methods}).

\section{The PhysioTRACE Protocol}
\label{sec:protocol}
PhysioTRACE asks one main question about a frozen encoder: whether a task head built on it relies on how its inputs were recorded. A complete answer takes four steps (Figure~\ref{fig:four-axis-audit}), each addressing what the previous one leaves open. \textbf{Recover} asks whether there is anything to rely on, by measuring how much provenance the embedding exposes. Accessibility does not show use, since a head can ignore what a probe decodes. \textbf{Stress} therefore looks at behavior: on held-out records, it reverses which provenance groups co-occur with which targets and checks whether the exposed head, fitted under the original pairing, degrades more than a control head. Detecting reliance does not yet locate or remove it, so \textbf{Intervene} removes a provenance component localized on training data and checks whether the failure disappears. Since a removal can also discard task signal or succeed by chance, \textbf{Verify} certifies it only if it beats matched random controls and keeps utility within a declared margin. Algorithm~\ref{alg:audit} summarizes the complete audit, and Appendix~\ref{app:worked-example} follows one record through all four steps.

Formally, the audit takes embeddings $h_i=f_\theta(x_i)\in\mathbb R^d$, task labels $y_i$, and provenance labels $u_i\in\mathcal G$ with $K=|\mathcal G|$, on an identity-disjoint train/test split. Two linear task heads are fitted on the training embeddings and then frozen. The \emph{exposed head} $f_s$ is fitted under a provenance--target association, either present in the data or induced by the audit design. The \emph{control head} $f_c$ is fitted on the same embeddings with independence weights that make $y$ and $u$ independent in the training distribution. The control head is the comparator throughout: behavior that $f_s$ and $f_c$ share is not attributable to the association. Each audit also declares the role of its provenance factor (Appendix Table~\ref{tab:invariance-shift-types}): a near nuisance, such as a re-referenced view of the same record, or a confounded factor, such as site or device, that also indexes who was recorded. The confounded case is the common one in collected data, and the protocol's defaults are chosen for it.

\subsection{Recover: Linear Provenance Information}
\label{sec:lpi}
We quantify how accessible $u$ is in $h$ as held-out predictive $\mathcal V$-information \citep{Xu2020Usable} for a calibrated linear family. With a multinomial logistic probe $q_\psi(u\mid h)$ and the Jeffreys-smoothed training prevalence $q_0(u)$ as the constant predictor, Linear Provenance Information over $N$ held-out native units is
\begin{equation}
\mathrm{LPI}=\frac1N\sum_{i=1}^N\log_2\frac{q_\psi(u_i\mid h_i)}{q_0(u_i)}.
\label{eq:lpi}
\end{equation}
\begin{algorithm}[t]
\small
\caption{PhysioTRACE audit of one frozen encoder}
\label{alg:audit}
\textbf{Input:} train/test embeddings with $(y,u)$; regimes aligned, independent, reversed; utility metric $\mathcal M$ and margin $\delta$; effect size $\varepsilon$; number of random controls $R$.
\begin{enumerate}\setlength{\itemsep}{1pt}\setlength{\parskip}{0pt}
\item \textbf{Recover.} Fit and calibrate the probe $q_\psi$ on training data; report LPI (Eq.~\ref{eq:lpi}) on test data.
\item Fit $f_s$ under the association and $f_c$ with independence weights on training data; freeze both.
\item \textbf{Stress.} Reweight the test set for each regime (Eq.~\ref{eq:workshop-fixed-marginals}); compute $G(\mathrm{id})$ (Eq.~\ref{eq:excess-vulnerability}).
\item If $G(\mathrm{id})<\varepsilon$ or its lower 95\% bound is not above zero, \textbf{return} \emph{no reliance}.
\item \textbf{Intervene.} Estimate $\mathbf U$ on training data by paired or observational localization (Section~\ref{sec:intervene}); compute $G(T_{\mathbf U})$ with Eq.~\ref{eq:workshop-intervention}.
\item \textbf{Verify.} Compute $G$ for $R$ rank- and energy-matched random subspaces. \emph{Specific}: the reduction $G(\mathrm{id})-G(T_{\mathbf U})$ exceeds every random reduction. \emph{Safe}: the lower 95\% bound of $\mathcal M(T_{\mathbf U})-\mathcal M(\mathrm{id})$ is at least $-\delta$.
\item \textbf{Return} \emph{reliance, remedy certified} if specific and safe; otherwise \emph{reliance, remedy not certified}.
\end{enumerate}
\end{algorithm}
The probe is fitted on training embeddings, selected by validation log loss, and temperature-calibrated on validation logits; no test score enters selection. LPI is measured in bits per native provenance-labeled unit and reported with identity-aware bootstrap intervals that resample subjects, patients, or records together with all of their views. This matters for physiological data: on ERP, whose 15,312 test trials come from four subjects, resampling trials instead would narrow the intervals 14--31-fold and separate 27 rather than 9 of 28 model pairs (Appendix Table~\ref{tab:lpi-naive-ci}). Unlike probe accuracy \citep{Hewitt2019Control,Voita2020MDL}, LPI is calibrated and prevalence-referenced, so values remain interpretable when group balance differs. Like any $\mathcal V$-information estimate it depends on the probe family and its calibration: it is not mutual information, a non-positive value means no gain for this family, and it does not show that a task head uses $u$. That question is answered by the remaining three axes.

\subsection{Stress: fixed-marginal association shift}
A regime $e$ specifies a joint distribution $\pi_e(y,u)$ whose target and provenance marginals are fixed by the training design, so that only the association changes. Each of the same $n_{\mathrm{te}}$ held-out observations is reweighted by
\begin{equation}
\omega_i^{(e)}\propto\pi_e(y_i,u_i)\,/\,\hat p_{\mathrm{te}}(y_i,u_i),
\label{eq:workshop-fixed-marginals}
\end{equation}
normalized to mean one, where $\hat p_{\mathrm{te}}$ is the empirical held-out joint; this requires observed support wherever $\pi_e>0$ (Appendix~\ref{app:workshop-protocol-details}). The aligned regime reproduces the association under which $f_s$ was fitted, the independent regime removes it, and the reversed regime inverts it. Let $\mathcal A_e(f;T)$ be the weighted held-out AUROC of a fixed head $f$ after representation transform $T$ ($T=\mathrm{id}$ without intervention), and $D(f;T)=\mathcal A_{\mathrm{aligned}}(f;T)-\mathcal A_{\mathrm{reversed}}(f;T)$. The \emph{excess vulnerability}
\begin{equation}
G(T)=D(f_s;T)-D(f_c;T)
\label{eq:excess-vulnerability}
\end{equation}
is the part of the exposed head's sensitivity to the association that the control head does not share. Because records, labels, embeddings, and heads are identical across regimes, $G$ measures decision reliance on the association rather than ordinary distribution shift. When each record is observed under several provenance views, a same-record logit contrast and the rate of decision flips across views add record-level evidence (Appendix~\ref{app:workshop-protocol-details}).

\subsection{Intervene: train-localized suppression}
\label{sec:intervene}
Given a train-fitted centering $\mu_{\mathrm{loc}}$, a support-aware whitening operator $W\in\mathbb R^{r\times d}$ onto the effective covariance rank $r$, with pseudoinverse $W^\dagger$, and an orthonormal rank-$p$ basis $\mathbf U$ in whitened coordinates, the intervention
\begin{equation}
T_{\mathbf U}(h)=h-W^\dagger\mathbf U\mathbf U^\top W(h-\mu_{\mathrm{loc}})
\label{eq:workshop-intervention}
\end{equation}
removes the localized component and leaves everything outside it, including directions outside the numerical covariance support, unchanged. Both heads stay fixed. The protocol estimates $\mathbf U$ from training data in one of two ways.

\textit{Paired localization.} When every training record is observed under all $K$ provenance views, we average each view's whitened deviation from its record mean over training records and take $\mathbf U$ as the top $p$ right singular vectors of the resulting $K\times r$ matrix, so $p\le K-1$ (Appendix Equation~\ref{eq:workshop-paired-subspace}). This uses embeddings, record pairing, and provenance labels, but no task labels.

\textit{Observational localization.} Without paired views, provenance groups generally differ in whom they contain and hence in target composition, so an unconditional provenance probe would partly recover the target itself. Localization therefore conditions on the target. We subtract training class means from the standardized embeddings shared by both heads, whiten the residuals, and fit an independence-weighted logistic probe for a binary provenance label; $\mathbf U$ is the unit normal of its separator, not a task-head direction (Appendix Equation~\ref{eq:workshop-observational-subspace}). Fitting uses training task and provenance labels only; applying $T_{\mathbf U}$ to a new embedding needs neither.

\subsection{Verify: specificity, utility, and verdict}
\label{sec:reliance-audit}
Suppression is \emph{specific} if the reduction $G(\mathrm{id})-G(T_{\mathbf U})$ exceeds that of each of $R$ Haar-random rank-$p$ subspaces drawn in the same whitened support and removing the same localization-set energy fraction $p/r$. It is \emph{safe} if the lower 95\% bound of the ordinary-utility change $\mathcal M(T_{\mathbf U})-\mathcal M(\mathrm{id})$ is at least $-\delta$, for a utility metric $\mathcal M$ and margin $\delta$ declared before evaluation. A refitted provenance probe reports residual accessibility but does not enter the decision, since reduced decodability alone says nothing about behavior. The audit returns one of three verdicts (Algorithm~\ref{alg:audit}): \emph{no reliance}, when $G(\mathrm{id})$ is below a declared effect size $\varepsilon$ or not bounded away from zero; \emph{reliance, remedy certified}, when reliance is detected and suppression is specific and safe; and \emph{reliance, remedy not certified}, when reliance is detected but suppression fails either test. Recover alone never yields a verdict.

\Needspace{8\baselineskip}
\section{Applying PhysioTRACE}
\label{sec:controlled-acquisition}\label{sec:real-domain-transfer}
Each subsection tests one of the three claims of Section~\ref{sec:intro} on data suited to it. Q1 compares accessibility and transfer across many models on benchmarks where acquisition is mixed with other domain factors; Q2 uses paired EEG views in which the shortcut is known by construction; Q3 uses real ECG device metadata, in which it is not. Table~\ref{tab:audit-outcomes} summarizes all five audits and their settings, declared before evaluation; seeds, folds, and resampling units are specified in Appendix~\ref{app:evidence-units}.

\subsection{Q1: Can accessibility or transfer stand in for a reliance test?}
\label{sec:broad-screen}

\begin{figure}[t]
\centering
\includegraphics[width=\linewidth]{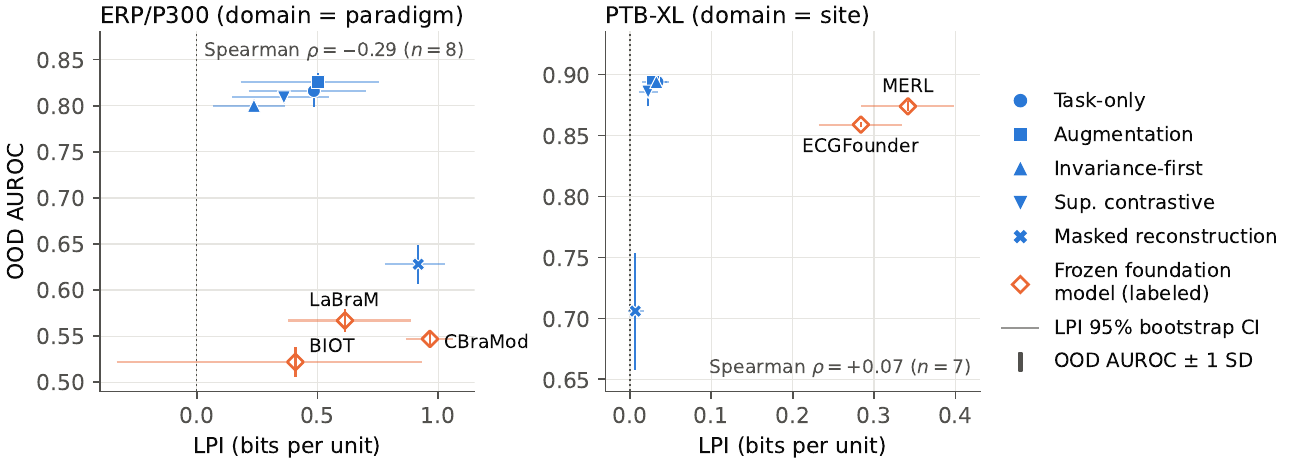}
\caption{Accessibility and transfer decouple. Each point is a shared-backbone objective (blue; five encoder seeds) or a frozen foundation model (orange) on ERP/P300 (left; domains are paradigms) or PTB-XL (right; domains are sites). Horizontal bars: LPI with identity-aware 95\% intervals. Vertical bars: OOD AUROC, $\pm1$ SD over five fits. LPI is comparable only within a dataset; values are in Appendix Table~\ref{tab:lpi-transfer-main}.}
\label{fig:lpi-transfer}
\end{figure}

Q1 uses P300 ERP transfer across amyotrophic lateral sclerosis (ALS) P300, covert GeoSpell, and overt P300 \citep{Riccio2013ALS,Arico2014P300Jitter,Aloise2012Geospell} and PTB-XL site transfer \citep{Wagner2020}, which mix acquisition with population, interface, era, and workflow, and compares five shared-backbone objectives with frozen LaBraM \citep{Jiang2024}, BIOT \citep{Yang2023}, CBraMod \citep{Wang2025CBraMod}, ECGFounder \citep{Li2025ECGFounder}, and MERL \citep{Liu2024MERL}. Shifted transfer and provenance accessibility are the two measurements routinely read as evidence about provenance risk, and Figure~\ref{fig:lpi-transfer} shows that neither predicts the other. On ERP, masked reconstruction, CBraMod, and LaBraM have the highest LPI, and the three frozen EEG models transfer worst. On PTB-XL, masked reconstruction has almost no site information yet transfers worst, whereas MERL and ECGFounder retain the most and transfer well. The sign of the relation thus changes between datasets, and the HMC Sleep Staging Database \citep{AlvarezEstevez2022HMC} reproduces the decoupling for three frozen EEG encoders on a five-class task with four references, using an ordinary reference probe (Appendix~\ref{app:post-hmc}). Neither number can therefore validate the other, and neither measures whether a predictor \emph{uses} provenance: Q2 shows a relied-upon component that transfer metrics cannot see even though a probe recovers it easily.

\subsection{Q2: Does the audit recover a known shortcut?}
\label{sec:nmt-audit}
\begin{table}[t]
\centering\small
\setlength{\tabcolsep}{4pt}
\caption{PhysioTRACE verdicts for all five audits. $G$ (Equation~\ref{eq:excess-vulnerability}) is shown before and after removing the train-localized component, heads fixed. A remedy is certified if it beats all random controls and the 95\% lower bound of the utility change stays above $-\delta$. NMT: rank 2, 10 controls per seed; ECG: rank 1, 100 controls, $\varepsilon=0.01$. Dashes: axis not reached; intervals: Appendix Table~\ref{tab:audit-outcomes-full}.}
\label{tab:audit-outcomes}
\vspace{2pt}
\begin{tabular*}{\linewidth}{@{\extracolsep{\fill}}lccccc@{}}
\toprule
& \multicolumn{2}{c}{Paired EEG (NMT)} & \multicolumn{3}{c}{Observational ECG (PTB-XL devices)} \\
\cmidrule(lr){2-3}\cmidrule(l){4-6}
& Task-only & Task + recon. & ECGFounder & \makecell{ECGFounder,\\weak pair} & CLEF-Small \\
\midrule
Excess vulnerability $G$ & 0.197 & 0.225 & 0.023 & 0.00007 & 0.060 \\
$G$ after suppression & $-$0.001 & 0.001 & 0.002 & -- & 0.022 \\
Beats all random controls & yes & yes & yes & -- & yes \\
Utility change & 0.000 & 0.001 & $-$0.004 & -- & $-$0.039 \\
\quad 95\% lower bound & $-$0.004 & $-$0.001 & $-$0.006 & -- & $-$0.044 \\
\quad tolerated loss $\delta$ & 0.02$^{\ast}$ & 0.02 & 0.01 & -- & 0.01 \\
\midrule
Reliance detected & yes & yes & yes & no & yes \\
Remedy certified & yes & yes & yes & -- & \textbf{no} \\
\bottomrule
\end{tabular*}

\vspace{2pt}
{\raggedright\footnotesize $^{\ast}$Prespecified for the reconstruction encoder; applied to task-only as a secondary check.\par}
\end{table}

Paired NMT is a setting in which the answer is known. Each NMT Scalp EEG record \citep{Khan2022NMT} is observed under average, Cz, and linked-ear reference, so physiology and diagnosis are fixed while only provenance varies; reference is thus a near nuisance (Appendix Table~\ref{tab:invariance-shift-types}). The association is induced by design: the exposed head is trained where one reference co-occurs with the abnormal class, the control head where reference is independent of diagnosis, and each reference serves in turn as the shortcut. We audit ten encoders, five seeds each of a task-only and a task-plus-reconstruction objective, on one record-disjoint split (Appendix~\ref{app:post-nmt}).

\textit{Stress detects the induced reliance.} The control head is unaffected by the association, whereas the exposed head degrades steadily from aligned to reversed (Figure~\ref{fig:acquisition-leakage}(c)), giving $G=0.20$--$0.22$ for both objectives (Table~\ref{tab:audit-outcomes}). Changing only the reference view of the same record flips the exposed head's decision for $44.5\%$ of records, against $13.8\%$ for the control (Appendix~\ref{app:nmt-regimes}). Reliance therefore follows from the association under which the head is fitted, not from the reconstruction objective (Appendix~\ref{app:post-nmt}).

\begin{figure}[t]
\centering
\includegraphics[width=\linewidth]{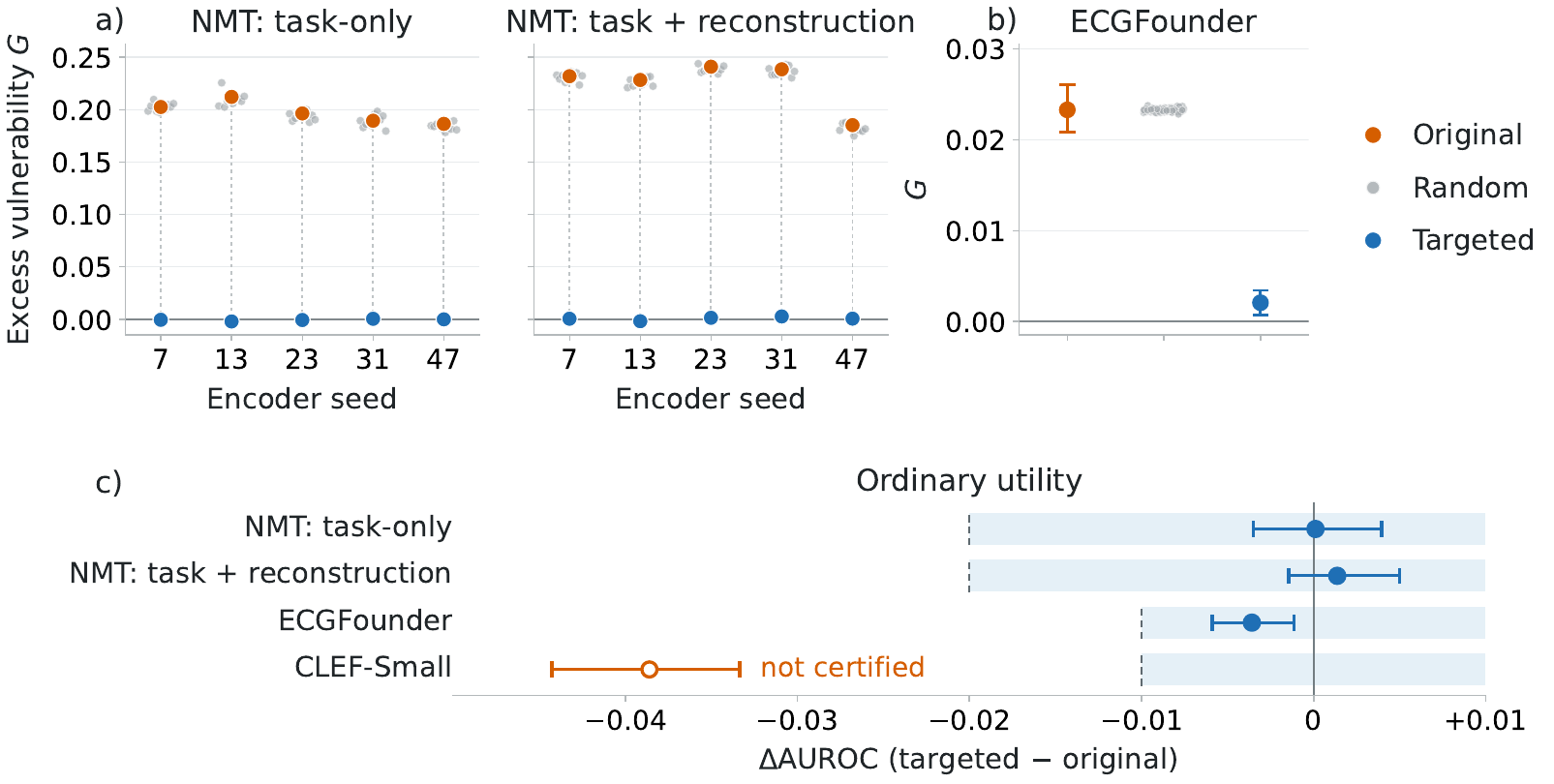}
\caption{Targeted removal is specific, and Verify refuses it when it costs utility. \textbf{(a)} NMT, per encoder seed, averaged over reference rotations; \textbf{(b)} ECGFounder, means over five patient-disjoint folds (each random control is also fold-averaged); the two panels use different $G$ scales. In both, targeted removal reduces $G$ beyond every rank- and energy-matched random control. \textbf{(c)} Utility change (targeted minus original) with 95\% bootstrap intervals (Appendix~\ref{app:evidence-units}); shading marks the declared margin, 0.02 for NMT and 0.01 for ECG. NMT and ECGFounder stay inside it and are certified; CLEF-Small does not.}
\label{fig:fixed-head-audit-summary}
\end{figure}

\textit{Intervene and Verify certify the remedy.} Removing the rank-two reference subspace, localized without diagnosis labels, brings $G$ to within $0.003$ of zero in every encoder seed and returns the exposed head to control-level behavior, whereas matched random subspaces leave $G$ unchanged (Figure~\ref{fig:fixed-head-audit-summary}(a)). For task-plus-reconstruction encoders, a refitted reference probe falls from $92\%$ to $36\%$ (chance $33\%$), yet in-domain and cross-reference AUROC change by less than $0.002$. The relied-upon component thus carries no measurable ordinary utility: transfer metrics are blind to it, and only $G$ shows that the head relies on it. Both objectives are certified.

\subsection{Q3: Does the audit work on observational data, and can it refuse?}
\label{sec:ecg-audit}
Observational data offer neither paired views nor a designed association. We use frozen embeddings of 8,888 PTB-XL ECGs from the two most frequent devices at site 0, whose NORM prevalence differs markedly ($82\%$ versus $29\%$), with five patient-disjoint outer folds. The exposed head is fitted on this natural association and the control head with independence weights; every device-by-target cell is populated in every held-out fold (Appendix~\ref{app:post-ecg-sensitivity}). Device is a confounded factor (Appendix Table~\ref{tab:invariance-shift-types}): in PTB-XL it also indexes recording era, workflow, and case mix. A head fitted on age, sex, and recording year alone reaches $G=0.061$ on this cohort (Appendix Table~\ref{tab:post-ecg-sensitivity}), more than the ECGFounder embedding; the device association is thus largely one of patient mix and workflow. This is the typical observational setting (Section~\ref{sec:intro}), and the audit accordingly tests reliance on the device-associated collection signal that a deployed head would encounter, not a hardware effect in isolation.

\textit{ECGFounder: reliance, remedy certified.} Device remains decodable within diagnosis strata (conditional AUROC $0.96$), and the exposed head is more vulnerable than its control in every fold ($G=0.023$). Removing the rank-one direction reduces $G$ by $91\%$, beyond all 100 random controls, and flattens the exposed head's response across all five association strengths (Figure~\ref{fig:fixed-head-audit-summary}(b); Appendix Table~\ref{tab:clef-curves}), while utility stays inside the margin (Table~\ref{tab:audit-outcomes}). The only loss is the aligned-regime advantage the shortcut provided. A refitted probe stays above chance: the remedy removes only the component the head relies on and leaves the rest of the device-associated signal in place, as intended when device also indexes who was recorded.

\textit{Weak-association pair: no reliance.} On a site-2 device pair whose natural device--diagnosis odds ratio is only $1.04$, the same audit under stress of the primary pair's strength gives $G=0.00007$, below $\varepsilon$ and not bounded away from zero, so it stops before Intervene (Table~\ref{tab:audit-outcomes}; Appendix~\ref{app:post-ecg-sensitivity}). The stress procedure therefore does not manufacture excess vulnerability where the head has no association to exploit.

\textit{CLEF-Small: reliance, remedy not certified.} On the same ECGs, folds, and regimes, the frozen single-lead CLEF-Small encoder \citep{Shu2025CLEF} is more than twice as vulnerable ($G=0.060$), about as vulnerable as the age/sex/year head (excess $-0.002$ [$-0.010$, $0.007$]; Appendix~\ref{app:clef}). It is also less device-accessible than ECGFounder (conditional AUROC $0.90$ versus $0.96$; Appendix Table~\ref{tab:audit-outcomes-full}), so within one audit accessibility and reliance order the two encoders oppositely, as Q1 predicts. Rank-one projection removes $63\%$ of $G$ and beats every random control, but costs $0.039$ AUROC against a $0.01$ margin (Figure~\ref{fig:fixed-head-audit-summary}(c)), and the projected head's response inverts (Appendix~\ref{app:clef}). The localized direction carries task signal together with device signal, the entangled case described in Section~\ref{sec:intro}, so the audit detects the reliance but refuses the linear remedy.

\Needspace{6\baselineskip}
\section{Discussion, Scope, and Conclusion}
\label{sec:discussion}
A physiological foundation model can pass every transfer benchmark and still carry a deployable shortcut, and a fixed-head association stress reveals it where transfer metrics do not. The three questions of Section~\ref{sec:controlled-acquisition} make this concrete. The relation between accessibility and shifted transfer changes sign between datasets, so neither can stand in for a reliance test (Q1). When the association is induced by design, the audit detects the reliance, localizes a low-rank component that carries no ordinary utility, and certifies its removal for both training objectives (Q2). On observational ECG it certifies one remedy, returns no reliance where the device carries almost no association with diagnosis, and refuses a remedy that also removes task signal (Q3). For PFM evaluation, this means that transfer and probe scores cannot support a claim that a predictor is robust to how its inputs were recorded; that claim needs a behavioral test.

Selectivity is the appropriate default rather than a refinement. Because collected sources differ in whom they recorded, removing whatever a probe can decode would, by default, remove target-relevant signal with the artifact; for CLEF-Small it would have cost $0.039$ AUROC. The audit avoids this because it localizes conditionally, compares against a control head and matched random projections, and certifies a remedy only within a declared utility margin. Where it does certify, the remedy removes the advantage the shortcut gave the exposed head under its training association and keeps ordinary utility within that margin, which is what a fix should do.

\textbf{Scope and limitations.} A certified remedy supports the evaluated association and linear intervention, not safety under every deployment shift. Empirical endpoints are binary and multiclass classification; regression and forecasting need endpoint-specific loss and horizon-aware stress designs and remain untested extensions. NMT is record-disjoint rather than established patient-disjoint, and its five-seed audit covers one architecture at $d=128$. ERP test support comprises four physical subjects, so we report leave-one-subject-out sensitivity alongside cluster intervals. The ECG device effect is not separated from era and case mix beyond the metadata comparators of Appendix~\ref{app:post-ecg-sensitivity}. Linear probes, low-rank projections, and conditional bootstrap intervals do not establish nonlinear invariance or complete removal. A rank-one observational localization can capture task signal along with provenance; Verify detects this, as for CLEF-Small, and higher-rank localizations that separate the two are natural extensions it can adjudicate.

The audit is not specific to physiology. Scanner/site, staining batch, sequencing platform, and dataset/study identity can similarly index how observations were produced rather than the intended measurement, and the observational mode needs only identity-disjoint splits, composition checks, target-conditional probes, control heads, and sufficient joint support. PhysioTRACE turns the claim that a physiological foundation model is robust to how its inputs were recorded into a test that can pass, fail, or refuse a remedy, so that the claim can be checked rather than assumed.

\Needspace{8\baselineskip}
\subsection*{Ethics statement}
The experiments are secondary analyses of previously published physiological datasets and existing checkpoints under their access conditions. The authors did not recruit participants, collect new human-subject data, or deploy diagnostic models. Raw physiological recordings and external checkpoints are not redistributed. Section~\ref{sec:discussion} discusses the risks of acquisition-associated shortcuts and the limits of the audit's interpretation.

\subsection*{Reproducibility statement}
Section~\ref{sec:protocol} and Appendices~\ref{app:workshop-protocol-details}, \ref{app:post-lpi}, \ref{app:post-nmt}, and~\ref{app:clef} specify estimands, splits, hyperparameter selection, calibration, localization, controls, and bootstrap units; Appendix~\ref{app:evidence-units} lists seeds, folds, and resampling units per experiment. Code, benchmark schemas and cards, experiment runners, and the saved numerical artifacts (CSV/JSON results with source and input hashes, exported probe states, bootstrap draws, and pre-test selection locks) are available at \url{\repourl}. Hardware and software versions are listed in Appendix~\ref{app:compute}, and dataset and checkpoint terms in Appendix~\ref{app:asset_licenses}.

\bibliographystyle{plainnat}
\bibliography{references}

@article{Aloise2012Geospell,
  author = {Aloise, Fabio and Aric{\`o}, Pietro and Schettini, Francesca and Riccio, Angela and Salinari, Serenella and Mattia, Donatella and Babiloni, Fabio and Cincotti, Febo},
  title = {A Covert Attention {P300}-Based Brain--Computer Interface: {GeoSpell}},
  journal = {Ergonomics},
  volume = {55},
  number = {5},
  pages = {538--551},
  year = {2012},
  doi = {10.1080/00140139.2012.661084},
  url = {https://doi.org/10.1080/00140139.2012.661084}
}

@misc{AlvarezEstevez2022HMC,
  author = {Diego Alvarez-Estevez and Roselyne Rijsman},
  title = {{Haaglanden Medisch Centrum} Sleep Staging Database},
  howpublished = {PhysioNet},
  year = {2022},
  note = {Version 1.1},
  doi = {10.13026/t79q-fr32},
  url = {https://doi.org/10.13026/t79q-fr32}
}

@article{Arico2014P300Jitter,
  author = {Aric{\`o}, Pietro and Aloise, Fabio and Schettini, Francesca and Salinari, Serenella and Mattia, Donatella and Cincotti, Febo},
  title = {Influence of {P300} Latency Jitter on Event Related Potential-Based Brain--Computer Interface Performance},
  journal = {Journal of Neural Engineering},
  volume = {11},
  number = {3},
  pages = {035008},
  year = {2014},
  doi = {10.1088/1741-2560/11/3/035008},
  url = {https://doi.org/10.1088/1741-2560/11/3/035008}
}

@article{Aristimunha2025,
  author = {Bruno Aristimunha and Dung Truong and Pierre Guetschel and Seyed Yahya Shirazi and Isabelle Guyon and others},
  title = {{EEG} Foundation Challenge: From Cross-Task to Cross-Subject {EEG} Decoding},
  journal = {arXiv preprint arXiv:2506.19141},
  year = {2025},
  url = {https://arxiv.org/abs/2506.19141},
  doi = {10.48550/arXiv.2506.19141}
}

@inproceedings{Belrose2023LEACE,
  author = {Nora Belrose and David Schneider-Joseph and Shauli Ravfogel and Ryan Cotterell and Edward Raff and Stella Biderman},
  title = {{LEACE}: Perfect Linear Concept Erasure in Closed Form},
  booktitle = {Advances in Neural Information Processing Systems},
  volume = {36},
  year = {2023},
  url = {https://proceedings.neurips.cc/paper_files/paper/2023/hash/d066d21c619d0a78c5b557fa3291a8f4-Abstract-Conference.html},
  pages = {66044--66063},
  doi = {10.52202/075280-2884}
}

@article{Bender2018,
  author = {Emily M. Bender and Batya Friedman},
  title = {Data Statements for Natural Language Processing: Toward Mitigating System Bias and Enabling Better Science},
  journal = {Transactions of the Association for Computational Linguistics},
  volume = {6},
  pages = {587--604},
  year = {2018},
  doi = {10.1162/tacl_a_00041},
  url = {https://aclanthology.org/Q18-1041/}
}

@article{Castro2020,
  title   = {Causality Matters in Medical Imaging},
  author  = {Castro, Daniel C. and Walker, Ian and Glocker, Ben},
  journal = {Nature Communications},
  volume  = {11},
  pages   = {3673},
  year    = {2020},
  doi     = {10.1038/s41467-020-17478-w}
}

@article{Dong2019,
  author = {Li Dong and Xiaobo Liu and Lingling Zhao and Yongxiu Lai and Diankun Gong and Tiejun Liu and Dezhong Yao},
  title = {A Comparative Study of Different {EEG} Reference Choices for Event-Related Potentials Extracted by Independent Component Analysis},
  journal = {Frontiers in Neuroscience},
  volume = {13},
  pages = {1068},
  year = {2019},
  doi = {10.3389/fnins.2019.01068},
  url = {https://www.frontiersin.org/articles/10.3389/fnins.2019.01068}
}

@article{Elazar2021Amnesic,
  author = {Yanai Elazar and Shauli Ravfogel and Alon Jacovi and Yoav Goldberg},
  title = {Amnesic Probing: Behavioral Explanation with Amnesic Counterfactuals},
  journal = {Transactions of the Association for Computational Linguistics},
  volume = {9},
  pages = {160--175},
  year = {2021},
  doi = {10.1162/tacl_a_00359},
  url = {https://aclanthology.org/2021.tacl-1.10/}
}

@article{Gebru2021,
  author = {Timnit Gebru and Jamie Morgenstern and Briana Vecchione and Jennifer Wortman Vaughan and Hanna Wallach and Hal Daum{\'e} III and Kate Crawford},
  title = {Datasheets for Datasets},
  journal = {Communications of the ACM},
  volume = {64},
  number = {12},
  pages = {86--92},
  year = {2021},
  doi = {10.1145/3458723},
  url = {https://doi.org/10.1145/3458723}
}

@article{Geirhos2020,
  author = {Robert Geirhos and J{\"o}rn-Henrik Jacobsen and Claudio Michaelis and Richard Zemel and Wieland Brendel and Matthias Bethge and Felix A. Wichmann},
  title = {Shortcut Learning in Deep Neural Networks},
  journal = {Nature Machine Intelligence},
  volume = {2},
  number = {11},
  pages = {665--673},
  year = {2020},
  doi = {10.1038/s42256-020-00257-z},
  url = {https://www.nature.com/articles/s42256-020-00257-z}
}

@article{Goldberger2000PhysioNet,
  author = {Goldberger, Ary L. and Amaral, Luis A. N. and Glass, Leon and Hausdorff, Jeffrey M. and Ivanov, Plamen Ch. and Mark, Roger G. and Mietus, Joseph E. and Moody, George B. and Peng, Chung-Kang and Stanley, H. Eugene},
  title = {{PhysioBank}, {PhysioToolkit}, and {PhysioNet}: Components of a New Research Resource for Complex Physiologic Signals},
  journal = {Circulation},
  volume = {101},
  number = {23},
  pages = {e215--e220},
  year = {2000},
  doi = {10.1161/01.CIR.101.23.e215},
  url = {https://doi.org/10.1161/01.CIR.101.23.e215}
}

@article{Gorgolewski2016,
  author = {Krzysztof J. Gorgolewski and Tibor Auer and Vince D. Calhoun and R. Cameron Craddock and Samir Das and Eugene P. Duff and Guillaume Flandin and Satrajit S. Ghosh and Tristan Glatard and Yaroslav O. Halchenko and Daniel A. Handwerker and Michael Hanke and David Keator and Xiangrui Li and Zachary Michael and Camille Maumet and B. Nolan Nichols and Thomas E. Nichols and John Pellman and Jean-Baptiste Poline and Ariel Rokem and Gunnar Schaefer and Vanessa Sochat and William Triplett and Jessica A. Turner and Ga{\"e}l Varoquaux and Russell A. Poldrack},
  title = {The Brain Imaging Data Structure, a Format for Organizing and Describing Outputs of Neuroimaging Experiments},
  journal = {Scientific Data},
  volume = {3},
  pages = {160044},
  year = {2016},
  doi = {10.1038/sdata.2016.44},
  url = {https://www.nature.com/articles/sdata201644}
}

@inproceedings{Hewitt2019Control,
  title     = {Designing and Interpreting Probes with Control Tasks},
  author    = {Hewitt, John and Liang, Percy},
  booktitle = {Proceedings of the 2019 Conference on Empirical Methods in Natural Language Processing and the 9th International Joint Conference on Natural Language Processing (EMNLP-IJCNLP)},
  pages     = {2733--2743},
  year      = {2019},
  doi       = {10.18653/v1/D19-1275}
}

@article{Jayaram2018,
  author = {Vinay Jayaram and Alexandre Barachant},
  title = {{MOABB}: Trustworthy Algorithm Benchmarking for {BCIs}},
  journal = {Journal of Neural Engineering},
  volume = {15},
  number = {6},
  pages = {066011},
  year = {2018},
  doi = {10.1088/1741-2552/aadea0},
  url = {https://iopscience.iop.org/article/10.1088/1741-2552/aadea0}
}

@inproceedings{Jiang2024,
  author = {Wei-Bang Jiang and Li-Ming Zhao and Bao-Liang Lu},
  title = {Large Brain Model for Learning Generic Representations with Tremendous {EEG} Data in {BCI}},
  booktitle = {The Twelfth International Conference on Learning Representations},
  year = {2024},
  url = {https://openreview.net/forum?id=QzTpTRVtrP}
}

@article{Kastrati2025,
  author = {Ard Kastrati and Josua B{\"u}rki and Jonas Lauer and Cheng Xuan and Raffaele Iaquinto and Roger Wattenhofer},
  title = {{EEG-Bench}: A Benchmark for {EEG} Foundation Models in Clinical Applications},
  journal = {arXiv preprint arXiv:2512.08959},
  year = {2025},
  url = {https://arxiv.org/abs/2512.08959},
  doi = {10.48550/arXiv.2512.08959}
}

@article{Kayser2010,
  author = {J{\"u}rgen Kayser and Craig E. Tenke},
  title = {In Search of the {Rosetta Stone} for Scalp {EEG}: Converging on Reference-Free Techniques},
  journal = {Clinical Neurophysiology},
  volume = {121},
  number = {12},
  pages = {1973--1975},
  year = {2010},
  doi = {10.1016/j.clinph.2010.04.030},
  url = {https://doi.org/10.1016/j.clinph.2010.04.030}
}

@article{Khan2022NMT,
  author = {Khan, Hassan Aqeel and {Ul Ain}, Rahat and Kamboh, Awais Mehmood and Butt, Hammad Tanveer and Shafait, Saima and Alamgir, Wasim and Stricker, Didier and Shafait, Faisal},
  title = {{The NMT Scalp EEG Dataset}: An Open-Source Annotated Dataset of Healthy and Pathological {EEG} Recordings for Predictive Modeling},
  journal = {Frontiers in Neuroscience},
  volume = {15},
  pages = {755817},
  year = {2022},
  doi = {10.3389/fnins.2021.755817},
  url = {https://doi.org/10.3389/fnins.2021.755817}
}

@inproceedings{Kirichenko2023DFR,
  title     = {Last Layer Re-Training is Sufficient for Robustness to Spurious Correlations},
  author    = {Kirichenko, Polina and Izmailov, Pavel and Wilson, Andrew Gordon},
  booktitle = {International Conference on Learning Representations},
  year      = {2023},
  url       = {https://openreview.net/forum?id=THOOBy1uWVH}
}

@inproceedings{Koh2021,
  author = {Pang Wei Koh and Shiori Sagawa and Henrik Marklund and Sang Michael Xie and Marvin Zhang and Akshay Balsubramani and Weihua Hu and Michihiro Yasunaga and Richard Lanas Phillips and Irena Gao and Tony Lee and Etienne David and Ian Stavness and Wei Guo and Berton Earnshaw and Imran Haque and Sara Beery and Jure Leskovec and Anshul Kundaje and Emma Pierson and Sergey Levine and Chelsea Finn and Percy Liang},
  title = {{WILDS}: A Benchmark of in-the-Wild Distribution Shifts},
  booktitle = {Proceedings of the 38th International Conference on Machine Learning},
  volume = {139},
  pages = {5637--5664},
  year = {2021},
  url = {https://proceedings.mlr.press/v139/koh21a.html}
}

@article{Kostas2021,
  author = {Demetres Kostas and St{\'e}phane Aroca-Ouellette and Frank Rudzicz},
  title = {{BENDR}: Using Transformers and a Contrastive Self-Supervised Learning Task to Learn from Massive Amounts of {EEG} Data},
  journal = {Frontiers in Human Neuroscience},
  volume = {15},
  pages = {653659},
  year = {2021},
  doi = {10.3389/fnhum.2021.653659},
  url = {https://doi.org/10.3389/fnhum.2021.653659}
}

@article{Kuruppu2025,
  author = {Gayal Kuruppu and Neeraj Wagh and Vaclav Kremen and Sandipan Pati and Gregory Worrell and Yogatheesan Varatharajah},
  title = {{EEG} Foundation Models: A Critical Review of Current Progress and Future Directions},
  journal = {arXiv preprint arXiv:2507.11783},
  year = {2025},
  url = {https://arxiv.org/abs/2507.11783},
  doi = {10.48550/arXiv.2507.11783}
}

@article{Lee2025SynthSleepNet,
  author = {Cheol-Hui Lee and Hakseung Kim and Byung Chul Yoon and Dong-Joo Kim},
  title = {Toward Foundational Model for Sleep Analysis Using a Multimodal Hybrid-Self-Supervised Learning Framework},
  journal = {IEEE Transactions on Cybernetics},
  year = {2025},
  doi = {10.1109/TCYB.2025.3603608}
}

@article{Li2025ECGFounder,
  author = {Jun Li and Aaron D. Aguirre and Valdery Moura Junior and Jiarui Jin and Che Liu and Lanhai Zhong and Chenxi Sun and Gari Clifford and M. Brandon Westover and Shenda Hong},
  title = {An Electrocardiogram Foundation Model Built on over 10 Million Recordings},
  journal = {{NEJM} AI},
  volume = {2},
  number = {7},
  pages = {AIoa2401033},
  year = {2025},
  doi = {10.1056/AIoa2401033},
  url = {https://ai.nejm.org/doi/10.1056/AIoa2401033}
}

@inproceedings{Liu2024MERL,
  author = {Che Liu and Zhongwei Wan and Cheng Ouyang and Anand Shah and Wenjia Bai and Rossella Arcucci},
  title = {Zero-Shot {ECG} Classification with Multimodal Learning and Test-time Clinical Knowledge Enhancement},
  booktitle = {Proceedings of the 41st International Conference on Machine Learning},
  series = {Proceedings of Machine Learning Research},
  volume = {235},
  pages = {31949--31963},
  year = {2024},
  publisher = {PMLR},
  url = {https://proceedings.mlr.press/v235/liu24bg.html}
}

@inproceedings{Makar2022Shortcut,
  title     = {Causally Motivated Shortcut Removal Using Auxiliary Labels},
  author    = {Makar, Maggie and Packer, Ben and Moldovan, Dan and Blalock, Davis and Halpern, Yoni and D'Amour, Alexander},
  booktitle = {Proceedings of the 25th International Conference on Artificial Intelligence and Statistics},
  series    = {Proceedings of Machine Learning Research},
  volume    = {151},
  pages     = {739--766},
  year      = {2022},
  url       = {https://proceedings.mlr.press/v151/makar22a.html}
}

@article{McKeen2025,
  author = {Kaden McKeen and Sameer Masood and Augustin Toma and Barry Rubin and Bo Wang},
  title = {{ECG}-{FM}: An Open Electrocardiogram Foundation Model},
  journal = {JAMIA Open},
  volume = {8},
  number = {5},
  pages = {ooaf122},
  year = {2025},
  doi = {10.1093/jamiaopen/ooaf122},
  url = {https://academic.oup.com/jamiaopen/article/8/5/ooaf122/8287827}
}

@inproceedings{Mitchell2019,
  author = {Margaret Mitchell and Simone Wu and Andrew Zaldivar and Parker Barnes and Lucy Vasserman and Ben Hutchinson and Elena Spitzer and Inioluwa Deborah Raji and Timnit Gebru},
  title = {Model Cards for Model Reporting},
  booktitle = {Proceedings of the Conference on Fairness, Accountability, and Transparency},
  pages = {220--229},
  year = {2019},
  doi = {10.1145/3287560.3287596},
  url = {https://dl.acm.org/doi/10.1145/3287560.3287596}
}

@book{Nunez2006,
  author = {Paul L. Nunez and Ramesh Srinivasan},
  title = {Electric Fields of the Brain: The Neurophysics of {EEG}},
  edition = {2nd},
  publisher = {Oxford University Press},
  address = {New York},
  year = {2006},
  doi = {10.1093/acprof:oso/9780195050387.001.0001},
  url = {https://doi.org/10.1093/acprof:oso/9780195050387.001.0001}
}

@article{Obeid2016,
  author = {Iyad Obeid and Joseph Picone},
  title = {The {Temple University Hospital} {EEG} Data Corpus},
  journal = {Frontiers in Neuroscience},
  volume = {10},
  pages = {196},
  year = {2016},
  doi = {10.3389/fnins.2016.00196},
  url = {https://www.frontiersin.org/articles/10.3389/fnins.2016.00196}
}

@inproceedings{Ouahidi2025,
  author = {Yassine El Ouahidi and Jonathan Lys and Philipp Th{\"o}lke and Nicolas Farrugia and Bastien Pasdeloup and Vincent Gripon and Karim Jerbi and Giulia Lioi},
  title = {{REVE}: A Foundation Model for {EEG} -- Adapting to Any Setup with Large-Scale Pretraining on 25,000 Subjects},
  booktitle = {Advances in Neural Information Processing Systems},
  volume = {38},
  pages = {22541--22577},
  year = {2025},
  url = {https://proceedings.neurips.cc/paper_files/paper/2025/hash/20a917f77773ac0fa8bea2bdd6606b66-Abstract-Conference.html},
  doi = {10.52202/085713-0760}
}

@misc{PTBXLPhysioNet2020,
  author = {Wagner, Patrick and Strodthoff, Nils and Bousseljot, Ralf-Dieter and Samek, Wojciech and Schaeffter, Tobias},
  title = {{PTB-XL, a Large Publicly Available Electrocardiography Dataset}},
  year = {2020},
  month = apr,
  howpublished = {PhysioNet, version 1.0.1},
  doi = {10.13026/x4td-x982},
  url = {https://doi.org/10.13026/x4td-x982}
}

@article{Pernet2019,
  author = {Cyril R. Pernet and Stefan Appelhoff and Krzysztof J. Gorgolewski and Guillaume Flandin and Christophe Phillips and Arnaud Delorme and Robert Oostenveld},
  title = {{EEG}-{BIDS}, an Extension to the Brain Imaging Data Structure for Electroencephalography},
  journal = {Scientific Data},
  volume = {6},
  pages = {103},
  year = {2019},
  doi = {10.1038/s41597-019-0104-8},
  url = {https://www.nature.com/articles/s41597-019-0104-8}
}

@inproceedings{Puli2022NuRD,
  title     = {Out-of-distribution Generalization in the Presence of Nuisance-Induced Spurious Correlations},
  author    = {Puli, Aahlad Manas and Zhang, Lily H. and Oermann, Eric Karl and Ranganath, Rajesh},
  booktitle = {International Conference on Learning Representations},
  year      = {2022},
  url       = {https://openreview.net/forum?id=12RoR2o32T}
}

@inproceedings{Pushkarna2022,
  author = {Mahima Pushkarna and Andrew Zaldivar and Oddur Kjartansson},
  title = {Data Cards: Purposeful and Transparent Dataset Documentation for Responsible {AI}},
  booktitle = {Proceedings of the 2022 {ACM} Conference on Fairness, Accountability, and Transparency},
  pages = {1776--1826},
  year = {2022},
  doi = {10.1145/3531146.3533231},
  url = {https://dl.acm.org/doi/10.1145/3531146.3533231}
}

@inproceedings{Ravfogel2020INLP,
  author = {Shauli Ravfogel and Yanai Elazar and Hila Gonen and Michael Twiton and Yoav Goldberg},
  title = {Null It Out: Guarding Protected Attributes by Iterative Nullspace Projection},
  booktitle = {Proceedings of the 58th Annual Meeting of the Association for Computational Linguistics},
  pages = {7237--7256},
  year = {2020},
  doi = {10.18653/v1/2020.acl-main.647},
  url = {https://aclanthology.org/2020.acl-main.647/}
}

@article{Riccio2013ALS,
  author = {Riccio, Angela and Simione, Luca and Schettini, Francesca and Pizzimenti, Alessia and Inghilleri, Maurizio and Belardinelli, Marta Olivetti and Mattia, Donatella and Cincotti, Febo},
  title = {Attention and {P300}-Based {BCI} Performance in People with Amyotrophic Lateral Sclerosis},
  journal = {Frontiers in Human Neuroscience},
  volume = {7},
  pages = {732},
  year = {2013},
  doi = {10.3389/fnhum.2013.00732},
  url = {https://doi.org/10.3389/fnhum.2013.00732}
}

@article{Roberts2021,
  author = {Michael Roberts and Derek Driggs and Matthew Thorpe and Julian Gilbey and Michael Yeung and Stephan Ursprung and Angelica I. Aviles-Rivero and Christian Etmann and Cathal McCague and Lucian Beer and Jonathan R. Weir-McCall and Zhongzhao Teng and Effrossyni Gkrania-Klotsas and AIX-COVNET and James H. F. Rudd and Evis Sala and Carola-Bibiane Sch{\"o}nlieb},
  title = {Common Pitfalls and Recommendations for Using Machine Learning to Detect and Prognosticate for {COVID}-19 Using Chest Radiographs and {CT} Scans},
  journal = {Nature Machine Intelligence},
  volume = {3},
  number = {3},
  pages = {199--217},
  year = {2021},
  doi = {10.1038/s42256-021-00307-0},
  url = {https://www.nature.com/articles/s42256-021-00307-0}
}

@inproceedings{Sagawa2020GroupDRO,
  title     = {Distributionally Robust Neural Networks for Group Shifts: On the Importance of Regularization for Worst-Case Generalization},
  author    = {Sagawa, Shiori and Koh, Pang Wei and Hashimoto, Tatsunori B. and Liang, Percy},
  booktitle = {International Conference on Learning Representations},
  year      = {2020},
  url       = {https://openreview.net/forum?id=ryxGuJrFvS}
}

@article{Shu2025CLEF,
  author = {Yuxuan Shu and Peter H. Charlton and Fahim Kawsar and Jussi Hernesniemi and Mohammad Malekzadeh},
  title = {{CLEF}: Clinically-Guided Contrastive Learning for Electrocardiogram Foundation Models},
  journal = {arXiv preprint arXiv:2512.02180},
  year = {2025},
  url = {https://arxiv.org/abs/2512.02180},
  doi = {10.48550/arXiv.2512.02180}
}

@article{Song2025CREMA,
  author = {Junho Song and Jong-Hwan Jang and DongGyun Hong and {Joon-myoung} Kwon and Yong-Yeon Jo},
  title = {{CREMA}: A Contrastive Regularized Masked Autoencoder for Robust {ECG} Diagnostics across Clinical Domains},
  journal = {arXiv preprint arXiv:2407.07110},
  year = {2025},
  url = {https://arxiv.org/abs/2407.07110},
  doi = {10.48550/arXiv.2407.07110}
}

@article{Strodthoff2021,
  author = {Nils Strodthoff and Patrick Wagner and Tobias Schaeffter and Wojciech Samek},
  title = {Deep Learning for {ECG} Analysis: Benchmarks and Insights from {PTB}-{XL}},
  journal = {{IEEE} Journal of Biomedical and Health Informatics},
  volume = {25},
  number = {5},
  pages = {1519--1528},
  year = {2021},
  doi = {10.1109/JBHI.2020.3022989},
  url = {https://ieeexplore.ieee.org/document/9190034}
}

@inproceedings{Veitch2021Counterfactual,
  title     = {Counterfactual Invariance to Spurious Correlations in Text Classification},
  author    = {Veitch, Victor and D'Amour, Alexander and Yadlowsky, Steve and Eisenstein, Jacob},
  booktitle = {Advances in Neural Information Processing Systems},
  volume    = {34},
  year      = {2021},
  url       = {https://proceedings.neurips.cc/paper_files/paper/2021/hash/8710ef761bbb29a6f9d12e4ef8e4379c-Abstract.html},
  pages = {16196--16208}
}

@inproceedings{Voita2020MDL,
  title     = {Information-Theoretic Probing with Minimum Description Length},
  author    = {Voita, Elena and Titov, Ivan},
  booktitle = {Proceedings of the 2020 Conference on Empirical Methods in Natural Language Processing (EMNLP)},
  pages     = {183--196},
  year      = {2020},
  doi       = {10.18653/v1/2020.emnlp-main.14}
}

@article{Wagner2020,
  author = {Patrick Wagner and Nils Strodthoff and Ralf-Dieter Bousseljot and Dieter Kreiseler and F. I. Lunze and Wojciech Samek and Tobias Schaeffter},
  title = {{PTB}-{XL}, a Large Publicly Available Electrocardiography Dataset},
  journal = {Scientific Data},
  volume = {7},
  pages = {154},
  year = {2020},
  doi = {10.1038/s41597-020-0495-6},
  url = {https://www.nature.com/articles/s41597-020-0495-6}
}

@article{Wan2025,
  author = {Zhijiang Wan and Qianhao Yu and Jia Mao and Wenfeng Duan and Cheng Ding},
  title = {{OpenECG}: Benchmarking {ECG} Foundation Models with Public 1.2 Million Records},
  journal = {arXiv preprint arXiv:2503.00711},
  year = {2025},
  url = {https://arxiv.org/abs/2503.00711},
  doi = {10.48550/arXiv.2503.00711}
}

@inproceedings{Wang2024,
  author = {Guangyu Wang and Wenchao Liu and Yuhong He and Cong Xu and Lin Ma and Haifeng Li},
  title = {{EEGPT}: Pretrained Transformer for Universal and Reliable Representation of {EEG} Signals},
  booktitle = {Advances in Neural Information Processing Systems},
  volume = {37},
  pages = {39249--39280},
  year = {2024},
  url = {https://proceedings.neurips.cc/paper_files/paper/2024/hash/4540d267eeec4e5dbd9dae9448f0b739-Abstract-Conference.html},
  doi = {10.52202/079017-1239}
}

@inproceedings{Wang2025CBraMod,
  author = {Jiquan Wang and Sha Zhao and Zhiling Luo and Yangxuan Zhou and Haiteng Jiang and Shijian Li and Tao Li and Gang Pan},
  title = {{CBraMod}: A Criss-Cross Brain Foundation Model for {EEG} Decoding},
  booktitle = {The Thirteenth International Conference on Learning Representations},
  year = {2025},
  url = {https://openreview.net/forum?id=NPNUHgHF2w}
}

@article{Xiong2025,
  author = {Wei Xiong and Jiangtong Li and Jie Li and Kun Zhu and Changjun Jiang},
  title = {{EEG-FM-Bench}: A Comprehensive Benchmark for the Systematic Evaluation and Diagnostic Analyses of {EEG} Foundation Models},
  journal = {arXiv preprint arXiv:2508.17742},
  year = {2025},
  url = {https://arxiv.org/abs/2508.17742},
  doi = {10.48550/arXiv.2508.17742}
}

@inproceedings{Xu2020Usable,
  title     = {A Theory of Usable Information under Computational Constraints},
  author    = {Xu, Yilun and Zhao, Shengjia and Song, Jiaming and Stewart, Russell and Ermon, Stefano},
  booktitle = {International Conference on Learning Representations},
  year      = {2020},
  url       = {https://openreview.net/forum?id=r1eBeyHFDH}
}

@inproceedings{Yang2023,
  author = {Chaoqi Yang and M. Brandon Westover and Jimeng Sun},
  title = {{BIOT}: Biosignal Transformer for Cross-Data Learning in the Wild},
  booktitle = {Advances in Neural Information Processing Systems},
  volume = {36},
  year = {2023},
  url = {https://proceedings.neurips.cc/paper_files/paper/2023/hash/f6b30f3e2dd9cb53bbf2024402d02295-Abstract-Conference.html},
  pages = {78240--78260},
  doi = {10.52202/075280-3420}
}

@article{Yao2001,
  author = {Dezhong Yao},
  title = {A Method to Standardize a Reference of Scalp {EEG} Recordings to a Point at Infinity},
  journal = {Physiological Measurement},
  volume = {22},
  number = {4},
  pages = {693--711},
  year = {2001},
  doi = {10.1088/0967-3334/22/4/305},
  url = {https://iopscience.iop.org/article/10.1088/0967-3334/22/4/305}
}

@article{Zech2018,
  author = {John R. Zech and Marcus A. Badgeley and Manway Liu and Anthony B. Costa and Joseph J. Titano and Eric Karl Oermann},
  title = {Variable Generalization Performance of a Deep Learning Model to Detect Pneumonia in Chest Radiographs: A Cross-Sectional Study},
  journal = {PLOS Medicine},
  volume = {15},
  number = {11},
  pages = {e1002683},
  year = {2018},
  doi = {10.1371/journal.pmed.1002683},
  url = {https://journals.plos.org/plosmedicine/article?id=10.1371/journal.pmed.1002683}
}
\clearpage
\appendix
\raggedbottom
\renewcommand{\topfraction}{0.9}\renewcommand{\bottomfraction}{0.8}\renewcommand{\textfraction}{0.07}\renewcommand{\floatpagefraction}{0.85}\setcounter{topnumber}{3}\setcounter{bottomnumber}{2}\setcounter{totalnumber}{5}
\etocdepthtag.toc{appendix}
\etocsettagdepth{main}{none}
\etocsettagdepth{appendix}{subsection}
\etocsettocstyle{\noindent{\large\bfseries Appendix contents}\par\medskip}{\clearpage}
\tableofcontents
\begingroup
\color{revisiongreen}
\FloatBarrier
\section{Full PhysioTRACE estimands and localization}
\label{app:workshop-protocol-details}\label{app:post-protocol}

{This section gives the complete reporting estimands, fixed-marginal stress construction, mode-specific localization operators, and verification criteria summarized in Section~\ref{sec:protocol}.}

\FloatBarrier
\subsection{Shifted-utility reporting}

{Let $\mathcal G$ be the set of provenance groups, with $K=|\mathcal G|$, and let $S_{a\rightarrow b}$ be a higher-is-better downstream score, such as balanced accuracy or AUROC, from a head fitted on provenance group $a$ and evaluated on group $b$. The common shifted-utility layer reports}
\begin{equation}
\begin{aligned}
 S_{\mathrm{ID}}&=\frac{1}{K}\sum_{a\in\mathcal G}S_{a\rightarrow a}, &
 S_{\mathrm{OOD}}&=\frac{1}{K(K-1)}\sum_{\substack{a,b\in\mathcal G\\a\neq b}}S_{a\rightarrow b},\\
 S_{\mathrm{worst}}&=\min_{\substack{a,b\in\mathcal G\\a\neq b}}S_{a\rightarrow b}, &
 \Delta_{\mathrm{shift}}&=S_{\mathrm{ID}}-S_{\mathrm{OOD}}.
\end{aligned}
\label{eq:workshop-transfer-metrics}
\end{equation}
{A mixed-provenance head has no single source group $a$. If its score on target group $b$ is $S_{\mathrm{mix}\rightarrow b}$, its shifted utility is $K^{-1}\sum_b S_{\mathrm{mix}\rightarrow b}$ and its worst-group utility is $\min_b S_{\mathrm{mix}\rightarrow b}$; these are reported separately from the diagonal and off-diagonal source-specific estimands.}

\FloatBarrier
\subsection{Fixed-marginal stress and reliance estimands}

{Each regime joint $\pi_e$ keeps the target and provenance marginals fixed by the training design rather than estimated from test data, and the normalized density-ratio weights of Equation~\ref{eq:workshop-fixed-marginals} are
\begin{equation}
\begin{gathered}
\sum_u\pi_e(y,u)=\pi_Y(y),\qquad\sum_y\pi_e(y,u)=\pi_U(u),\\
\tilde\omega_i^{(e)}=\frac{\pi_e(y_i,u_i)}{\hat p_{\mathrm{te}}(y_i,u_i)},\qquad
\omega_i^{(e)}=\frac{\tilde\omega_i^{(e)}}{n_{\mathrm{te}}^{-1}\sum_j\tilde\omega_j^{(e)}},
\end{gathered}
\label{eq:stress-weights-full}
\end{equation}
which are well defined under the overlap condition $\hat p_{\mathrm{te}}(y,u)>0$ wherever $\pi_e(y,u)>0$.}
{The core regimes are aligned, independent, and reversed; aligned reproduces the association used to fit the exposed head. NMT evaluates all $K$ real views and fixes $\pi_Y(y)=1/2$ and $\pi_U(u)=1/K$. For observational ECG fold $k$, let $\kappa_k>1$ be the natural device--target odds ratio (OR) magnitude estimated strictly from its outer-training design set. The five regimes satisfy $\operatorname{OR}(\pi_{k,e})=\kappa_k^{\alpha_e}$ for $\alpha_e\in\{1,\frac12,0,-\frac12,-1\}$, corresponding to aligned, intermediate aligned, independent, intermediate reversed, and reversed. Scores and $G$ are computed within each outer fold and then averaged equally across folds.}

{For NMT, one shortcut head $f_s^{(g)}$ is fitted for each designated shortcut reference $g\in\mathcal G$, together with one shared control head $f_c$. The reported $G$ averages the per-$g$ instance of Equation~\ref{eq:excess-vulnerability}. Paired data also permit a direct same-record reliance statistic under representation transform $T$. Let $h_{i,u}=f_\theta(x_i^{(u)})$ be the embedding of held-out record $i$ under view $u$, let $\eta_f$ be the task logit, and let $n_y$ be the number of held-out records with class $y$. Define the class-balanced record expectation as $\mathbb E_i^{\mathrm{bal}}[a_i]=\frac12\sum_{y\in\{0,1\}}n_y^{-1}\sum_{i:y_i=y}a_i$. Then}
\begin{equation}
\begin{aligned}
\Lambda_{\mathrm{pair}}(T)=\frac{1}{K}\sum_{g\in\mathcal G}\mathbb E_i^{\mathrm{bal}}
\Bigg[&\eta_{f_s^{(g)}}(T(h_{i,g}))
-\frac{1}{K-1}\sum_{u\neq g}\eta_{f_s^{(g)}}(T(h_{i,u}))\\
&-\eta_{f_c}(T(h_{i,g}))
+\frac{1}{K-1}\sum_{u\neq g}\eta_{f_c}(T(h_{i,u}))\Bigg].
\end{aligned}
\label{eq:workshop-paired-reliance}
\end{equation}
{Because all views belong to the same record, $\Lambda_{\mathrm{pair}}(T)$ is a signed excess logit contrast under provenance change while physiology and diagnosis remain fixed, class-balanced across records and averaged over shortcut-reference rotations. We report $\Lambda_{\mathrm{pair}}(\mathrm{id})$ before intervention and $\Lambda_{\mathrm{pair}}(T_{\mathbf U})$ afterward.}

\FloatBarrier
\subsection{Train-localized suppression}

{Let $H_{\mathrm{loc,tr}}=[\xi_1^\top;\ldots;\xi_{N_{\mathrm{loc}}}^\top]\in\mathbb R^{N_{\mathrm{loc}}\times d}$ be the train-only localization matrix. Its empirical mean and covariance are $\mu_{\mathrm{loc}}=N_{\mathrm{loc}}^{-1}\sum_j\xi_j$ and $\Sigma_{\mathrm{loc}}=N_{\mathrm{loc}}^{-1}\sum_j(\xi_j-\mu_{\mathrm{loc}})(\xi_j-\mu_{\mathrm{loc}})^\top$. The support-aware whitening operator $W\in\mathbb R^{r\times d}$ satisfies $W\Sigma_{\mathrm{loc}}W^\top=I_r$, where $r$ is the effective covariance rank. For a predeclared rank $p\le r$, the orthonormal basis $\mathbf U\in\mathbb R^{r\times p}$ satisfies $\mathbf U^\top\mathbf U=I_p$ and is applied through Equation~\ref{eq:workshop-intervention}. The subtraction form preserves components outside the numerical localization support.}

\paragraph{{Paired localization.}}
{For $n$ training records observed under all $K$ provenance views, $H_{\mathrm{loc,tr}}$ contains the $N_{\mathrm{loc}}=nK$ embeddings $h_{i,u}=f_\theta(x_i^{(u)})$. In whitened coordinates $\tilde h=W(h-\mu_{\mathrm{loc}})$, within-record centering removes the component shared across views:}
\begin{equation}
\begin{aligned}
\bar{\tilde h}_i&=\frac{1}{K}\sum_{u=1}^{K}\tilde h_{i,u},&
m_u&=\frac{1}{n}\sum_{i=1}^{n}(\tilde h_{i,u}-\bar{\tilde h}_i),\\
M&=[m_1^\top;\ldots;m_K^\top]=L\Sigma_MV^\top,&
\mathbf U&=V_p.
\end{aligned}
\label{eq:workshop-paired-subspace}
\end{equation}
{Here $L$, $\Sigma_M$, and $V$ are the thin SVD factors, and $V_p\in\mathbb R^{r\times p}$ contains the first $p$ right singular vectors. Because $\sum_u m_u=0$, $p\le\operatorname{rank}(M)\le K-1$. NMT confirms full contrast rank and uses the predeclared $p=K-1=2$. Localization uses training embeddings, matched record identities, and provenance labels, but not diagnosis labels, validation/test embeddings, or task-head weights.}

\paragraph{{Observational localization.}}
{For ECG outer fold $k$, let $\check h_{k,i}=A_k(h_i)$ be the output of the affine StandardScaler fitted on that outer-training fold and shared by both fixed task heads. Define train-estimated target means $\mu_{k,y}=\mathbb E_{\mathrm{tr},k}[\check h_k\mid Y=y]$, residuals $\rho_i=\check h_{k,i}-\mu_{k,y_i}$, and their training mean $\bar\rho_k$. The localization matrix contains the outer-training residuals, and $W_k$ is fitted only on their covariance. We whiten them as $\tilde\rho_i=W_k(\rho_i-\bar\rho_k)$ and fit an independence-weighted binary device probe, where $\sigma$ is the sigmoid:}
\begin{equation}
q_{\phi,k}(u=1\mid\tilde\rho_i)=\sigma(v_k^\top\tilde\rho_i+b_k),\qquad
\mathbf U_k=\frac{v_k}{\lVert v_k\rVert}\in\mathbb R^{r_k\times 1}.
\label{eq:workshop-observational-subspace}
\end{equation}
{Conditional device AUROC is computed separately within each target stratum and outer test fold, then averaged equally without pooling scores across folds. Residualization is used for conditional probing and localization, but the projection is applied to the full standardized embedding:}
\begin{equation}
T_{k,\mathbf U_k}(\check h)=
\check h-W_k^\dagger\mathbf U_k\mathbf U_k^\top W_k(\check h-\bar\rho_k).
\label{eq:workshop-observational-intervention}
\end{equation}
{Both already-fitted task heads consume $T_{k,\mathbf U_k}(\check h)$. Target labels inform outer-training conditional localization, but applying the projection to a new embedding requires neither its target nor its provenance label. The operation builds on linear concept-erasure methods such as iterative null-space projection (INLP) \citep{Ravfogel2020INLP}, Amnesic Probing \citep{Elazar2021Amnesic}, and least-squares concept erasure (LEACE) \citep{Belrose2023LEACE}; PhysioTRACE contributes acquisition-specific paired or conditional localization and behavioral verification.}

\FloatBarrier
\subsection{Verification criteria}

{Let the utility metric $\mathcal M(T)$ be a prespecified higher-is-better ordinary-utility estimand evaluated with its own already-fitted head or heads. We report}
\begin{equation}
\Delta_G(\mathbf U)=G(\mathrm{id})-G(T_{\mathbf U}),\qquad
\Delta\mathcal M(\mathbf U)=
\mathcal M(T_{\mathbf U})-\mathcal M(\mathrm{id}).
\label{eq:workshop-verification}
\end{equation}
{For the NMT reconstruction encoder, the prespecified utility metric $\mathcal M$ is mean in-domain AUROC from separate source-reference-specific heads with $\delta=0.02$; task-only uses the same margin as a secondary check, and mean off-diagonal cross-reference AUROC is an additional preservation check. For ECG, $\mathcal M$ is the independent-regime AUROC of the fixed control head with $\delta=0.01$.}

{Let $\widetilde H_{\mathrm{loc,tr}}\in\mathbb R^{N_{\mathrm{loc}}\times r}$ contain the whitened rows from the exact localization support. By construction, $\widetilde H_{\mathrm{loc,tr}}^\top\widetilde H_{\mathrm{loc,tr}}=N_{\mathrm{loc}}I_r$. Specificity is tested with $R$ Haar-random orthonormal bases $Q_j\in\mathbb R^{r\times p}$, $Q_j^\top Q_j=I_p$, in that same support. Every rank-$p$ basis removes the same aggregate localization-set energy:}
\begin{equation}
\frac{\lVert\widetilde H_{\mathrm{loc,tr}}Q_j\rVert_F^2}
{\lVert\widetilde H_{\mathrm{loc,tr}}\rVert_F^2}
=\frac{p}{r}
=\frac{\lVert\widetilde H_{\mathrm{loc,tr}}\mathbf U\rVert_F^2}
{\lVert\widetilde H_{\mathrm{loc,tr}}\rVert_F^2},
\qquad j=1,\ldots,R.
\label{eq:workshop-energy-matching}
\end{equation}
{A complete Verify result requires that targeted suppression reduce $G$ beyond these rank- and energy-matched controls, that a newly fitted provenance probe quantify residual accessibility without altering either task head, and that the lower 95\% confidence bound satisfies $\operatorname{LCB}_{95\%}[\Delta\mathcal M(\mathbf U)]\ge-\delta$. For the $d=128$ NMT audit, intervals use 2,000 crossed encoder-seed and label-stratified record bootstrap draws over five encoder seeds and the same 185 held-out records. Each record draw is shared across objectives, views, task heads, interventions, and random controls; seeds are resampled as matched objective pairs. The split remains record-disjoint, not patient-disjoint. ECG intervals use 2,000 shared within-fold patient-cluster bootstraps. NMT inference includes variability across the five fitted encoder seeds, whereas ECG inference conditions on the frozen checkpoint. Both condition on the split, fitted task heads, association tables, train-localized subspaces, and random-control realizations. The NMT seed-level diagnostics and utility check are detailed in Appendix~\ref{app:post-nmt}; calibrated accessibility uses the separate uncertainty protocol in Appendix~\ref{app:post-lpi}. These criteria establish selective suppression of a train-localized, behaviorally relevant linear provenance component, not complete linear or nonlinear invariance.}

\FloatBarrier
\subsection{Relation to linear erasure and amnesic probing}
\label{app:erasure-delta}
Table~\ref{tab:erasure-delta} separates what PhysioTRACE reuses from prior linear-removal methods and what it adds. The projection operator is not new; the added elements define what is stressed, against which comparator, and what counts as a certified remedy.

\begin{table*}[htbp]
\centering\small
\setlength{\tabcolsep}{4pt}
\caption{PhysioTRACE relative to INLP \citep{Ravfogel2020INLP}, Amnesic Probing \citep{Elazar2021Amnesic}, and LEACE \citep{Belrose2023LEACE}.}
\label{tab:erasure-delta}
\begin{tabular}{@{}>{\raggedright\arraybackslash}p{1.9cm}>{\raggedright\arraybackslash}p{3.1cm}>{\raggedright\arraybackslash}p{3.1cm}>{\raggedright\arraybackslash}p{5.0cm}@{}}
\toprule
Method & Removal operator & How the removed component is chosen & How removal is evaluated \\
\midrule
INLP & Iterated null-space projection & Successive linear classifiers for the property & Property decodability after removal; downstream bias or task metrics \\
Amnesic Probing & INLP projection & Linear classifiers for the property & Change in the model's own task behavior, against random directions of equal dimension \\
LEACE & Closed-form least-squares erasure & Cross-covariance of embedding and property & Guarantee that no linear classifier beats a constant predictor; downstream effects \\
\midrule
PhysioTRACE & Whitened rank-$p$ projection (Eq.~\ref{eq:workshop-intervention}) & Paired within-record view contrasts, or a target-conditional probe fitted on training data & \textbf{Added:} fixed-marginal association stress on the same held-out records (Eq.~\ref{eq:workshop-fixed-marginals}); excess vulnerability against a control head fitted without the association (Eq.~\ref{eq:excess-vulnerability}); paired same-record contrast (Eq.~\ref{eq:workshop-paired-reliance}); fixed task heads; utility check against a declared margin. \textbf{Reused:} rank- and energy-matched random controls \\
\bottomrule
\end{tabular}
\end{table*}

\endgroup

\FloatBarrier
\section{Worked example: one NMT record through the four axes}
\label{app:worked-example}

We follow encoder seed 13 of the task-plus-reconstruction objective and held-out record 513, a normal recording observed in all three reference views. All values are read from the saved audit outputs.

\paragraph{Recover.} A calibrated reference probe on this encoder's embeddings reaches $93.3\%$ accuracy on the 555 held-out record views (chance $33.3\%$) and $\mathrm{LPI}=1.31\,[1.22,1.38]$ bits per view, against a $\log_2 3=1.585$-bit prevalence baseline (Equation~\ref{eq:lpi}). Reference is therefore highly accessible, but this alone says nothing about decisions.

\paragraph{Stress.} Two linear heads are fitted on the same training embeddings. The exposed head sees a training set in which the designated shortcut reference is associated with the abnormal class; the control head sees reference independent of diagnosis. On the same 185 held-out records, reweighted by Equation~\ref{eq:workshop-fixed-marginals}, the exposed head's AUROC falls from $0.828$ (aligned) through $0.722$ (independent) to $0.599$ (reversed), while the control stays at $0.768$ in all three regimes. Averaging over shortcut rotations gives $G=(0.828-0.599)-(0.768-0.768)=0.228$ (Equation~\ref{eq:excess-vulnerability}). For record 513 with AVG as the shortcut, the exposed head assigns abnormal probability $0.759$ to the AVG view and $0.140$ and $0.185$ to the CZREF and LE views: the same recording changes class at the $0.5$ threshold only because its reference changed. The control assigns $0.356$, $0.335$, and $0.316$.

\paragraph{Intervene.} Paired localization (Equation~\ref{eq:workshop-paired-subspace}) uses only training records, their three views, and reference labels. In this encoder's whitened support of effective rank $96$, the rank-two basis removes $2/96=2.1\%$ of the localization energy. After suppression (Equation~\ref{eq:workshop-intervention}) with both heads unchanged, the exposed head assigns $0.423$, $0.342$, and $0.316$ to record 513, so all three views are classified normal, and a refitted reference probe falls to $37.8\%$.

\paragraph{Verify.} Across all records, $G$ falls from $0.228$ to $-0.002$, whereas each of the ten rank- and energy-matched random subspaces leaves it between $0.221$ and $0.231$. Mean in-domain AUROC changes from $0.768$ to $0.770$ ($+0.002\,[-0.003,0.008]$), inside the $-0.02$ margin. The component the exposed head relied on is thus localized, specific, and removable without measurable utility loss for this encoder; Table~\ref{tab:audit-outcomes} aggregates the same computation over five seeds.

{\color{revisiongreen}
\FloatBarrier
\section{Calibrated accessibility on ERP/P300 and PTB-XL}
\label{app:post-lpi}\label{app:lpi-naive-ci}

\paragraph{Coverage and role in the audit.}
We complete calibrated Linear Provenance Information (LPI) reporting for 55 cached-embedding conditions: five shared-backbone objectives at encoder seeds $7,13,23,42,52$ in each benchmark, plus frozen LaBraM, BIOT, and CBraMod on ERP/P300 and frozen ECGFounder and MERL on PTB-XL. These yield 15 model--benchmark summaries. The five frozen checkpoints each receive one deterministic calibrated probe; repeated evaluation fits are not independent encoder seeds. This analysis completes the selected ERP/PTB \textbf{Recover} grid without retraining encoders or task heads, or changing the transfer and behavioral audits.

\begin{table}[htbp]
\centering\small
\setlength{\tabcolsep}{4pt}
\caption{Transfer and calibrated provenance accessibility. AUROC: mean $\pm$ SD over five training/evaluation fits. OOD and worst: mean and minimum AUROC over held-out source--target group pairs (Appendix Equation~\ref{eq:workshop-transfer-metrics}). LPI: bits per ERP trial or PTB ECG, with identity-aware 95\% bootstrap intervals; compare only within datasets. ERP includes four test subjects. Figure~\ref{fig:lpi-transfer} plots these values; the complete grid is in Table~\ref{tab:post-lpi-complete}.}
\label{tab:lpi-transfer-main}
\begin{tabular*}{\linewidth}{@{\extracolsep{\fill}}llccc@{}}
\toprule
Setting & Model / objective & OOD AUROC & Worst AUROC & LPI [95\% interval] \\
\midrule
ERP & Task-only & $0.816\pm0.017$ & $0.754\pm0.013$ & $0.485\ [0.215,0.702]$ \\
& Augmentation-only & $0.826\pm0.010$ & $0.755\pm0.020$ & $0.503\ [0.182,0.754]$ \\
& Masked reconstruction & $0.628\pm0.021$ & $0.514\pm0.046$ & $0.917\ [0.779,1.028]$ \\
& CBraMod, frozen & $0.547\pm0.008$ & $0.515\pm0.005$ & $0.966\ [0.865,1.061]$ \\
\midrule
PTB & Task-only & $0.894\pm0.005$ & $0.854\pm0.008$ & $0.034\ [0.020,0.048]$ \\
& Masked reconstruction & $0.706\pm0.048$ & $0.638\pm0.051$ & $0.0067\ [-0.0019,0.0176]$ \\
& ECGFounder, frozen & $0.859\pm0.002$ & $0.790\pm0.004$ & $0.284\ [0.232,0.335]$ \\
& MERL, frozen & $0.874\pm0.004$ & $0.822\pm0.018$ & $0.342\ [0.284,0.398]$ \\
\bottomrule
\end{tabular*}
\end{table}

\paragraph{Probability probe and prevalence baseline.}
A StandardScaler and an unweighted multinomial logistic regression are fitted on training embeddings only. We select $C\in\{0.001,0.01,0.1,1,10\}$ by uncalibrated validation negative log likelihood, breaking ties toward smaller $C$, then fit one temperature $\tau\in[0.05,20]$ on the selected validation logits. The probe random state is 42; no train-plus-validation refit or probe-specific subsampling is used. For training group count $n^{\mathrm{tr}}_u$, the frequency-only baseline is $q_0(u)=(n^{\mathrm{tr}}_u+0.5)/(N_{\mathrm{tr}}+0.5K)$. For native held-out unit $i$,
\begin{equation}
\ell_{ji}=\log_2 q_{\psi,j}(u_i\mid h_{ji})-\log_2 q_0(u_i),
\qquad
\widehat{\mathrm{LPI}}=\frac{1}{J}\sum_{j=1}^{J}\frac{1}{N}\sum_{i=1}^{N}\ell_{ji},
\label{eq:post-lpi-native}
\end{equation}
where $J=5$ encoder seeds for trained variants and $J=1$ for a frozen checkpoint. Units are bits per ERP trial or ECG record, not bits per participant. LPI is the held-out log-score gain of this probe procedure, not exact mutual information or evidence that a task head uses provenance. Zero or negative LPI means no improvement over the training-prevalence baseline for this procedure on this support, not absence of information for all decoders. These unweighted calibrated probes are distinct from the class-weighted classification-accuracy probes of Appendix~\ref{app:post-broad}; their accuracy values are not interchangeable with LPI.

\paragraph{Support and physical identity.}
Trials and ECGs remain separate observations for scoring. Table~\ref{tab:post-lpi-support} reports their independent resampling units. ERP group counts in ALS/overt/covert order are $(21000,12096,12096)$ for training, $(4200,1728,1728)$ for validation, and $(8400,3456,3456)$ for test. PTB site counts are $(2500,2500,2500)$, $(500,500,500)$, and $(487,500,500)$, respectively. The PTB cohort is the existing site-capped benchmark, not a sample of clinical site prevalence. There is no subject/patient overlap across splits.

\begin{table}[htbp]
\centering\color{revisiongreen}
\small
\setlength{\tabcolsep}{5pt}
\caption{Native-unit support and physical clusters. ERP clusters are participants across all sessions and domains; PTB clusters are patients across ECGs. ERP has only four independent test participants and two validation participants.}
\label{tab:post-lpi-support}
\begin{tabular*}{\linewidth}{@{\extracolsep{\fill}}lccc@{}}
\toprule
Benchmark & Train units/clusters & Validation units/clusters & Test units/clusters \\
\midrule
ERP/P300 & $45{,}192/12$ & $7{,}656/2$ & $15{,}312/4$ \\
PTB-XL & $7{,}500/6{,}916$ & $1{,}500/1{,}374$ & $1{,}487/1{,}324$ \\
\bottomrule
\end{tabular*}
\end{table}

Both healthy test participants contribute to both healthy ERP domains; domain-prefixed identifiers must not count them as different people. All sessions and domains of one participant stay together. PTB has 133 test patients with multiple ECGs, up to five per patient; 26 patients have different task labels across their ECGs. We therefore use unstratified patient resampling, not an invented patient-level task label. Embedding and checkpoint hashes, metadata, split support, and participant identities were verified. Trained PTB caches lack sample IDs; their canonical row order was checked against all saved label/group/patient arrays and, for test rows, exact ECG IDs in saved prediction files. Only identity fields, not prediction scores, were used for this check.

\paragraph{Native-weighted cluster uncertainty.}
Let $n_c$ be the native-unit count of physical cluster $c$ and $a_{jc}=\sum_{i\in c}\ell_{ji}$. In bootstrap draw $b$, let $m_{bc}$ be cluster multiplicities and $k_{bj}$ encoder-seed multiplicities, with $\sum_j k_{bj}=J$. We compute
\begin{equation}
\widehat{\mathrm{LPI}}^{(b)}=
\frac{1}{J}\sum_{j=1}^{J} k_{bj}
\frac{\sum_c m_{bc}a_{jc}}{\sum_c m_{bc}n_c}.
\label{eq:post-lpi-bootstrap}
\end{equation}
Thus resampling preserves native-unit weighting rather than taking an equal-cluster mean. We use 10,000 unstratified cluster draws, bootstrap seed 20260908, and percentile 95\% intervals. Trained rows cross encoder-seed and cluster resampling; frozen rows resample clusters conditional on one checkpoint. Compatible models share ordered cluster draws. Probes are not refitted within draws. Intervals condition on the split, training data, fitted probes, and calibration, and are pointwise intervals, not simultaneous model-ranking guarantees.

\begin{table}[htbp]
\centering\color{revisiongreen}
\small
\setlength{\tabcolsep}{3pt}
\caption{Complete calibrated LPI grid: 55 conditions summarized in 15 model--benchmark rows. Each trained entry averages five encoder seeds; each frozen entry uses one checkpoint. ERP values are bits per trial, PTB values bits per ECG. ERP test support comprises four physical subjects; leave-one-subject-out ranges are given below. Different provenance definitions and supports preclude a cross-dataset ranking by absolute LPI.}
\label{tab:post-lpi-complete}
\begin{tabular*}{\linewidth}{@{\extracolsep{\fill}}lcc@{}}
\toprule
Model & ERP LPI [95\% interval] & PTB LPI [95\% interval] \\
\midrule
Task-only & $0.4849\ [0.2152,0.7024]$ & $0.0339\ [0.0200,0.0477]$ \\
Augmentation-only & $0.5026\ [0.1820,0.7541]$ & $0.0287\ [0.0151,0.0437]$ \\
Invariance-first & $0.2368\ [0.0673,0.3665]$ & $0.0326\ [0.0194,0.0458]$ \\
Masked reconstruction & $0.9171\ [0.7792,1.0282]$ & $0.0067\ [-0.0019,0.0176]$ \\
Supervised contrastive & $0.3609\ [0.1467,0.5478]$ & $0.0224\ [0.0110,0.0340]$ \\
\midrule
LaBraM, frozen & $0.6138\ [0.3799,0.8850]$ & n/a \\
BIOT, frozen & $0.4085\ [-0.3320,0.9320]$ & n/a \\
CBraMod, frozen & $0.9655\ [0.8653,1.0612]$ & n/a \\
ECGFounder, frozen & n/a & $0.2838\ [0.2320,0.3348]$ \\
MERL, frozen & n/a & $0.3415\ [0.2843,0.3982]$ \\
\bottomrule
\end{tabular*}
\end{table}

\paragraph{Results and sensitivity.}
Masked reconstruction has the largest ERP LPI among shared-backbone variants at each of five encoder seeds; CBraMod also retains substantial accessibility (Table~\ref{tab:post-lpi-complete}). This is a descriptive within-grid pattern, not a causal reconstruction ablation or a population-level superiority test. In PTB, MERL and ECGFounder have positive site accessibility, whereas masked reconstruction provides little log-score gain. Combined with their transfer profiles, these results motivate joint reporting rather than a universal accessibility--transfer ordering. The PTB site analysis is separate from the ECGFounder device audit, which uses another provenance variable and support.

Because ERP test support comprises four physical subjects, we also report fixed-probe leave-one-subject-out (LOSO) LPI ranges: $[0.3773,0.6030]$ for task-only, $[0.3827,0.6453]$ for augmentation-only, $[0.1740,0.3120]$ for invariance-first, $[0.8627,0.9759]$ for reconstruction, $[0.2697,0.4504]$ for supervised contrastive, $[0.4580,0.7040]$ for LaBraM, $[0.1493,0.7161]$ for BIOT, and $[0.9132,1.0063]$ for CBraMod. These are sensitivity ranges across four omissions, not confidence intervals or refitted folds. BIOT's gain is $-0.647$ bits per trial for one healthy participant; positive LOSO point estimates do not override its zero-crossing cluster interval.

\paragraph{Why identity-aware intervals.}
Table~\ref{tab:lpi-naive-ci} compares the reported intervals with a naive bootstrap that resamples native units independently, recomputed from the saved per-unit log-score gains without refitting any probe. On ERP, where 15,312 test trials come from four physical subjects, the naive intervals are 14--31 times narrower and would separate 27 of 28 model pairs, against 9 with identity-aware resampling; BIOT's naive interval excludes zero, whereas the reported interval includes it. On PTB, where 1,487 ECGs come from 1,324 patients, the two differ by a factor of 1.1--1.5 and separate 13 and 12 of 21 pairs.

\begin{table}[htbp]
\centering\small
\setlength{\tabcolsep}{4pt}
\caption{LPI intervals under naive unit-level resampling versus the reported identity-aware resampling (subjects or patients, crossed with encoder seeds for trained variants). Ratio: reported width divided by naive width. 2,000 naive draws per row.}
\label{tab:lpi-naive-ci}
\begin{tabular*}{\linewidth}{@{\extracolsep{\fill}}llcccc@{}}
\toprule
& Model / objective & LPI & Naive 95\% interval & Reported 95\% interval & Ratio \\
\midrule
ERP & Task-only & 0.485 & [0.474, 0.496] & [0.215, 0.702] & 22.8 \\
 & Augmentation-only & 0.503 & [0.491, 0.514] & [0.182, 0.754] & 24.8 \\
 & Invariance-first & 0.237 & [0.229, 0.245] & [0.067, 0.367] & 19.0 \\
 & Masked reconstruction & 0.917 & [0.908, 0.926] & [0.779, 1.028] & 14.3 \\
 & Supervised contrastive & 0.361 & [0.352, 0.369] & [0.147, 0.548] & 22.9 \\
 & LaBraM, frozen & 0.614 & [0.598, 0.630] & [0.380, 0.885] & 15.5 \\
 & BIOT, frozen & 0.409 & [0.388, 0.428] & [$-$0.332, 0.932] & 31.5 \\
 & CBraMod, frozen & 0.966 & [0.959, 0.972] & [0.865, 1.061] & 15.1 \\
\midrule
PTB & Masked reconstruction & 0.007 & [0.000, 0.013] & [$-$0.002, 0.018] & 1.5 \\
 & Task-only & 0.034 & [0.022, 0.046] & [0.020, 0.048] & 1.2 \\
 & Augmentation-only & 0.029 & [0.018, 0.039] & [0.015, 0.044] & 1.4 \\
 & Supervised contrastive & 0.022 & [0.013, 0.032] & [0.011, 0.034] & 1.2 \\
 & Invariance-first & 0.033 & [0.021, 0.044] & [0.019, 0.046] & 1.2 \\
 & ECGFounder, frozen & 0.284 & [0.239, 0.328] & [0.232, 0.335] & 1.2 \\
 & MERL, frozen & 0.342 & [0.287, 0.394] & [0.284, 0.398] & 1.1 \\
\bottomrule
\end{tabular*}
\end{table}

PTB masked-reconstruction seed 52 yields $-0.001765$ bits with $T=19.999987$, at the predefined upper calibration boundary. This probe was retained and the temperature range was not expanded after test evaluation. All 275 candidate logistic regressions converged; selected models required at most 671 of 5,000 allowed iterations. Table~\ref{tab:post-lpi-diagnostics} reports accuracy, balanced accuracy, and macro-F1 of the same calibrated probes. Near-zero LPI does not certify complete provenance invariance or safe downstream behavior.

\begin{table}[htbp]
\centering\color{revisiongreen}
\small
\setlength{\tabcolsep}{4pt}
\caption{Diagnostics from the calibrated LPI probes, not the class-weighted classification-accuracy probes. Entries average encoder seeds for trained variants and use one checkpoint for frozen models. Accuracy (Acc.), balanced accuracy (BA), and macro-F1 supplement prevalence-referenced LPI.}
\label{tab:post-lpi-diagnostics}
\begin{tabular*}{\linewidth}{@{\extracolsep{\fill}}lcccccc@{}}
\toprule
& \multicolumn{3}{c}{ERP/P300} & \multicolumn{3}{c}{PTB-XL} \\
\cmidrule(lr){2-4}\cmidrule(lr){5-7}
Model & Acc. & BA & Macro-F1 & Acc. & BA & Macro-F1 \\
\midrule
Task-only & 0.6819 & 0.5812 & 0.5802 & 0.4153 & 0.4154 & 0.4137 \\
Augmentation-only & 0.6924 & 0.5904 & 0.5889 & 0.4109 & 0.4110 & 0.4095 \\
Invariance-first & 0.5975 & 0.4872 & 0.4880 & 0.4140 & 0.4140 & 0.4126 \\
Masked reconstruction & 0.8017 & 0.7203 & 0.7193 & 0.3568 & 0.3567 & 0.3512 \\
Supervised contrastive & 0.6446 & 0.5332 & 0.5355 & 0.4020 & 0.4021 & 0.3990 \\
\midrule
LaBraM, frozen & 0.7386 & 0.6555 & 0.6589 & n/a & n/a & n/a \\
BIOT, frozen & 0.6637 & 0.5419 & 0.5434 & n/a & n/a & n/a \\
CBraMod, frozen & 0.7998 & 0.7077 & 0.7077 & n/a & n/a & n/a \\
ECGFounder, frozen & n/a & n/a & n/a & 0.5938 & 0.5933 & 0.5925 \\
MERL, frozen & n/a & n/a & n/a & 0.6200 & 0.6194 & 0.6186 \\
\bottomrule
\end{tabular*}
\end{table}

The train-selected ERP majority predictor selects ALS, with test accuracy $0.548589$; the training-frequency prior has test cross entropy $1.464940$ bits. Tied PTB training frequencies invoke a fixed tie rule selecting site 0, with test accuracy $0.327505$. The test-oracle majority fraction $0.336247$ is descriptive only, not used in selection. The PTB prior is uniform, with cross entropy $1.584963$ bits; a constant predictor has BA $1/3$ in both benchmarks.

\paragraph{Reproducibility and inference scope.}
The model grid and protocol were locked before fitting; all 55 calibrated states and 275 validation candidate scores were jointly locked before test prediction. All conditions, including negative estimates, were retained without test-driven reruns. Its Recover coverage complements the distinct NMT and ECG fixed-head evidence; it does not establish task-head reliance by itself.

The released code includes the ERP/PTB LPI runner, its locked inventories and estimator snapshot, and the resulting per-condition LPI, leave-one-subject-out scores, group supports, confusion counts, native-unit scores, bootstrap draws, and exported probes. Cached embeddings, metadata, and test identities are hash-verified. The NMT calibrated analysis is reported in Appendix~\ref{app:post-nmt}.
}

\FloatBarrier
\section{Complete broad transfer screens}
\label{app:post-broad}
\postrev{The following tables report transfer and classification-probe results separately from the calibrated LPI grid: Table~\ref{tab:nmt-results} for NMT, Table~\ref{tab:erp-results} for ERP/P300, and Table~\ref{tab:ecg-results} for PTB-XL. The NMT screen uses a weighted ResNet encoder and its own set of training objectives, distinct from the matched task-only and task-plus-reconstruction pair audited in Appendix~\ref{app:post-nmt}; it is reported for completeness and for the three frozen EEG encoders, and it enters no audit verdict. Values are mean $\pm$ standard deviation across five encoder/evaluation fits; frozen-model fits reuse one pretrained checkpoint. These standard deviations are not the cluster-bootstrap intervals of Appendix~\ref{app:post-lpi}. Objective-family comparisons differ in training signals and are not causal ablations of a reconstruction term.}
\begin{table*}[htbp]
\centering
\small
\caption{\sloppy NMT reference-shift results. Scores are balanced accuracy across AVG,
CZREF, and LE reference views; embedding-probe accuracy is unavailable for AVG-only. Reference classes are balanced (majority-class accuracy $0.333$), so probe accuracy equals balanced accuracy. LaBraM is an
external frozen EEG encoder audit under the same linear-probe protocol, not a trained objective-family
baseline. \rev{BIOT and CBraMod are additional frozen external encoders; variation reflects repeated evaluation fits of each fixed checkpoint.}}
\label{tab:nmt-results}
\begingroup
\setlength{\tabcolsep}{3pt}
\renewcommand{\arraystretch}{1.08}
\fitwidth{%
\begin{tabular}{lccccc}
\toprule
Variant & In-domain & Shifted & Worst-case & Gap & Reference probe \\
\midrule
AVG-only & 0.666 $\pm$ 0.017 & 0.555 $\pm$ 0.050 & 0.543 $\pm$ 0.052 & 0.111 $\pm$ 0.049 & n/a \\
Contrastive only & 0.629 $\pm$ 0.026 & 0.627 $\pm$ 0.024 & 0.592 $\pm$ 0.043 & 0.002 $\pm$ 0.004 & 0.678 $\pm$ 0.019 \\
Supervised contrastive & 0.632 $\pm$ 0.021 & 0.629 $\pm$ 0.022 & 0.610 $\pm$ 0.028 & 0.003 $\pm$ 0.007 & 0.707 $\pm$ 0.019 \\
Canonical recon. + invariance & 0.622 $\pm$ 0.033 & 0.574 $\pm$ 0.023 & 0.520 $\pm$ 0.022 & 0.048 $\pm$ 0.013 & 0.898 $\pm$ 0.085 \\
\addlinespace
LaBraM frozen & 0.645 $\pm$ 0.022 & 0.631 $\pm$ 0.018 & 0.567 $\pm$ 0.032 & 0.014 $\pm$ 0.005 & 0.739 $\pm$ 0.004 \\
\rev{BIOT frozen} & \rev{0.637 $\pm$ 0.011} & \rev{0.625 $\pm$ 0.007} & \rev{0.590 $\pm$ 0.029} & \rev{0.012 $\pm$ 0.007} & \rev{0.532 $\pm$ 0.003} \\
\rev{CBraMod frozen} & \rev{0.618 $\pm$ 0.018} & \rev{0.608 $\pm$ 0.016} & \rev{0.567 $\pm$ 0.029} & \rev{0.010 $\pm$ 0.007} & \rev{0.800 $\pm$ 0.006} \\
\bottomrule
\end{tabular}
}
\endgroup
\end{table*}

\postrev{A mixed-reference exposure baseline trained on all three references reaches $0.681\pm0.004$ mean-target and $0.674\pm0.004$ worst-target BA; because every target reference is seen in training, its gap is not comparable with the source-held-out gaps above.}

\begin{table*}[htbp]
\centering
\small
\caption{\sloppy Five-seed ERP/P300 domain-transfer summary across ALS P300, covert GeoSpell,
and overt P300. The first five rows are tuned ResNet objective-family runs. LaBraM is an external
frozen-encoder audit under the same source-domain linear-probe protocol; preprocessing details are
in Appendix~\ref{app:frozen-encoder-details}. \rev{BIOT and CBraMod are additional fixed-checkpoint audits.} Domain probes report multiclass accuracy; classes are imbalanced (majority-class accuracy $0.549$). Balanced accuracy of the frozen-encoder domain probes is $0.656$ (LaBraM), $0.548$ (BIOT), and $0.712$ (CBraMod), against a chance level of $0.333$.}

\label{tab:erp-results}
\begingroup
\setlength{\tabcolsep}{3pt}
\renewcommand{\arraystretch}{1.08}
\fitwidth{%
\begin{tabular}{lccccc}
\toprule
Variant & ID & OOD & Worst OOD & Gap & Domain probe \\
\midrule
Masked reconstruction & 0.693 $\pm$ 0.017 & 0.628 $\pm$ 0.021 & 0.514 $\pm$ 0.046 & 0.065 $\pm$ 0.037 & 0.800 $\pm$ 0.008 \\
Task only & 0.818 $\pm$ 0.016 & 0.816 $\pm$ 0.017 & 0.754 $\pm$ 0.013 & 0.002 $\pm$ 0.003 & 0.677 $\pm$ 0.022 \\
Augmentation only & 0.828 $\pm$ 0.009 & 0.826 $\pm$ 0.010 & 0.755 $\pm$ 0.020 & 0.002 $\pm$ 0.002 & 0.690 $\pm$ 0.029 \\
Supervised contrastive & 0.815 $\pm$ 0.005 & 0.809 $\pm$ 0.007 & 0.743 $\pm$ 0.012 & 0.006 $\pm$ 0.004 & 0.638 $\pm$ 0.029 \\
Invariance first & 0.801 $\pm$ 0.008 & 0.800 $\pm$ 0.007 & 0.736 $\pm$ 0.007 & 0.001 $\pm$ 0.001 & 0.582 $\pm$ 0.010 \\
\addlinespace
LaBraM frozen & 0.614 $\pm$ 0.008 & 0.567 $\pm$ 0.012 & 0.477 $\pm$ 0.011 & 0.046 $\pm$ 0.016 & 0.731 $\pm$ 0.012 \\
\rev{BIOT frozen} & \rev{0.532 $\pm$ 0.020} & \rev{0.522 $\pm$ 0.016} & \rev{0.493 $\pm$ 0.015} & \rev{0.010 $\pm$ 0.019} & \rev{0.666 $\pm$ 0.016} \\
\rev{CBraMod frozen} & \rev{0.603 $\pm$ 0.010} & \rev{0.547 $\pm$ 0.008} & \rev{0.515 $\pm$ 0.005} & \rev{0.057 $\pm$ 0.008} & \rev{0.803 $\pm$ 0.002} \\
\bottomrule
\end{tabular}
}
\endgroup
\end{table*}

\begin{table*}[htbp]
\centering
\small
\caption{\sloppy Five-seed PTB-XL cross-site summary over three site groups. The first five rows are
tuned ResNet objective-family runs. ECGFounder is an external 12-lead pretrained checkpoint
evaluated as a frozen encoder under the same linear-probe protocol. \postrev{Task scores are AUROC; site probes report ordinary multiclass accuracy from frozen embeddings.} \rev{MERL is an additional frozen ECG-report-contrastive encoder; we report its site-probe score under the same held-out-group audit.} Majority-class site accuracy is $0.328$; balanced accuracy of the frozen-encoder site probes is $0.563$ (ECGFounder) and $0.599$ (MERL).}
\label{tab:ecg-results}
\begingroup
\setlength{\tabcolsep}{3pt}
\renewcommand{\arraystretch}{1.08}
\fitwidth{%
\begin{tabular}{lccccc}
\toprule
Variant & ID & OOD & Worst OOD & Gap & Site probe \\
\midrule
Masked reconstruction & 0.720 $\pm$ 0.048 & 0.706 $\pm$ 0.048 & 0.638 $\pm$ 0.051 & 0.014 $\pm$ 0.009 & 0.374 $\pm$ 0.034 \\
Task only & 0.898 $\pm$ 0.004 & 0.894 $\pm$ 0.005 & 0.854 $\pm$ 0.008 & 0.004 $\pm$ 0.004 & 0.413 $\pm$ 0.041 \\
Augmentation only & 0.897 $\pm$ 0.003 & 0.894 $\pm$ 0.003 & 0.853 $\pm$ 0.010 & 0.003 $\pm$ 0.003 & 0.385 $\pm$ 0.057 \\
Supervised contrastive & 0.886 $\pm$ 0.013 & 0.886 $\pm$ 0.012 & 0.836 $\pm$ 0.021 & -0.000 $\pm$ 0.004 & 0.380 $\pm$ 0.019 \\
Invariance first & 0.897 $\pm$ 0.005 & 0.894 $\pm$ 0.003 & 0.850 $\pm$ 0.007 & 0.004 $\pm$ 0.003 & 0.427 $\pm$ 0.035 \\
\addlinespace
ECGFounder frozen & 0.872 $\pm$ 0.003 & 0.859 $\pm$ 0.002 & 0.790 $\pm$ 0.004 & 0.013 $\pm$ 0.002 & 0.564 $\pm$ 0.007 \\
\rev{MERL frozen} & \rev{0.892 $\pm$ 0.007} & \rev{0.874 $\pm$ 0.004} & \rev{0.822 $\pm$ 0.018} & \rev{0.018 $\pm$ 0.003} & \rev{0.599 $\pm$ 0.011} \\
\bottomrule
\end{tabular}
}
\endgroup
\end{table*}

{\color{revisiongreen}
\FloatBarrier
\section{Five-seed paired NMT audit}
\label{app:post-nmt}

\paragraph{Matched training and fixed evaluation support.}
The complete paired audit uses EEGNetSmall with a $128$-dimensional embedding and encoder seeds $7,13,23,31,47$. Task-only training sets the reconstruction weight to zero; task-plus-reconstruction uses masked-denoising reconstruction with weight $0.5$ and mask ratio $0.15$. Initialization is identical within each objective pair. Both arms use mean pooling over ten windows per record/reference view, AdamW with learning rate $3\times10^{-4}$ and weight decay $10^{-4}$, batches of $24$ complete record/reference views, dropout $0.1$, and input-noise standard deviation $0.02$. The training budget is $16$ epochs; validation-task-loss early stopping uses patience $5$ and minimum improvement $10^{-4}$, followed by restoration of the best validation checkpoint.

All runs use the same $1{,}871/331/185$ train/validation/test records, with AVG, CZREF, and LE views retained for every record. The test support contains $95$ normal and $90$ abnormal records, hence $555$ provenance-labelled record views. Views and their constituent windows stay in the same split. The split is record-disjoint; patient-disjointness is not asserted. Source, preprocessing, split, paired initialization, and prediction-support identities were checked before aggregation.

\paragraph{Head selection, association stress, and localization.}
Aligned, independent, and reversed regimes preserve the class marginal at $0.5$ and each reference marginal at $1/3$. Each reference serves in turn as the designated shortcut; its aligned probability conditional on the positive class is $0.60$. Regularization $C\in\{0.01,0.1,1,10\}$ is selected once per encoder by balanced-validation AUROC, then shared by its control head and all three exposed heads. Final heads are fitted on train plus validation and remain fixed during intervention. We average $G(T)$ from Equation~\ref{eq:excess-vulnerability} over the three shortcut rotations. Train-only, equally weighted within-record reference contrasts localize the rank-two basis in whitened coordinates; Equation~\ref{eq:workshop-intervention} suppresses it without diagnosis labels or task-head weights entering localization. Each fitted encoder is also evaluated with ten equal-rank, equal-energy random subspaces and refitted reference probes.

\paragraph{Training and test uncertainty.}
Point estimates average the five encoder seeds. Percentile $95\%$ intervals use $2{,}000$ crossed encoder-seed and label-stratified record bootstrap draws. The same seed and record draws are shared across objectives, reference views, heads, interventions, and controls; all views of a sampled record move together. Random controls are averaged within each encoder, not counted as independent training replicates. The intervals condition on the split, training sample, fitted heads, localization rule, and sampled control subspaces. Per-seed checks resample records only. Thus, the aggregate quantifies variation across these five fitted runs and the held-out records without treating their $555$ views as independent observations.

\begin{table}[htbp]
\centering\color{revisiongreen}
\normalsize
\setlength{\tabcolsep}{6pt}
\caption{Seed-level paired NMT audit. Probe columns give reference accuracy before suppression and after refitting on the targeted representation; balanced accuracy is identical because every reference has $185$ test views. Chance and majority accuracy are $1/3$. Every targeted projection has lower residual $G$ than all ten matched random controls for that encoder. These diagnostic probes use validation-selected regularization followed by train-plus-validation fitting, without temperature calibration; they are distinct from the calibrated LPI probes in Table~\ref{tab:post-nmt-lpi}.}
\label{tab:post-nmt-seeds}
\begin{tabular*}{\linewidth}{@{\extracolsep{\fill}}llrrrr@{}}
\toprule
Objective & Seed & $G(\mathrm{id})$ & $G(T_{\mathbf U})$
& Probe, original & Probe, refitted \\
\midrule
Task-only & 7  & $0.2024$ & $-0.0004$ & $0.8468$ & $0.3315$ \\
          & 13 & $0.2121$ & $-0.0020$ & $0.8865$ & $0.3387$ \\
          & 23 & $0.1963$ & $-0.0007$ & $0.8757$ & $0.3604$ \\
          & 31 & $0.1894$ & $ 0.0005$ & $0.9117$ & $0.3459$ \\
          & 47 & $0.1864$ & $ 0.0000$ & $0.9009$ & $0.3514$ \\
\addlinespace
Task + reconstruction & 7  & $0.2318$ & $ 0.0006$ & $0.9117$ & $0.3586$ \\
          & 13 & $0.2282$ & $-0.0018$ & $0.9333$ & $0.3784$ \\
          & 23 & $0.2409$ & $ 0.0015$ & $0.9405$ & $0.3568$ \\
          & 31 & $0.2384$ & $ 0.0027$ & $0.9369$ & $0.3586$ \\
          & 47 & $0.1852$ & $ 0.0007$ & $0.8685$ & $0.3495$ \\
\bottomrule
\end{tabular*}
\end{table}

\paragraph{Specific recovery and ordinary utility.}
Table~\ref{tab:post-nmt-seeds} gives the seed-level values. For reconstruction, mean $G$ falls from $0.2249\,[0.1928,0.2557]$ to $0.0007\,[-0.0019,0.0036]$; the paired reduction is $0.2242\,[0.1923,0.2542]$. The average-random-minus-targeted residual-$G$ contrast is $0.2221\,[0.1910,0.2516]$, establishing specificity beyond the descriptive ten-control ranking. Task-only independently reproduces recovery: $G$ falls from $0.1973\,[0.1727,0.2214]$ to $-0.0005\,[-0.0029,0.0017]$, with reduction $0.1978\,[0.1730,0.2221]$. The corresponding same-record excess logit contrasts fall from $1.620$ to $0.0047$ and from $1.335$ to $0.0035$, respectively. Small signed residuals are retained rather than truncated.

The prespecified utility check is the change in ordinary mean in-domain (ID) AUROC from separate source-reference heads for the reconstruction encoder, with a prespecified lower-confidence-bound margin of $-0.02$. Mean ID AUROC changes from $0.7774$ to $0.7788$, giving $+0.0014\,[-0.0015,0.0050]$ and satisfying the margin. Each reconstruction seed also passes its conditional record-bootstrap check. Applying the same margin to task-only yields $+0.0001\,[-0.0035,0.0040]$, with all five per-seed checks passing; for task-only this margin is a secondary check rather than the primary test. Reconstruction cross-reference AUROC changes by $+0.0005\,[-0.0018,0.0029]$, a separate preservation check rather than a substitute check. The result establishes selective suppression of the measured decision-relevant component, not complete provenance erasure or nonlinear invariance.

\paragraph{Between-objective interpretation.}
The reconstruction-minus-task contrast in original $G$ is $+0.0276\,[0.0084,0.0460]$. Its interpretation includes separately selected head regularization: seed $7$ selects $C=0.01$ for task-only and $C=10$ for reconstruction; all other encoder/objective pairs select $C=0.01$. Architecture and encoder-training controls therefore do not make this a reconstruction-only causal contrast. Crucially, this between-encoder caveat does not affect fixed-head intervention comparisons within an encoder. The induced association is a controlled positive stress test, not an estimate of clinical deployment prevalence.

\paragraph{Calibrated LPI on the matched embeddings.}
A complementary Recover analysis uses the four encoder seeds whose $d=128$ embeddings were retained ($13,23,31,47$), without retraining encoders or task heads. Seed $7$ retains its saved predictions, which suffice for the fixed-head audit above but not for fitting a new calibrated probe. Each probe uses a train-only StandardScaler and unweighted multinomial logistic regression. We select $C\in\{0.001,0.01,0.1,1,10\}$ by uncalibrated validation negative log likelihood, then fit one temperature on validation logits; there is no train-plus-validation refit. All eight fitted probes were jointly locked before test scoring. The frequency baseline uses training-only counts with Jeffreys (add-0.5) smoothing, giving the exact uniform reference prior here and baseline cross entropy $\log_2(3)=1.585$ bits.

We average log-score gains over the three views within each record, then records and encoder seeds. Because all records have three views, this equals the native record-view mean. Primary intervals use $10{,}000$ paired crossed seed/record bootstrap draws, stratified by diagnosis and retaining all views together. The split, training sample, fitted probes, and calibration are conditioned upon.

\begin{table}[htbp]
\centering\color{revisiongreen}
\normalsize
\setlength{\tabcolsep}{6pt}
\caption{NMT LPI in bits per provenance-labelled record view. Differences subtract task-only from reconstruction. Per-seed intervals resample paired records conditional on that fitted encoder pair; the mean-row interval additionally resamples encoder seeds and is the primary between-objective interval. The pooled task-only and reconstruction intervals are $[1.092,1.252]$ and $[1.161,1.369]$, respectively.}
\label{tab:post-nmt-lpi}
\begin{tabular*}{\linewidth}{@{\extracolsep{\fill}}lrrrr@{}}
\toprule
Encoder seed & Task-only LPI & Reconstruction LPI & Difference & $95\%$ interval \\
\midrule
13 & $1.165$ & $1.310$ & $+0.145$ & $[0.068,0.222]$ \\
23 & $1.125$ & $1.332$ & $+0.207$ & $[0.119,0.302]$ \\
31 & $1.190$ & $1.328$ & $+0.137$ & $[0.056,0.229]$ \\
47 & $1.226$ & $1.127$ & $-0.099$ & $[-0.168,-0.032]$ \\
\midrule
Four-seed mean & $1.177$ & $1.274$ & $+0.098$ & $[-0.041,0.212]$ \\
\bottomrule
\end{tabular*}
\end{table}

Both objectives retain substantial linearly accessible reference provenance. The reconstruction-minus-task difference is positive on average but seed-dependent, with its crossed interval including zero (Table~\ref{tab:post-nmt-lpi}). A secondary interval conditional on the four fixed encoders, $[0.058,0.140]$, is narrower but cannot replace the crossed interval for objective claims. The negative seed-$47$ contrast is retained. Thus, this analysis supports accessibility in both objectives, not a seed-invariant reconstruction increase. LPI quantifies log-score gain of the specified fitted linear probe, not exact mutual information or task-head use. Together with the five-seed fixed-head audit, these results establish why an objective label or an accessibility ranking cannot substitute for behavioral verification.
}

\FloatBarrier
\section{Regime-level NMT results for both objectives}
\label{app:nmt-regimes}

Tables~\ref{tab:nmt-regime-task-only} and~\ref{tab:nmt-regime-recon} give the full regime-level AUROC and flip-rate summaries behind the NMT columns of Table~\ref{tab:audit-outcomes}, for task-only and task-plus-reconstruction encoders, respectively. Entries are five-seed means over the three shortcut-reference rotations of weighted held-out AUROC on the same 185 records under the aligned (fitting association preserved), independent, and reversed regimes. The flip rate is the class-balanced fraction of records whose thresholded prediction differs across their three reference views. Random rows average the ten rank-two, energy-matched Haar subspaces of each encoder.

\begin{table}[htbp]
\centering\small
\setlength{\tabcolsep}{5pt}
\caption{Task-only encoders, five-seed means. The exposed head gains $0.056\,[0.039,0.074]$ over the control under the aligned regime and loses $0.037\,[0.026,0.049]$ under independence; targeted removal rescues reversed AUROC by $0.139\,[0.121,0.158]$ relative to the full representation.}
\label{tab:nmt-regime-task-only}
\begin{tabular*}{\linewidth}{@{\extracolsep{\fill}}llcccc@{}}
\toprule
Head & Representation & Aligned & Independent & Reversed & Flip rate \\
\midrule
Control & full & $0.769$ & $0.769$ & $0.769$ & $9.8\%$ \\
Control & targeted rank-2 removed & $0.770$ & $0.770$ & $0.770$ & $9.9\%$ \\
Exposed & full & $0.825$ & $0.732$ & $0.628$ & $38.5\%$ \\
Exposed & random rank-2 removed & $0.821$ & $0.729$ & $0.626$ & $38.0\%$ \\
Exposed & targeted rank-2 removed & $0.766$ & $0.766$ & $0.766$ & $11.1\%$ \\
\bottomrule
\end{tabular*}
\end{table}

\begin{table}[htbp]
\centering\small
\setlength{\tabcolsep}{5pt}
\caption{Task-plus-reconstruction encoders, five-seed means. The exposed head gains $0.062\,[0.043,0.083]$ over the control under the aligned regime and loses $0.042\,[0.030,0.055]$ under independence; targeted removal rescues reversed AUROC by $0.159\,[0.137,0.183]$ relative to the full representation.}
\label{tab:nmt-regime-recon}
\begin{tabular*}{\linewidth}{@{\extracolsep{\fill}}llcccc@{}}
\toprule
Head & Representation & Aligned & Independent & Reversed & Flip rate \\
\midrule
Control & full & $0.774$ & $0.774$ & $0.774$ & $13.8\%$ \\
Control & targeted rank-2 removed & $0.774$ & $0.774$ & $0.774$ & $15.5\%$ \\
Exposed & full & $0.836$ & $0.732$ & $0.611$ & $44.5\%$ \\
Exposed & random rank-2 removed & $0.832$ & $0.728$ & $0.609$ & $43.8\%$ \\
Exposed & targeted rank-2 removed & $0.771$ & $0.771$ & $0.771$ & $15.3\%$ \\
\bottomrule
\end{tabular*}
\end{table}

The control head is invariant to the regime by construction, because its fitting weights make reference independent of diagnosis; its small residual movement across representations reflects the projection, not the regime. After targeted removal the exposed head is numerically indistinguishable across regimes and within $0.003$--$0.004$ AUROC of the control head, which is the fixed-head recovery quantified by $G(T_{\mathbf U})$ in Table~\ref{tab:audit-outcomes}. Random removal leaves every regime value within $0.005$ of the full representation.

\FloatBarrier
\section{Second observational ECG model: CLEF-Small}
\label{app:clef}

\paragraph{Encoder and cohort.}
CLEF-Small \citep{Shu2025CLEF} is a contrastive single-lead ECG encoder pretrained on MIMIC-IV-ECG and released by its authors; we use the frozen public checkpoint, lead II at its native 500\,Hz, and its 256-dimensional embedding, with global scalar input normalization fitted on the site-0 primary training data. The cohort, official-fold-derived patient-disjoint outer folds, target tables, association regimes, head form and regularization ($C=1$), conditional-probe construction, rank-one localization, 100 equal-energy Haar controls, and bootstrap procedure are identical to the ECGFounder audit of Section~\ref{sec:ecg-audit}: 8,888 ECGs from 7,237 patients on devices CS-12 E and CS100 3.

\begin{table}[htbp]
\centering\small
\setlength{\tabcolsep}{4pt}
\caption{Association-response curves for both observational ECG models: equal means over five patient-disjoint outer folds of held-out AUROC under five fixed-marginal regimes. $\kappa$ is the fold-specific natural device--target odds ratio ($11.06$--$12.10$). Projection uses the train-localized rank-one conditional-probe normal with both heads fixed.}
\label{tab:clef-curves}
\begin{tabular*}{\linewidth}{@{\extracolsep{\fill}}lllccccc@{}}
\toprule
Model & Head & Representation & Aligned ($\kappa$) & $\sqrt\kappa$ & Indep. ($1$) & $1/\sqrt\kappa$ & Reversed ($1/\kappa$) \\
\midrule
ECGFounder & Exposed & full & $0.943$ & $0.929$ & $0.910$ & $0.891$ & $0.878$ \\
 & Exposed & projected & $0.911$ & $0.911$ & $0.911$ & $0.912$ & $0.913$ \\
 & Control & full & $0.936$ & $0.927$ & $0.914$ & $0.902$ & $0.895$ \\
 & Control & projected & $0.910$ & $0.910$ & $0.911$ & $0.912$ & $0.913$ \\
\midrule
CLEF-Small & Exposed & full & $0.903$ & $0.878$ & $0.845$ & $0.812$ & $0.790$ \\
 & Exposed & projected & $0.805$ & $0.815$ & $0.828$ & $0.840$ & $0.847$ \\
 & Control & full & $0.884$ & $0.872$ & $0.856$ & $0.841$ & $0.831$ \\
 & Control & projected & $0.782$ & $0.798$ & $0.817$ & $0.835$ & $0.847$ \\
\bottomrule
\end{tabular*}
\end{table}

\paragraph{Recover and Stress.}
Held-out conditional device AUROC, averaged over target strata and folds, is $0.897$ for CLEF-Small (ECGFounder: $0.959$), and exceeds the target/age/sex metadata comparator by $0.147$ (paired lower bound above zero). Under association stress the exposed head loses $0.113\,[0.102,0.125]$ AUROC from aligned to reversed, the control head loses $0.053\,[0.043,0.063]$, and the excess vulnerability is $G=0.0599\,[0.0549,0.0650]$, about $2.6$ times the ECGFounder value on the same cohort. The exposed head's aligned advantage over the control head is $0.019\,[0.016,0.021]$ and the control head's reversed advantage is $0.041\,[0.035,0.047]$. All primary validity gates for behavioral shortcut harm pass (Table~\ref{tab:clef-gates}).

\paragraph{Intervene.}
Rank-one projection of the train-localized conditional-probe normal, with both heads fixed, reduces $G$ to $0.0224\,[0.0202,0.0247]$, a reduction of $0.0375\,[0.0327,0.0426]$ ($63\%$), and rescues the exposed head's reversed AUROC by $0.057\,[0.046,0.068]$. Both effects exceed all 100 rank- and energy-matched random directions (descriptive rank $1/101$, not a randomization $p$-value). A refitted conditional device probe remains at $0.830$.

\paragraph{Verify: rejected.}
The control head's independent-regime AUROC changes by $-0.0386\,[-0.0443,-0.0334]$, so the lower bound lies far below the predeclared $-0.01$ margin and the utility check fails. Table~\ref{tab:clef-curves} shows why: after projection both heads' curves invert, rising from $0.78$--$0.81$ under the aligned regime to $0.85$ under reversal, and the residual $G$ of $0.022$ is comparable to ECGFounder's entire original vulnerability. The rank-one direction identified by the target-residualized probe therefore removes task-relevant variance together with device-associated variance in this embedding. Under the PhysioTRACE decision rule this is a detected, replicated reliance whose proposed linear remedy is not certified; it is reported alongside the certified ECGFounder result rather than omitted.

\begin{table}[htbp]
\centering\small
\setlength{\tabcolsep}{5pt}
\caption{CLEF-Small validity and verification gates. Estimates are point values from the corresponding bootstrap contrasts or probe summaries.}
\label{tab:clef-gates}
\begin{tabular*}{\linewidth}{@{\extracolsep{\fill}}llcc@{}}
\toprule
Gate & Criterion & Estimate & Result \\
\midrule
Primary excess gap $G$ & $\ge 0.01$ and 95\% lower bound $>0$ & $0.0599$ & pass \\
Control reversed advantage & $\ge 0.01$ and 95\% lower bound $>0$ & $0.0413$ & pass \\
Exposed aligned noninferiority & 95\% lower bound $\ge -0.01$ & $0.0186$ & pass \\
Conditional device recoverability & 95\% lower bound $>0.5$ & $0.897$ & pass \\
Embedding probe above coarse metadata & paired 95\% lower bound $>0$ & $0.147$ & pass \\
Targeted projection reduces $G$ & 95\% lower bound $>0$ & $0.0375$ & pass \\
Targeted projection reversed rescue & $\ge 0.01$ and 95\% lower bound $>0$ & $0.0570$ & pass \\
Targeted projection exceeds random controls & exceeds all 100 (descriptive rank) & $1/101$ & pass \\
Independent-regime utility margin & 95\% lower bound $\ge -0.01$ & $-0.0386$ & \textbf{fail} \\
\bottomrule
\end{tabular*}
\end{table}

\paragraph{Sensitivities.}
On the common 1990--1999 calendar support CLEF-Small retains $G=0.0501\,[0.0455,0.0553]$, and on human-validated ECGs $G=0.0063\,[0.0037,0.0089]$; both are positive; as for ECGFounder, these are sensitivity analyses on smaller cells rather than primary results, and the device factor continues to index era, workflow, and case mix. Against the age/sex/year metadata head ($G=0.061$ on the full cohort), the CLEF-Small excess is $-0.002\,[-0.010,0.007]$ on the full cohort and $0.040\,[0.035,0.047]$ on the common calendar support; for ECGFounder the corresponding values are $-0.038\,[-0.045,-0.031]$ and $0.004\,[0.001,0.007]$. Intervals use 1,000 within-fold patient-cluster draws and condition on the frozen embeddings, folds, heads, and localization.

\begingroup
\color{revisiongreen}
\FloatBarrier
\section{Scope, robustness, and sensitivity analyses}
\label{app:additional-review-results}
\label{app:post-support}

\FloatBarrier
\subsection{Held-out HMC controlled reference-shift evaluation}
\label{app:post-hmc}

The HMC Sleep Staging Database v1.1 contains 151 polysomnography recordings with five sleep stages \citep{AlvarezEstevez2022HMC}. For each selected 30-second epoch, we keep the recording, stage label, and active C4 electrode fixed while constructing C4-M1, C4-F4, C4-O2, and C4-CAR3. If $f=F4-M1$, $c=C4-M1$, and $o=O2-M1$, these views are $c$, $c-f$, $c-o$, and $c-(f+c+o)/3$. Thus the M1 term cancels in the derived differences without estimating an unavailable electrode. CAR3 denotes this sparse three-channel average, not a full-head reference.

The frozen audit uses the fixed recording-disjoint 106/23/22 train/validation/test split and 20 deterministic epochs per recording. Final train and test supports contain 8,480 and 1,760 reference views, respectively. Five standardized, class-balanced linear-probe fits use seeds 7, 13, 23, 42, and 52; each model's checkpoint, selected epochs, and embeddings remain fixed. Probes fit on training data and evaluate on the held-out test recordings. The displayed standard deviations quantify evaluation-fit variability, not uncertainty over independent encoder trainings or recording populations.

\begin{table*}[htbp]
\centering\color{revisiongreen}
\small
\setlength{\tabcolsep}{4pt}
\caption{\sloppy Final HMC held-out-test reference audit, with mean and sample standard deviation across five evaluation fits per fixed checkpoint. Task balanced-accuracy (BA) chance is 0.20; reference groups are balanced, so probe accuracy and BA coincide, with chance 0.25.}
\label{tab:app-hmc-reference}
\begin{tabular*}{\linewidth}{@{\extracolsep{\fill}}lccccc@{}}
\toprule
Frozen encoder & \makecell{Same-reference\\BA} & \makecell{Shifted\\BA} & \makecell{Worst-reference\\BA} & \makecell{Shift\\loss} & \makecell{Reference-probe\\BA} \\
\midrule
BIOT & $0.560\pm0.005$ & $0.529\pm0.006$ & $0.488\pm0.006$ & $0.031\pm0.002$ & $0.388\pm0.014$ \\
CBraMod & $0.487\pm0.010$ & $0.410\pm0.019$ & $0.300\pm0.046$ & $0.077\pm0.014$ & $0.544\pm0.007$ \\
LaBraM & $0.456\pm0.011$ & $0.439\pm0.016$ & $0.395\pm0.033$ & $0.017\pm0.008$ & $0.335\pm0.009$ \\
\bottomrule
\end{tabular*}
\end{table*}

In Table~\ref{tab:app-hmc-reference}, BIOT has the strongest shifted and worst-reference utility, while CBraMod has the greatest reference accessibility and weakest transfer. LaBraM has the lowest probe but intermediate transfer, so accessibility and robustness are not interchangeable. This is external-model screening across architectures and pretraining histories, not an isolated causal objective comparison. The fixed single-channel input adapters also bound model-ranking interpretations.

\FloatBarrier
\subsection{ECGFounder: association specificity and confounding sensitivities}
\label{app:post-ecg-sensitivity}

The primary device audit uses 8,888 ECGs from 7,237 patients for the two predominant site-0 devices, CS-12 E and CS100 3. Five patient-disjoint outer folds are formed as $(\mathrm{strat\_fold}-1)\bmod5$. Both fixed task heads share an outer-training StandardScaler and use $C=1$, without hyperparameter search. The fixed-marginal target tables use official PTB-XL folds 1--8 intersected with outer train; outer-test data do not define their own target marginals or association strength. The natural device--target odds ratios in these design sets range from 11.06 to 12.10.

All four device-by-target cells are present. The primary design has at least 83 ECGs per held-out cell; worst-fold reversed effective sample size is 756 ECGs or 543 patients, and the largest mean-one weight is 5.31. The same patient, including any ECGs from both devices, stays in one outer fold and bootstrap cluster. These checks support the within-observed-support stress construction.

Conditional device AUROC is $0.959\,[0.953,0.964]$ for ECGFounder, compared with $0.750\,[0.732,0.767]$ for target, age, and sex, and $0.839\,[0.826,0.853]$ after adding recording year. These are conditional accessibility checks, not estimates of an isolated hardware signal.

\begin{table*}[htbp]
\centering\color{revisiongreen}
\small
\setlength{\tabcolsep}{5pt}
\caption{\sloppy ECGFounder sensitivity analyses. The common-calendar cohort contains 7,324 ECGs from 5,957 patients, with minimum held-out cell support 56; the human-validated subset contains 3,147 ECGs and a minimum cell of seven. Both fall below the primary minimum-cell gate of 75 and are reported as sensitivity analyses. The weak-association pair is CS-12/AT-60 3 at site 2, with natural training odds ratio 1.04.}
\label{tab:post-ecg-sensitivity}
\begin{tabular}{>{\raggedright\arraybackslash}p{3.6cm}>{\raggedright\arraybackslash}p{3.8cm}>{\raggedright\arraybackslash}p{5.5cm}}
\toprule
Sensitivity & Result (95\% interval) & Interpretation \\
\midrule
Full-cohort metadata comparator & Metadata age/sex/year $G=0.0614\,[0.0555,0.0675]$ & Calendar and workflow are material confounds; the primary embedding result is not a hardware effect beyond these variables. \\
Calendar-matched cohort (1990--1999) & ECGFounder $G=0.0136\,[0.0116,0.0156]$; paired excess over metadata $=0.0040\,[0.0009,0.0071]$ & A coarse linear calendar comparator does not reproduce the entire embedding effect on this shared support; nonlinear era/workflow confounding remains possible. \\
Human-validated ECGs & $G=0.0022\,[0.0012,0.0032]$ & The direction is positive but below the primary 0.01 meaningful-effect gate; sparse cells make this a sensitivity check. \\
Weak-association site-2 device pair & $G=0.000074$ $[-0.000005,0.000158]$ & The interval lies within $\pm0.01$ under fold-matched stress, consistent with association specificity rather than a generic reweighting effect. \\
\bottomrule
\end{tabular}
\end{table*}

The sensitivity analyses in Table~\ref{tab:post-ecg-sensitivity} refine the interpretation of the positive primary result: the audited channel is device-associated acquisition, era, and workflow provenance, rather than hardware causality. The reversed association is a controlled within-support stress, not a claim that this deployment shift occurred naturally.

With fixed heads, the primary targeted projection reduces $G$ from $0.0233$ to $0.0021\,[0.0007,0.0034]$ and improves natural-head reversed AUROC by $0.0342\,[0.0287,0.0397]$. The independent-regime utility change for the control head is $-0.0036\,[-0.0059,-0.0012]$, within the predeclared $-0.01$ margin, not exactly zero. The refitted conditional probe remains accessible at $0.589\,[0.575,0.603]$. Targeted suppression outperforms all 100 rank- and energy-matched controls; its descriptive specificity rank of $1/101$ is not an exchangeability-based randomization $p$-value.

Primary behavioral and intervention intervals use 2,000 shared within-fold patient-cluster draws; sensitivity intervals use 1,000. Behavioral bootstrap weights restore the fixed target-cell masses, while conditional-probe AUROC is computed within each target stratum using patient-bootstrap multiplicities. All intervals condition on the existing checkpoint, split, heads, target tables, localization, and random directions.

\FloatBarrier
\subsection{ERP architecture robustness}

The ERP evaluation was repeated with EEGNetSmall and ShallowConvNet for three focal objectives (Figure~\ref{fig:app-erp-encoder-robustness}; Table~\ref{tab:app-erp-eeg-encoder-check}). Under both additional backbones, masked reconstruction has the weakest shifted utility and highest domain-probe accuracy among the compared objectives. This is an architecture check, not a backbone-specific search or a causal identification of the reconstruction term. These runs use the evaluation-fit protocol of Appendix~\ref{app:post-broad} and are distinct from the calibrated LPI analysis.

\begin{figure*}[htbp]
\centering\color{revisiongreen}
\includegraphics[width=\textwidth]{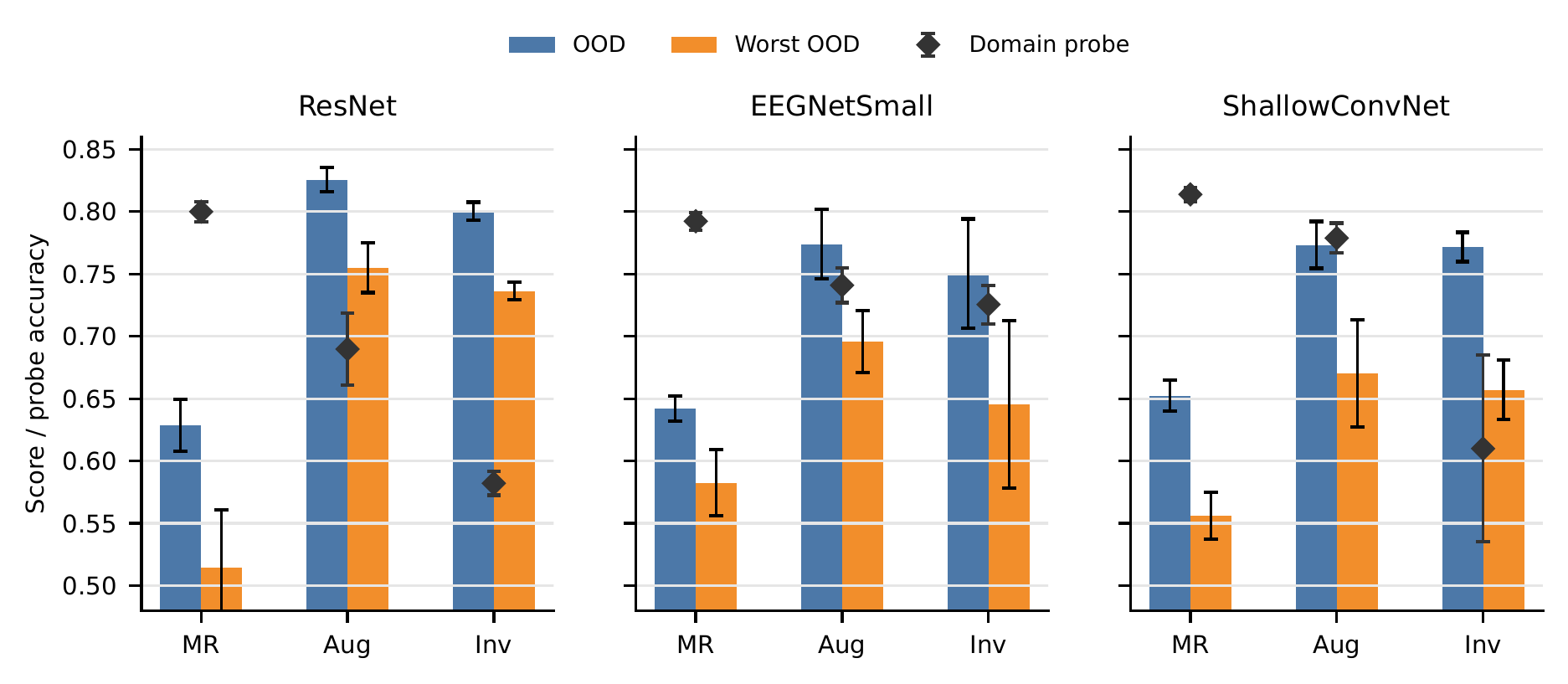}
\caption{\sloppy ERP architecture check across ResNet, EEGNetSmall, and ShallowConvNet. Bars show OOD and worst-OOD utility; black diamonds show ordinary domain-probe accuracy, not LPI.}
\label{fig:app-erp-encoder-robustness}
\label{fig:erp-encoder-robustness}
\end{figure*}

\begin{table*}[htbp]
\centering\color{revisiongreen}
\small
\setlength{\tabcolsep}{4pt}
\caption{\sloppy ERP architecture robustness: mean and sample standard deviation over five seeds. Domain probe denotes ordinary classification accuracy.}
\label{tab:app-erp-eeg-encoder-check}
\begin{tabular*}{\linewidth}{@{\extracolsep{\fill}}llccc@{}}
\toprule
Encoder & Variant & OOD AUROC & Worst OOD AUROC & Domain probe \\
\midrule
EEGNetSmall & Masked reconstruction & $0.642\pm0.010$ & $0.583\pm0.026$ & $0.792\pm0.007$ \\
EEGNetSmall & Augmentation only & $0.774\pm0.028$ & $0.696\pm0.025$ & $0.741\pm0.014$ \\
EEGNetSmall & Invariance first & $0.750\pm0.044$ & $0.645\pm0.067$ & $0.726\pm0.015$ \\
ShallowConvNet & Masked reconstruction & $0.652\pm0.012$ & $0.556\pm0.019$ & $0.814\pm0.006$ \\
ShallowConvNet & Augmentation only & $0.773\pm0.019$ & $0.670\pm0.043$ & $0.779\pm0.012$ \\
ShallowConvNet & Invariance first & $0.772\pm0.012$ & $0.657\pm0.024$ & $0.610\pm0.075$ \\
\bottomrule
\end{tabular*}
\end{table*}
\endgroup

\FloatBarrier
\section{Evidence units, seeds, and uncertainty}
\label{app:evidence-units}

\paragraph{Splits and evidence units.}
NMT provides average (AVG), Cz (CZREF), and linked-ear (LE) reference views of the same records, with record-disjoint splits. ERP/P300 combines amyotrophic lateral sclerosis (ALS) P300, covert GeoSpell, and overt P300 with physical-subject-disjoint splits, including subjects shared across healthy domains. PTB-XL site transfer uses official patient folds and the existing site-capped benchmark cohort. These latter two settings mix acquisition, population, interface, era, and workflow: they are realistic deployment-shift screens, not causal estimates of a single hardware factor.

\paragraph{Seeds, fits, and folds.}
Five shared-backbone ERP/PTB objectives use encoder seeds 7, 13, 23, 42, and 52. LaBraM, BIOT, CBraMod, ECGFounder, MERL, and CLEF-Small each use one released checkpoint; repeated evaluation fits do not count as independent pretraining runs. The focused NMT audit uses a different, explicitly matched EEGNetSmall training recipe, $d=128$, and seeds 7, 13, 23, 31, and 47. The observational ECG audits use five patient-disjoint outer folds of one checkpoint each. These grids are not pooled as interchangeable training replicates.

\paragraph{Uncertainty.}
Uncertainty follows the independent unit of each experiment. Mechanism estimates use 2,000 crossed encoder-seed and label-stratified record bootstrap draws for NMT, and within-fold patient-cluster draws for ECG. Calibrated LPI uses 10,000 draws, crossing encoder seeds with physical clusters for trained variants and cluster-only uncertainty for fixed checkpoints. ERP has 15,312 test trials from four physical subjects; we therefore report leave-one-subject-out sensitivity alongside its cluster intervals. PTB LPI uses 1,487 ECGs from 1,324 patients. All views or repeated observations of an identity move together. All fitting choices are locked before test scoring; intervals condition on the split and selected probes, not a newly sampled training corpus.

\paragraph{Audit outcomes with intervals.}
Table~\ref{tab:audit-outcomes-full} gives the 95\% bootstrap intervals behind the point estimates of Table~\ref{tab:audit-outcomes}, using the resampling units above.

\providecommand{\ci}[1]{\\[-1pt]{\footnotesize[#1]}}
\begin{table}[htbp]
\centering\small
\setlength{\tabcolsep}{5pt}
\renewcommand{\arraystretch}{1.3}
\caption{Table~\ref{tab:audit-outcomes} with 95\% bootstrap intervals in brackets: crossed encoder-seed and record resampling for NMT, within-fold patient clusters for ECG. Recover reports reference LPI in bits per record view for NMT (baseline $\log_2 3=1.585$; encoder seeds 13, 23, 31, and 47, whose embeddings were retained; Appendix~\ref{app:post-nmt}) and device AUROC within diagnosis strata for ECG; it enters no verdict. The weak-association ECGFounder audit stops at Stress with $G=0.00007$ [$-$0.00001, 0.00016].}
\label{tab:audit-outcomes-full}
\vspace{3pt}
\begin{tabular*}{\linewidth}{@{\extracolsep{\fill}}lcccc@{}}
\toprule
& \multicolumn{2}{c}{Paired EEG (NMT)} & \multicolumn{2}{c}{Observational ECG (PTB-XL)} \\
\cmidrule(lr){2-3}\cmidrule(l){4-5}
& Task-only & Task + recon. & ECGFounder & CLEF-Small \\
\midrule
Recover & \makecell{1.18\ci{1.09, 1.25}} & \makecell{1.27\ci{1.16, 1.37}} & \makecell{0.959\ci{0.953, 0.964}} & \makecell{0.897\ci{0.887, 0.906}}  \\
Excess vulnerability $G$ & \makecell{0.197\ci{0.173, 0.221}} & \makecell{0.225\ci{0.193, 0.256}} & \makecell{0.023\ci{0.021, 0.026}} & \makecell{0.060\ci{0.055, 0.065}}  \\
$G$ after suppression & \makecell{$-$0.001\ci{$-$0.003, 0.002}} & \makecell{0.001\ci{$-$0.002, 0.004}} & \makecell{0.002\ci{0.001, 0.003}} & \makecell{0.022\ci{0.020, 0.025}}  \\
Beats random controls & 10/10 per seed & 10/10 per seed & 100/100 & 100/100  \\
Utility change & \makecell{0.000\ci{$-$0.004, 0.004}} & \makecell{0.001\ci{$-$0.001, 0.005}} & \makecell{$-$0.004\ci{$-$0.006, $-$0.001}} & \makecell{$-$0.039\ci{$-$0.044, $-$0.033}}  \\
\quad tolerated loss $\delta$ & 0.02 & 0.02 & 0.01 & 0.01  \\
\midrule
Reliance detected & yes & yes & yes & yes  \\
Remedy certified & yes & yes & yes & \textbf{no}  \\
\bottomrule
\end{tabular*}
\end{table}

\section{Additional tables, model coverage, compute, and assets}
\label{app:additional-materials}

\postrev{This section gives benchmark specifications, the literature inventory, and source-to-target results. Classification-probe values use their own fitting and observation units; they are not the calibrated LPI values of Appendix~\ref{app:post-lpi}. The paired NMT reliance audit is reported in Appendix~\ref{app:post-nmt}.}

\FloatBarrier
\subsection{Evaluation framing and provenance tables}

Tables~\ref{tab:invariance-shift-types}--\ref{tab:minimal-provenance-schema} give factor roles, shift-family coverage, and a minimal provenance schema.

\begin{table*}[htbp]
\centering
\small
\caption{\sloppy Factor roles used by PhysioTRACE. Near nuisances support matched-view robustness
tests, label-informative variables require an explicit and stable deployment rationale, and confounded
variables require conditional analysis rather than indiscriminate erasure.}
\label{tab:invariance-shift-types}
\begingroup
\setlength{\tabcolsep}{3pt}
\renewcommand{\arraystretch}{1.10}
\begin{tabular}{>{\raggedright\arraybackslash}p{2.2cm}>{\raggedright\arraybackslash}p{4.1cm}>{\raggedright\arraybackslash}p{3.4cm}>{\raggedright\arraybackslash}p{3.4cm}}
\toprule
Factor role & Examples & Desired representation behavior & Evaluation implication \\
\midrule
Near nuisance
&
Valid re-referencing, geometry-aware channel subsets, benign resampling, or bounded acquisition
changes known to preserve physiology and target semantics
&
Preserve task-relevant physiology while limiting unstable acquisition identity
&
Use matched views when available; require shifted and worst-group utility, fixed-head verification,
and a utility margin before suppression claims. \\
\midrule
Label-informative
&
An acquisition or protocol variable that is part of the intended measurement and has
domain-supported, stable deployment meaning
&
Model the factor explicitly rather than imposing invariance
&
State the operational rationale and audit whether its predictive role remains stable across
supported deployment groups. \\
\midrule
Confounded
&
Site, device, protocol, or workflow differences that also track population, label prevalence,
or clinical practice without matched counterfactual views
&
Separate accessibility from decision reliance and avoid biological or hardware-causal interpretation
&
Use identity-disjoint splits, composition reporting, target-conditional probes, fixed-marginal stress,
and control heads fitted without the association. \\
\bottomrule
\end{tabular}
\endgroup
\end{table*}

\begin{table*}[htbp]
\centering
\small
\caption{\sloppy Shift families covered by the provenance-aware protocol. The table distinguishes
families instantiated in the main experiments from provenance requirements needed for auditable
future evaluation.}
\label{tab:shift-family-coverage}
\begingroup
\setlength{\tabcolsep}{3pt}
\renewcommand{\arraystretch}{1.10}
\begin{tabular}{>{\raggedright\arraybackslash}p{2.4cm}>{\raggedright\arraybackslash}p{3.5cm}>{\raggedright\arraybackslash}p{3.5cm}>{\raggedright\arraybackslash}p{3.7cm}}
\toprule
Shift family & Controlled stress test & Natural held-out group & Instantiation and interpretation here \\
\midrule
Reference and montage
&
Valid re-referencing, reference mixing, channel dropping, subset sampling, or geometry-aware remapping
&
Held-out reference family, montage family, channel set, or reference pipeline
&
NMT is the controlled anchor: the task is fixed and the same records are evaluated under AVG, CZREF,
and LE reference views. \\
\midrule
Device, site, and protocol
&
Device-aware perturbations, acquisition-pipeline variants, or protocol-specific preprocessing changes
when physically justified
&
Held-out amplifier, device model, collection site, task interface, or clinical workflow
&
ERP/P300 tests real-domain EEG transfer across interface and population differences; PTB-XL tests
clinical ECG site transfer. Both require confound-aware interpretation. \\
\midrule
Sampling, filters, and timing
&
Resampling with anti-aliasing, bounded filter-chain perturbations, mild line-noise variation, or
clock-drift-like jitter
&
Held-out sampling rate, filter chain, artifact-removal policy, or segmentation convention
&
Included in the provenance schema and leakage audit. The present experiments do not claim causal
isolation of filter or timing effects. \\
\midrule
Quality and environment
&
Realistic impedance-like noise, dropped-channel patterns, sensor-contact variation, or bounded
motion/artifact perturbations
&
Held-out quality stratum, recording environment, wearable condition, or artifact regime
&
Treated as a reporting requirement here; future datasets can instantiate this family without changing
the protocol metrics. \\
\bottomrule
\end{tabular}
\endgroup
\end{table*}

\begin{table*}[htbp]
\centering
\small
\caption{\sloppy Minimal provenance schema for acquisition- and provenance-shift evaluation. The
fields are not assumed to be available in every dataset; robustness claims should state which fields
define groups and which are missing.}
\label{tab:minimal-provenance-schema}
\begingroup
\setlength{\tabcolsep}{4pt}
\renewcommand{\arraystretch}{1.10}
\begin{tabular}{>{\raggedright\arraybackslash}p{2.9cm}>{\raggedright\arraybackslash}p{10.5cm}}
\toprule
Category & Fields to record or standardize \\
\midrule
Device/amplifier
&
Manufacturer, model, amplifier, analog-to-digital converter (ADC) characteristics, firmware or acquisition software when available \\
\midrule
Reference or lead definition
&
Online reference, re-reference procedure, lead configuration, polarity conventions \\
\midrule
Montage or sensor geometry
&
Channel names, electrode positions, lead set, missing-channel policy, remapping or interpolation procedure \\
\midrule
Sampling and timing
&
Sampling rate, anti-alias settings, clock drift information, resampling procedure \\
\midrule
Filters and preprocessing
&
Hardware and software filter chain, notch settings, artifact removal, independent component analysis (ICA) or artifact subspace reconstruction (ASR) flags, normalization,
segmentation policy \\
\midrule
Quality and impedance
&
Per-channel impedance or quality indicators when available, dropped channels, sensor-contact metadata \\
\midrule
Protocol and site
&
Site identifier, protocol identifier, session metadata, collection environment, device workflow, subject or patient split unit \\
\bottomrule
\end{tabular}
\endgroup
\end{table*}

\FloatBarrier
\subsection{Frozen-encoder preprocessing details}
\label{app:frozen-encoder-details}

For EEG, we evaluate the official LaBraM-Base checkpoint \citep{Jiang2024}
without fine-tuning. NMT windows use the fixed record-level splits, 100 Hz,
2 s, 0.5-45 Hz preprocessing, and are resampled to a 400-sample,
200 Hz-compatible input before per-channel z-scoring and embedding extraction.
All AVG, CZREF, and LE reference views derived from the same underlying NMT
record are assigned to the same split, so no record appears in training
under one reference view and in evaluation under another. ERP/P300 epochs are
resampled to a 200-sample, 200 Hz-compatible input and evaluated with the same
subject-disjoint domain splits and probe seeds.

For ECG, we evaluate the public 12-lead ECGFounder checkpoint
\citep{Li2025ECGFounder} as a frozen encoder under the unchanged PTB-XL
site-shift protocol. High-resolution PTB-XL \texttt{filename\_hr} records are
read at 500 Hz, represented as standard 12-lead, 5000-sample inputs, normalized
by per-record z-scoring, and embedded using ECGFounder's penultimate
\texttt{deep\_features}. We then fit the same source-site normal-versus-abnormal
linear probes and the same multiclass site probe used for the residual-network (ResNet) rows.

\postrev{In the broad screens, frozen-encoder probes operate on windows and are fitted on train plus validation after protocol selection. The calibrated LPI analyses instead use train-only fitting, validation selection and calibration, and explicitly declared native-unit scoring with physical-cluster uncertainty (Appendix~\ref{app:post-lpi}). No frozen checkpoint is retrained across its repeated evaluation fits.}

\FloatBarrier
\subsection{Artifact contents}

Table~\ref{tab:app-artifact-contents} summarizes the PhysioTRACE artifact and the role each component plays in making the evaluation reusable. The artifact intentionally separates reusable evaluation code from saved paper-facing results: shared utilities define the split, probe, transfer-matrix, metric, and reporting behavior, while benchmark-local folders hold task-specific report code and paper-facing comma-separated values (CSV) and JavaScript Object Notation (JSON) outputs. Raw physiological datasets and external foundation-model checkpoints are not redistributed; the artifact documents the expected local data layout and official access requirements.

\postrev{The \path{additional_experiments/} package contains the paired NMT audit, the observational ECG audits, and the calibrated LPI analyses, with runners, locked selection manifests, saved predictions or probe states, bootstrap draws, and independently checked uncertainty estimates.}

\begin{table*}[htbp]
\centering
\footnotesize
\caption{\sloppy Contents of the PhysioTRACE artifact: a checklist, schema, reusable evaluation code, benchmark cards, notebooks, saved result artifacts, experiment runners, and scripts for regenerating paper outputs.}
\label{tab:app-artifact-contents}
\begingroup
\setlength{\tabcolsep}{3pt}
\renewcommand{\arraystretch}{1.12}
\begin{tabular}{@{}>{\raggedright\arraybackslash}p{4.0cm}>{\raggedright\arraybackslash}p{9.6cm}@{}}
\toprule
Artifact component & Purpose \\
\midrule
\texttt{README.md} & High-level artifact overview, repository layout, quickstart, data requirements, and expected output files. \\
\texttt{ARTIFACT.md} & Exact lightweight verification commands, data setup instructions, full-rerun entry points, and result-source map. \\
\texttt{PROTOCOL\_CHECKLIST.md} & Compact checklist for applying provenance-aware evaluation: acquisition groups, split unit, leakage controls, shift interpretation, required metrics, and saved outputs. \\
\texttt{pyproject.toml}, \texttt{requirements.txt}, \texttt{LICENSE} & Installable package metadata, Python dependency list, and software license. \\
\texttt{sharable\_modules/} & Reusable dataset registry, split preparation, preprocessing, model primitives, training helpers, frozen-probe utilities, transfer-matrix construction, pooled-target metrics, and common reporting semantics. \\
\texttt{benchmarks/} & Benchmark-specific modules layered above the shared utilities, including NMT reference-shift reporting, ERP/P300 domain-transfer reporting, PTB-XL site-shift reporting, external LaBraM/ECGFounder frozen-encoder utilities, and shared report-export mechanics. \\
\texttt{benchmarks/*/outputs/} & CSV/JSON result artifacts for the paper-facing runs: NMT ResNet and LaBraM, ERP/P300 ResNet, encoder check and LaBraM, and PTB-XL ResNet and ECGFounder. These files support table and figure regeneration without raw data. \\
\texttt{notebooks/} & Selected executable experiment notebooks for the NMT, ERP/P300, PTB-XL, LaBraM, and ECGFounder evaluations. Notebook outputs are cleared. \\
\texttt{paper\_assets/} & Scripts that regenerate paper-facing figures and appendix tables from the saved CSV/JSON artifacts. Generated files are written outside the tracked artifact source tree. \\
\texttt{benchmark\_cards/} & One card per paper-facing benchmark, documenting task, acquisition groups, split unit, metrics, known confounds, and data access requirements. \\
\path{additional_experiments/} & Installable package, runners, notebooks, tests, and saved result bundles for the paired NMT audit, the ECGFounder and CLEF-Small device audits, and the calibrated ERP/PTB and NMT LPI analyses (the ERP/PTB grid covers the LaBraM, BIOT, CBraMod, ECGFounder, and MERL checkpoints), including source hashes, selection locks, exported probe states, and bootstrap draws. \\
\path{schema/provenance_schema.json} & Machine-readable minimal provenance schema covering acquisition group, split unit, device/site/protocol fields, reference or lead definition, montage or sensor geometry, sampling, filters, preprocessing, and quality metadata. \\
\bottomrule
\end{tabular}
\endgroup
\end{table*}

\FloatBarrier
\subsection{Physiological foundation-model pretraining landscape}

\postrev{Tables~\ref{tab:app-pfm-data-coverage}--\ref{tab:app-pfm-methods} give the pretraining-landscape inventory as background, not as a ranking of the models evaluated here.} The purpose is not to exhaustively rank prior PFMs, but to make the design space concrete: recent models differ in corpus heterogeneity, channel handling, preprocessing signatures, and objective type, yet many still use reconstruction or masked prediction as a central pretraining signal. This motivates auditing whether task heads built on such encoders rely on recording provenance.

\smallskip
\noindent\textbf{Column notes.}
\textit{Multi-ch.} means multi-channel and indicates that the source corpora or model design address heterogeneous channel counts, montages, or lead availability. \textit{Multi-dev.} means multi-device and indicates explicit use of, evaluation on, or claimed robustness to multiple acquisition devices, sites, or data-collection sources. Objective type is abbreviated as R for reconstruction or masked modeling, C for contrastive or alignment, and H for hybrid.

\noindent\textbf{Signal and method abbreviations.}
PSG denotes polysomnography, EOG electrooculography, EMG electromyography, HAR human activity recognition, FFT fast Fourier transform, MSE mean squared error, ViT Vision Transformer, MAE masked autoencoder, and NT-Xent normalized temperature-scaled cross-entropy.

\noindent\textbf{Model abbreviations.}
REVE denotes Representation for EEG with Versatile Embeddings; EEGPT, a pretrained Transformer for EEG; BENDR, Bidirectional Encoder Representations from Transformers (BERT)-inspired Neural Data Representations; ECG-FM, an electrocardiography foundation model; and CREMA, a Contrastive Regularized Masked Autoencoder.

\noindent\textbf{Corpus abbreviations.}
PhysioMI/EEGMMIDB denotes the PhysioNet EEG Motor Movement/Imagery Dataset; HGD, the High-Gamma Dataset; TUH/TUEG, the Temple University Hospital EEG Corpus; SEED, the Shanghai Jiao Tong University Emotion EEG Dataset; M3CV, the Multi-subject, Multi-session, Multi-task Database for Investigation of EEG Commonality and Variability; SHHS, the Sleep Heart Health Study; CHB-MIT, the Children\textquotesingle s Hospital Boston-Massachusetts Institute of Technology Scalp EEG Database; IIIC, the ictal-interictal-injury continuum; TUAB, the TUH Abnormal EEG Corpus; TUEV, the TUH EEG Events Corpus; and MIMIC, the Medical Information Mart for Intensive Care. SHHS1 denotes SHHS cohort 1, CODE-15\% the 15\% subset of the Clinical Outcomes in Digital Electrocardiography cohort, and IKEM the Institute for Clinical and Experimental Medicine; SaMi-Trop is retained as the official cohort name.

\begin{table*}[htbp]
\centering
\footnotesize
\caption{\sloppy Physiological foundation-model pretraining setups: corpus coverage. Corpora differ widely in channels, devices, and sources, which are the provenance fields an audit needs.}
\label{tab:app-pfm-data-coverage}
\begingroup
\setlength{\tabcolsep}{2.5pt}
\renewcommand{\arraystretch}{1.12}
\begin{tabular}{@{}>{\raggedright\arraybackslash}p{1.2cm}>{\raggedright\arraybackslash}p{2.1cm}>{\raggedright\arraybackslash}p{7.2cm}cc@{}}
\toprule
Signal & Model & Pretraining corpus coverage & Multi-ch. & Multi-dev. \\
\midrule
EEG & REVE \citep{Ouahidi2025} & 92 public EEG datasets; approximately 25,000 subjects; designed for adaptation across heterogeneous EEG setups. & Yes & Yes \\
EEG & EEGPT \citep{Wang2024} & PhysioMI/EEGMMIDB, HGD, TUH EEG, SEED, and M3CV; broad EEG corpus spanning motor, clinical, emotion, and multi-session settings. & Yes & Yes \\
EEG, ECG & BIOT \citep{Yang2023} & SHHS, CHB-MIT, IIIC Seizure, TUAB, TUEV, and HAR-style biosignal or wearable sources; designed for cross-data learning in the wild. & Yes & Yes \\
EEG & BENDR \citep{Kostas2021} & TUEG/TUH EEG Corpus; large-scale EEG pretraining with a wav2vec-style contrastive objective. & No & Yes \\
EEG & CBraMod \citep{Wang2025CBraMod} & TUEG/TUH EEG Corpus; criss-cross EEG foundation-model pretraining. & No & Yes \\
EEG & LaBraM \citep{Jiang2024} & Large multi-dataset EEG pretraining corpus; discrete neural-token modeling for generic EEG representations. & Yes & Yes \\
ECG & ECG-FM \citep{McKeen2025} & PhysioNet 2021 and MIMIC-IV-ECG; 12-lead ECG pretraining with downstream clinical evaluation. & No & Yes \\
ECG & CREMA \citep{Song2025CREMA} & MIMIC-III, CODE-15\%, United Kingdom (UK) Biobank, SaMi-Trop, and IKEM ECG sources; designed for robust ECG diagnostics across clinical domains. & No & Yes \\
PSG & SynthSleepNet \citep{Lee2025SynthSleepNet} & SHHS1 polysomnography with selected EEG, EOG, ECG, and leg-EMG channels; multimodal sleep-analysis pretraining. & No & Yes \\
\bottomrule
\end{tabular}
\endgroup
\end{table*}

\begin{table*}[htbp]
\centering
\footnotesize
\caption{\sloppy Physiological foundation-model pretraining setups: preprocessing and objective design. The repeated use of reconstruction, masked prediction, or reconstruction-containing hybrid objectives means that many PFMs are trained to reproduce signal detail, including detail that can carry recording provenance.}
\label{tab:app-pfm-methods}
\begingroup
\setlength{\tabcolsep}{2.5pt}
\renewcommand{\arraystretch}{1.12}
\begin{tabular}{@{}>{\raggedright\arraybackslash}p{1.2cm}>{\raggedright\arraybackslash}p{2.0cm}>{\raggedright\arraybackslash}p{4.6cm}c>{\raggedright\arraybackslash}p{4.6cm}@{}}
\toprule
Signal & Model & Preprocessing signature & Type & Loss or objective \\
\midrule
EEG & REVE \citep{Ouahidi2025} & four-dimensional spatiotemporal positional encoding using electrode coordinates and time index; patching without requiring a fixed montage. & R & Masked autoencoding: reconstruct masked raw patches with an $L_1$ term plus a weighted global-token term. \\
EEG & EEGPT \citep{Wang2024} & Patch multichannel EEG; channel-identity embeddings and temporal rotary position encoding; masks over time and channels. & H & Alignment to a momentum target plus masked reconstruction with MSE-style losses on normalized targets. \\
EEG, ECG & BIOT \citep{Yang2023} & Resampling, per-channel amplitude normalization, fixed-length tokenization, FFT-energy features, and channel plus relative-position embeddings. & C & Contrastive alignment of sample embeddings under token and channel dropout, optimized with a similarity-matrix cross-entropy loss. \\
EEG & BENDR \citep{Kostas2021} & Scale and shift to a common range, resample to 256 Hz, map to 19 10/20 channels with missing channels zero-filled, and use 60 s segments. & C & wav2vec2-style contrastive loss on masked latents with in-sequence negatives plus an activation penalty. \\
EEG & CBraMod \citep{Wang2025CBraMod} & Convolutional feature encoder plus transformer; time- and frequency-aware patch features with patch masking. & R & Masked patch reconstruction with MSE-style losses in time/frequency feature space. \\
EEG & LaBraM \citep{Jiang2024} & Vector-quantized neural tokenizer via spectrum prediction; per-sample normalization; patches over channels and time. & R & Discrete masked-token prediction over learned neural codes. \\
ECG & ECG-FM \citep{McKeen2025} & Resample to 500 Hz, z-score normalization, non-overlapping 5 s segments, and lead-subset-aware fine-tuning. & H & Contrastive learning with ECG augmentations plus a continuous masked-signal objective. \\
ECG & CREMA \citep{Song2025CREMA} & 250 Hz resampling and fixed 10 s 12-lead ECG segments. & H & Contrastive-regularized masked autoencoding: reconstruction plus a contrastive regularizer. \\
PSG & SynthSleepNet \citep{Lee2025SynthSleepNet} & SHHS polysomnography with per-modality bandpass filtering; frozen modality encoders with a ViT-style multimodal MAE. & H & Hybrid objective combining MSE reconstruction on masked vectors with an NT-Xent contrastive term between signal and fusion representations. \\
\bottomrule
\end{tabular}
\endgroup
\end{table*}

\FloatBarrier
\subsection{Evaluation settings}

Table~\ref{tab:app-benchmark-settings} lists the provenance groups, utility metric, and seeds of each broad screen.

\begin{table*}[htbp]
\centering
\footnotesize
\caption{\sloppy \postrev{Broad-screen settings. Seeds index independent encoder trainings for trainable objectives and repeated evaluation fits for a fixed pretrained checkpoint in frozen-model rows; they are not interchangeable sources of uncertainty. The HMC evaluation (Appendix~\ref{app:post-hmc}), the paired NMT audit (Appendix~\ref{app:post-nmt}), the observational ECG audits (Appendices~\ref{app:post-ecg-sensitivity} and~\ref{app:clef}), and the calibrated LPI analyses (Appendix~\ref{app:post-lpi}) are specified separately.}}
\label{tab:app-benchmark-settings}
\begingroup
\setlength{\tabcolsep}{2.5pt}
\renewcommand{\arraystretch}{1.10}
\begin{tabular}{@{}>{\raggedright\arraybackslash}p{2.9cm}>{\raggedright\arraybackslash}p{2.6cm}>{\raggedright\arraybackslash}p{1.6cm}>{\raggedright\arraybackslash}p{2.4cm}>{\raggedright\arraybackslash}p{3.5cm}@{}}
\toprule
Evaluation & Groups & Metric & Seeds & Appendix role \\
\midrule
NMT clinical EEG reference shift & AVG, CZREF, LE & balanced accuracy & 7, 13, 23, 42, 52 & controlled EEG reference-shift evaluation \\
ERP/P300 domain transfer & ALS P300, covert GeoSpell, overt P300 & AUROC & 7, 13, 23, 42, 52 & primary real-domain EEG transfer evaluation \\
ERP encoder check & ALS P300, covert GeoSpell, overt P300 & AUROC & 7, 13, 23, 42, 52 & architecture robustness check \\
NMT frozen EEG encoders (LaBraM, BIOT, CBraMod) & AVG, CZREF, LE & balanced accuracy & 7, 13, 23, 42, 52 & external frozen EEG foundation-model audits \\
ERP frozen EEG encoders (LaBraM, BIOT, CBraMod) & ALS P300, covert GeoSpell, overt P300 & AUROC & 7, 13, 23, 42, 52 & external frozen EEG foundation-model audits \\
PTB-XL cross-site transfer & site 0, site 1, site 2 & AUROC & 7, 13, 23, 42, 52 & cross-modality clinical site-shift evaluation \\
PTB-XL frozen ECG encoders (ECGFounder, MERL) & site 0, site 1, site 2 & AUROC & 7, 13, 23, 42, 52 & external frozen 12-lead ECG foundation-model audits \\
\bottomrule
\end{tabular}
\endgroup
\end{table*}

\paragraph{Dataset access.}
NMT was used as an open clinical EEG normal-versus-abnormal dataset
\citep{Khan2022NMT}. ERP/P300 domains use the ALS P300 dataset
\citep{Riccio2013ALS} and the covert GeoSpell/overt P300 spelling datasets
\citep{Arico2014P300Jitter,Aloise2012Geospell}. PTB-XL was accessed from
PhysioNet version 1.0.1 \citep{PTBXLPhysioNet2020,Goldberger2000PhysioNet} and
is described in the peer-reviewed PTB-XL dataset paper \citep{Wagner2020}.

\FloatBarrier
\subsection{Paired seed-wise contrasts}

\postrev{Table~\ref{tab:app-paired-deltas} reports within-benchmark paired seed-wise differences, rather than subtracting independently rounded means. These descriptive checks neither isolate a single causal objective component nor establish universal ordering. Direct exposure baselines are not included in this table because their mean-target and worst-target estimands are not the source-held-out estimands of the objective rows.}

\begin{table*}[htbp]
\centering
\footnotesize
\caption{\sloppy \postrev{Paired seed-wise deltas, reported as mean and sample standard deviation. Positive shifted and worst-group deltas favor the method before the minus sign; negative gap deltas indicate a smaller ID--OOD gap. Probe deltas compare classification-probe accuracies, not calibrated LPI or downstream reliance.}}
\label{tab:app-paired-deltas}
\begingroup
\setlength{\tabcolsep}{2.5pt}
\renewcommand{\arraystretch}{1.10}
\begin{tabular*}{\linewidth}{@{\extracolsep{\fill}}>{\raggedright\arraybackslash}p{1.9cm}>{\raggedright\arraybackslash}p{3.3cm}cccc@{}}
\toprule
Evaluation & Contrast & Shifted/OOD $\Delta$ & Worst $\Delta$ & Gap $\Delta$ & Probe $\Delta$ \\
\midrule
NMT reference & Contrastive $-$ canonical reconstruction + invariance & 0.054 $\pm$ 0.019 & 0.071 $\pm$ 0.030 & -0.046 $\pm$ 0.009 & -0.221 $\pm$ 0.086 \\
NMT reference & Supervised contrastive $-$ canonical reconstruction + invariance & 0.056 $\pm$ 0.030 & 0.090 $\pm$ 0.038 & -0.045 $\pm$ 0.015 & -0.191 $\pm$ 0.101 \\
ERP/P300 domains & Augmentation $-$ masked reconstruction & 0.197 $\pm$ 0.028 & 0.241 $\pm$ 0.046 & -0.063 $\pm$ 0.038 & -0.110 $\pm$ 0.035 \\
ERP/P300 domains & Invariance $-$ masked reconstruction & 0.172 $\pm$ 0.021 & 0.222 $\pm$ 0.047 & -0.064 $\pm$ 0.038 & -0.218 $\pm$ 0.017 \\
PTB-XL sites & Task only $-$ masked reconstruction & 0.188 $\pm$ 0.050 & 0.217 $\pm$ 0.048 & -0.010 $\pm$ 0.006 & 0.038 $\pm$ 0.067 \\
PTB-XL sites & Augmentation $-$ masked reconstruction & 0.188 $\pm$ 0.049 & 0.216 $\pm$ 0.050 & -0.011 $\pm$ 0.008 & 0.010 $\pm$ 0.065 \\
\bottomrule
\end{tabular*}
\endgroup
\end{table*}

\FloatBarrier
\subsection{Source-to-target transfer heatmaps}

\begin{figure*}[htbp]
\centering
\includegraphics[width=\textwidth]{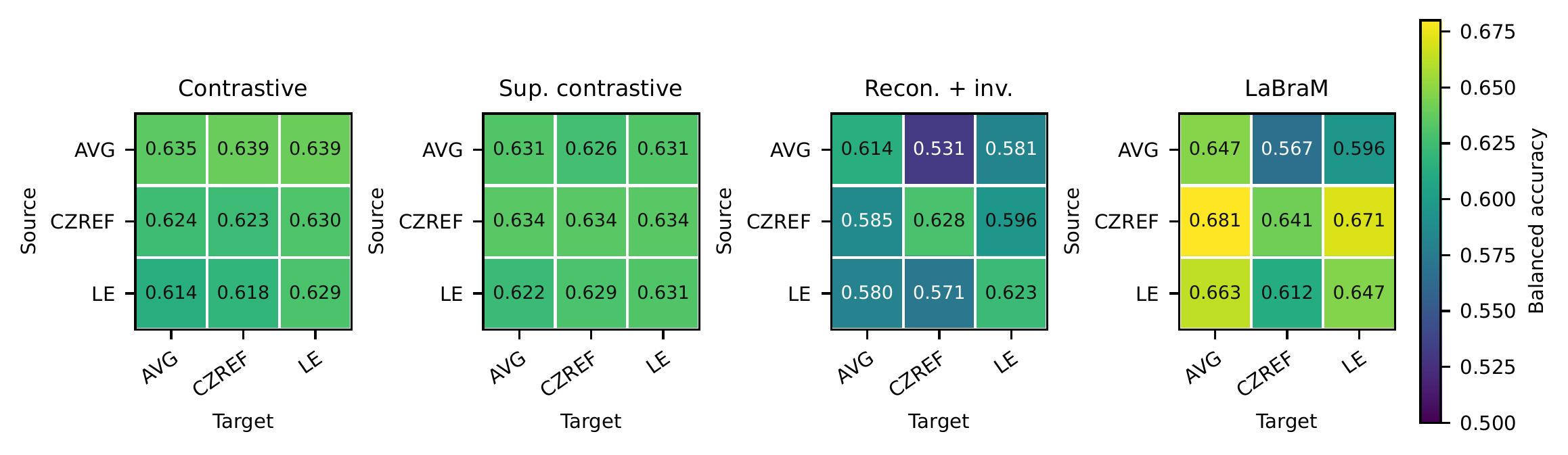}
\caption{\sloppy NMT source-reference to target-reference heatmaps for the key frozen objectives and the external frozen LaBraM audit. Entries are balanced accuracy from the five-seed targeted weighted ResNet or linear-probe runs. In the figure labels, Sup., recon., and inv. denote supervised, reconstruction, and invariance.}
\label{fig:app-nmt-transfer-heatmaps}
\end{figure*}

\begin{figure*}[htbp]
\centering
\includegraphics[width=\textwidth]{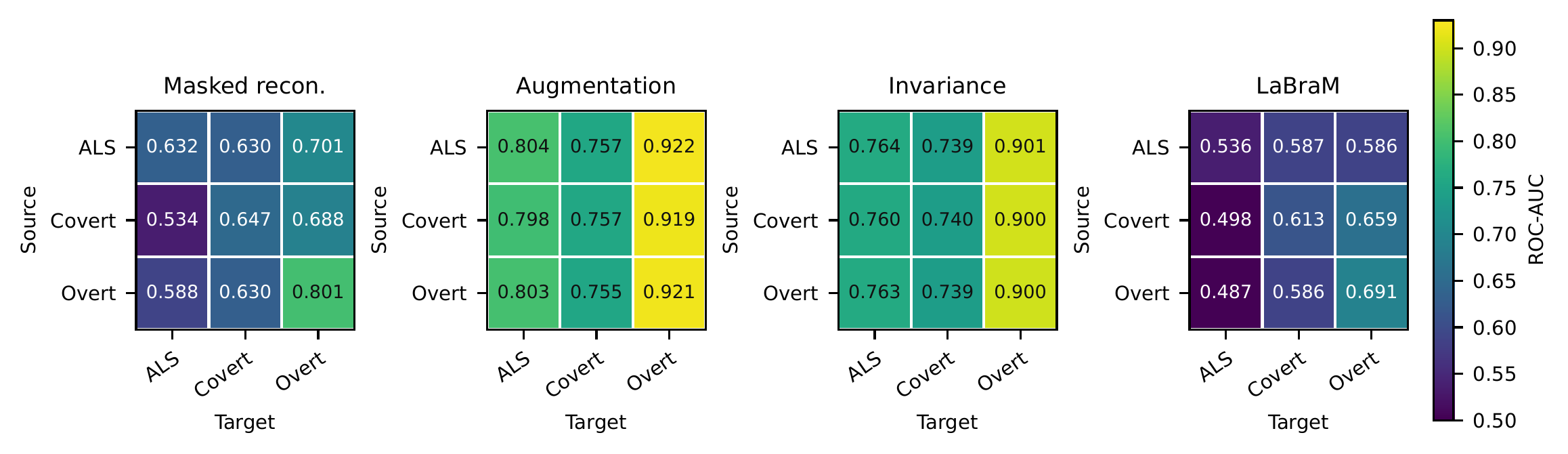}
\caption{\sloppy ERP/P300 source-domain to target-domain heatmaps for the key ResNet objectives and the secondary external frozen LaBraM audit. Entries are AUROC over ALS P300, covert GeoSpell, and overt P300 domains.}
\label{fig:app-erp-transfer-heatmaps}
\end{figure*}

\begin{figure*}[htbp]
\centering
\includegraphics[width=\textwidth]{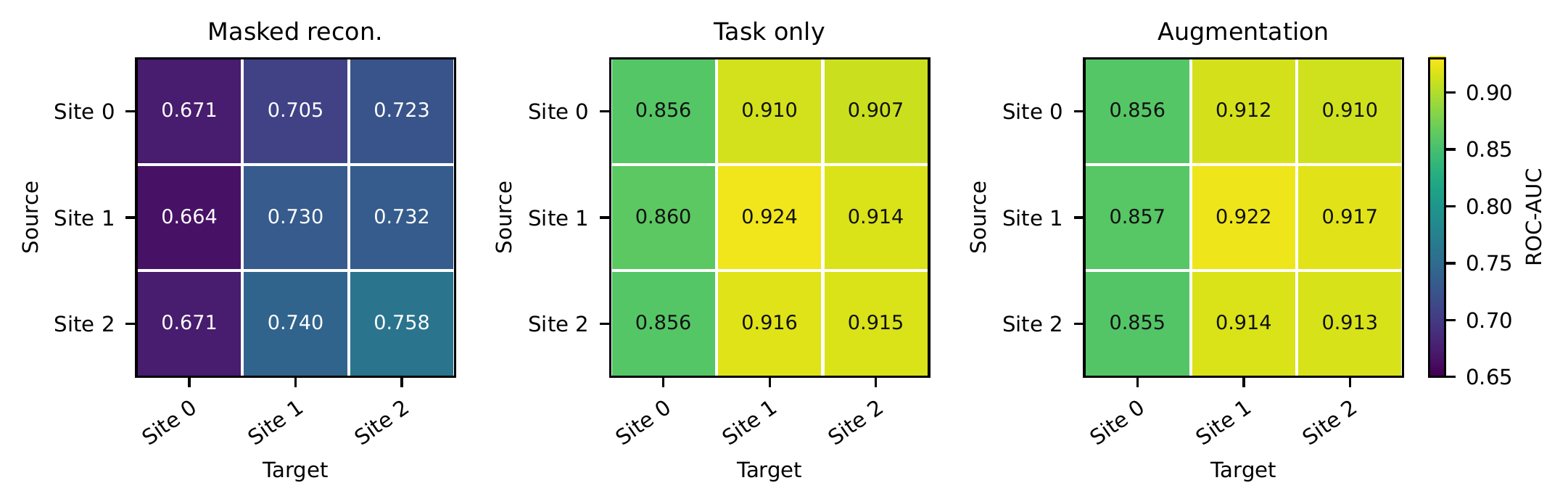}
\caption{\sloppy PTB-XL source-site to target-site heatmaps for the key ResNet objectives. Entries are AUROC over the three retained site groups.}
\label{fig:app-ptbxl-transfer-heatmaps}
\end{figure*}

Figures~\ref{fig:app-nmt-transfer-heatmaps}--\ref{fig:app-ptbxl-transfer-heatmaps} visualize the same source-to-target transfer matrices reported numerically below. Rows denote the source acquisition group used to fit the frozen-representation classifier, and columns denote the target acquisition group. The diagonal cells are in-domain evaluations and the off-diagonal cells are shifted evaluations.

\FloatBarrier
\Needspace{8\baselineskip}
\subsection{Detailed target and transfer tables}

\postrev{The following result tables expose target-specific and source-to-target structure. Table~\ref{tab:app-nmt-targets} reports direct and pooled-target NMT scores; these are not source-held-out transfer, and the mixed-reference row is an exposure baseline that sees every reference in training. Tables~\ref{tab:app-nmt-transfer}--\ref{tab:app-ptbxl-transfer} contain genuine source-specific matrices: diagonal entries are source-matched and off-diagonal entries are held-out-source-group evaluations.}

\begin{table*}[htbp]
\centering
\small
\caption{\sloppy \postrev{NMT target-reference summary. Entries are balanced accuracy by target view, with five encoder seeds for trainable models and five evaluation fits for frozen LaBraM. Mixed refs is an exposure baseline trained on all three references (Appendix~\ref{app:post-broad}); its scores are not source-held-out transfer.}}
\label{tab:app-nmt-targets}
\begingroup
\setlength{\tabcolsep}{2.5pt}
\renewcommand{\arraystretch}{1.10}
\begin{tabular*}{\linewidth}{@{\extracolsep{\fill}}lccc@{}}
\toprule
Variant & AVG target & CZREF target & LE target \\
\midrule
AVG-only & 0.666 $\pm$ 0.017 & 0.543 $\pm$ 0.052 & 0.568 $\pm$ 0.049 \\
Mixed refs & 0.682 $\pm$ 0.008 & 0.682 $\pm$ 0.004 & 0.678 $\pm$ 0.009 \\
Contrastive only & 0.652 $\pm$ 0.006 & 0.654 $\pm$ 0.013 & 0.662 $\pm$ 0.010 \\
Supervised contrastive & 0.632 $\pm$ 0.017 & 0.635 $\pm$ 0.020 & 0.638 $\pm$ 0.024 \\
Canonical recon. + invariance & 0.639 $\pm$ 0.041 & 0.636 $\pm$ 0.050 & 0.634 $\pm$ 0.048 \\
LaBraM frozen & 0.688 $\pm$ 0.011 & 0.628 $\pm$ 0.010 & 0.656 $\pm$ 0.014 \\
\bottomrule
\end{tabular*}
\endgroup
\end{table*}

\begin{table*}[htbp]
\centering
\small
\caption{\sloppy NMT source-reference transfer matrix for the key frozen objectives and the frozen LaBraM audit. Entries are balanced accuracy mean $\pm$ standard deviation over seeds. Diagonal entries are in-domain source-target evaluations; off-diagonal entries are shifted transfer evaluations.}
\label{tab:app-nmt-transfer}
\begingroup
\setlength{\tabcolsep}{2.5pt}
\renewcommand{\arraystretch}{1.10}
\begin{tabular}{@{}>{\raggedright\arraybackslash}p{2.1cm}>{\raggedright\arraybackslash}p{2.1cm}>{\raggedright\arraybackslash}p{2.2cm}>{\raggedright\arraybackslash}p{2.2cm}>{\raggedright\arraybackslash}p{2.2cm}>{\raggedright\arraybackslash}p{2.2cm}@{}}
\toprule
Source & Target & Contrastive only & Supervised contrastive & Canonical recon. + invariance & LaBraM frozen \\
\midrule
AVG & AVG & 0.635 $\pm$ 0.022 & 0.631 $\pm$ 0.016 & 0.614 $\pm$ 0.040 & 0.647 $\pm$ 0.042 \\
AVG & CZREF & 0.639 $\pm$ 0.016 & 0.626 $\pm$ 0.022 & 0.531 $\pm$ 0.035 & 0.567 $\pm$ 0.032 \\
AVG & LE & 0.639 $\pm$ 0.018 & 0.631 $\pm$ 0.030 & 0.581 $\pm$ 0.054 & 0.596 $\pm$ 0.042 \\
CZREF & AVG & 0.624 $\pm$ 0.040 & 0.634 $\pm$ 0.020 & 0.585 $\pm$ 0.029 & 0.681 $\pm$ 0.012 \\
CZREF & CZREF & 0.623 $\pm$ 0.033 & 0.634 $\pm$ 0.021 & 0.628 $\pm$ 0.037 & 0.641 $\pm$ 0.023 \\
CZREF & LE & 0.630 $\pm$ 0.039 & 0.634 $\pm$ 0.025 & 0.596 $\pm$ 0.036 & 0.671 $\pm$ 0.022 \\
LE & AVG & 0.614 $\pm$ 0.052 & 0.622 $\pm$ 0.031 & 0.580 $\pm$ 0.046 & 0.663 $\pm$ 0.033 \\
LE & CZREF & 0.618 $\pm$ 0.052 & 0.629 $\pm$ 0.027 & 0.571 $\pm$ 0.035 & 0.612 $\pm$ 0.010 \\
LE & LE & 0.629 $\pm$ 0.052 & 0.631 $\pm$ 0.032 & 0.623 $\pm$ 0.025 & 0.647 $\pm$ 0.015 \\
\bottomrule
\end{tabular}
\endgroup
\end{table*}

\begin{table*}[htbp]
\centering
\small
\caption{\sloppy ERP/P300 source-domain transfer matrix for the key ResNet objectives and the frozen LaBraM audit. Entries are AUROC mean $\pm$ standard deviation over seeds. Diagonal entries are in-domain source-target evaluations; off-diagonal entries are shifted transfer evaluations.}
\label{tab:app-erp-transfer}
\begingroup
\setlength{\tabcolsep}{2.5pt}
\renewcommand{\arraystretch}{1.10}
\begin{tabular}{@{}>{\raggedright\arraybackslash}p{2.1cm}>{\raggedright\arraybackslash}p{2.1cm}>{\raggedright\arraybackslash}p{2.2cm}>{\raggedright\arraybackslash}p{2.2cm}>{\raggedright\arraybackslash}p{2.2cm}>{\raggedright\arraybackslash}p{2.2cm}@{}}
\toprule
Source & Target & Masked recon. & Augmentation only & Invariance first & LaBraM frozen \\
\midrule
ALS P300 & ALS P300 & 0.632 $\pm$ 0.029 & 0.804 $\pm$ 0.016 & 0.764 $\pm$ 0.019 & 0.536 $\pm$ 0.024 \\
ALS P300 & Covert GeoSpell & 0.630 $\pm$ 0.030 & 0.757 $\pm$ 0.020 & 0.739 $\pm$ 0.008 & 0.587 $\pm$ 0.022 \\
ALS P300 & Overt P300 & 0.701 $\pm$ 0.069 & 0.922 $\pm$ 0.006 & 0.901 $\pm$ 0.005 & 0.586 $\pm$ 0.028 \\
Covert GeoSpell & ALS P300 & 0.534 $\pm$ 0.070 & 0.798 $\pm$ 0.016 & 0.760 $\pm$ 0.018 & 0.498 $\pm$ 0.020 \\
Covert GeoSpell & Covert GeoSpell & 0.647 $\pm$ 0.040 & 0.757 $\pm$ 0.019 & 0.740 $\pm$ 0.008 & 0.613 $\pm$ 0.009 \\
Covert GeoSpell & Overt P300 & 0.688 $\pm$ 0.118 & 0.919 $\pm$ 0.006 & 0.900 $\pm$ 0.006 & 0.659 $\pm$ 0.030 \\
Overt P300 & ALS P300 & 0.588 $\pm$ 0.013 & 0.803 $\pm$ 0.017 & 0.763 $\pm$ 0.021 & 0.487 $\pm$ 0.025 \\
Overt P300 & Covert GeoSpell & 0.630 $\pm$ 0.019 & 0.755 $\pm$ 0.020 & 0.739 $\pm$ 0.007 & 0.586 $\pm$ 0.018 \\
Overt P300 & Overt P300 & 0.801 $\pm$ 0.010 & 0.921 $\pm$ 0.006 & 0.900 $\pm$ 0.007 & 0.691 $\pm$ 0.017 \\
\bottomrule
\end{tabular}
\endgroup
\end{table*}

\begin{table*}[htbp]
\centering
\small
\caption{\sloppy PTB-XL source-site transfer matrix for the key ResNet objectives and the frozen ECGFounder audit. Entries are AUROC mean $\pm$ standard deviation over seeds. Diagonal entries are in-domain source-target evaluations; off-diagonal entries are shifted transfer evaluations.}
\label{tab:app-ptbxl-transfer}
\begingroup
\setlength{\tabcolsep}{2.5pt}
\renewcommand{\arraystretch}{1.10}
\begin{tabular}{@{}>{\raggedright\arraybackslash}p{2.1cm}>{\raggedright\arraybackslash}p{2.1cm}>{\raggedright\arraybackslash}p{2.2cm}>{\raggedright\arraybackslash}p{2.2cm}>{\raggedright\arraybackslash}p{2.2cm}>{\raggedright\arraybackslash}p{2.2cm}@{}}
\toprule
Source & Target & Masked recon. & Task only & Augmentation only & ECGFounder frozen \\
\midrule
Site 0 & Site 0 & 0.671 $\pm$ 0.050 & 0.856 $\pm$ 0.011 & 0.856 $\pm$ 0.009 & 0.822 $\pm$ 0.002 \\
Site 0 & Site 1 & 0.705 $\pm$ 0.078 & 0.910 $\pm$ 0.021 & 0.912 $\pm$ 0.015 & 0.906 $\pm$ 0.002 \\
Site 0 & Site 2 & 0.723 $\pm$ 0.053 & 0.907 $\pm$ 0.016 & 0.910 $\pm$ 0.016 & 0.910 $\pm$ 0.002 \\
Site 1 & Site 0 & 0.664 $\pm$ 0.064 & 0.860 $\pm$ 0.006 & 0.857 $\pm$ 0.006 & 0.791 $\pm$ 0.004 \\
Site 1 & Site 1 & 0.730 $\pm$ 0.072 & 0.924 $\pm$ 0.004 & 0.922 $\pm$ 0.004 & 0.899 $\pm$ 0.002 \\
Site 1 & Site 2 & 0.732 $\pm$ 0.075 & 0.914 $\pm$ 0.003 & 0.917 $\pm$ 0.002 & 0.893 $\pm$ 0.003 \\
Site 2 & Site 0 & 0.671 $\pm$ 0.030 & 0.856 $\pm$ 0.010 & 0.855 $\pm$ 0.011 & 0.793 $\pm$ 0.007 \\
Site 2 & Site 1 & 0.740 $\pm$ 0.057 & 0.916 $\pm$ 0.008 & 0.914 $\pm$ 0.010 & 0.864 $\pm$ 0.005 \\
Site 2 & Site 2 & 0.758 $\pm$ 0.040 & 0.915 $\pm$ 0.002 & 0.913 $\pm$ 0.002 & 0.896 $\pm$ 0.005 \\
\bottomrule
\end{tabular}
\endgroup
\end{table*}

\Needspace{4\baselineskip}
\FloatBarrier
\subsection{Compute resources}
\label{app:compute}

\postrev{Experiments were run on two local workstations: one with an NVIDIA GeForce RTX 5090 GPU (32 GB), an AMD Ryzen 9 9900X CPU (12 cores, 24 threads), and 64 GB DDR5 RAM, and an Apple-silicon workstation using the PyTorch 2.7.1 Metal Performance Shaders (MPS) backend. Each $d=128$ NMT encoder in the paired audit trains in under two minutes. Calibrated probes, stress weighting, localization, random controls, and bootstrap intervals run on CPU from saved embeddings, without retraining encoders or task heads. Saved manifests record software versions, source and input hashes, and selection locks. Table and figure regeneration from the numerical artifacts is CPU-only.}

\FloatBarrier
\subsection{Existing assets, licenses, and terms of use}
\label{app:asset_licenses}

Table~\ref{tab:asset_licenses} lists the datasets, pretrained checkpoints, and
external code assets. We credit the original creators through the corresponding
citations in the main paper and use each asset only through its official access path. The PhysioTRACE artifact does not redistribute raw physiological datasets or external pretrained
checkpoints; users are instructed to obtain those assets from the official sources and to comply with
their licenses and terms of use.

\postrev{Code licenses are not assumed to license pretrained weights; each checkpoint is used under the terms published with it and identified by its hash, and no external weights are redistributed.}

License abbreviations used below are Creative Commons Attribution 4.0 International (CC BY 4.0), Creative Commons Attribution-ShareAlike 4.0 International (CC BY-SA 4.0), Creative Commons Attribution-NonCommercial-NoDerivatives 4.0 International (CC BY-NC-ND 4.0), and the Berkeley Software Distribution 3-Clause License (BSD 3-Clause). The MIT License is retained under its standard name.

\begin{table}[htbp]
\centering
\small
\caption{Asset inventory and recorded licenses or terms of use. Raw physiological
datasets and external pretrained checkpoints are not redistributed with the PhysioTRACE artifact.}
\label{tab:asset_licenses}
\begin{tabular}{@{}>{\raggedright\arraybackslash}p{0.18\linewidth}>{\raggedright\arraybackslash}p{0.23\linewidth}>{\raggedright\arraybackslash}p{0.20\linewidth}>{\raggedright\arraybackslash}p{0.29\linewidth}@{}}
\toprule
Asset & Source / citation & License or terms & Use in this paper \\
\midrule
NMT Scalp EEG Dataset & Khan et al.~\citep{Khan2022NMT}; official NMT dataset/code page & Dataset: CC BY-SA 4.0; associated code repository: BSD 3-Clause & Used for the controlled EEG reference-shift evaluation across AVG, CZREF, and linked-ear reference views. \\
PTB-XL ECG Dataset & Wagner et al.~\citep{Wagner2020}; PhysioNet version 1.0.1 & CC BY 4.0 & Used for the ECG cross-site stress test with patient-level official folds and site-defined provenance groups. \\
Brain/Neural Computer Interaction (BNCI) 2014-008 ALS P300 Dataset & Riccio et al.~\citep{Riccio2013ALS}; BNCI Horizon 2020 dataset 008-2014 & CC BY-NC-ND 4.0 & Used as the ALS P300 domain in the ERP/P300 real-domain transfer evaluation. \\
BNCI 2014-009 Covert and Overt ERP-based brain-computer interface (BCI) Dataset & Aricò et al.~\citep{Arico2014P300Jitter}; Aloise et al.~\citep{Aloise2012Geospell}; BNCI Horizon 2020 dataset 009-2014 & CC BY-NC-ND 4.0 & Used for the covert GeoSpell and overt P300 domains in the ERP/P300 transfer evaluation. \\
LaBraM checkpoint and code & Jiang et al.~\citep{Jiang2024}; official LaBraM repository & MIT License & Used only as a frozen EEG encoder for the NMT, ERP/P300, and HMC evaluations; no LaBraM checkpoint is redistributed in our artifact. \\
ECGFounder checkpoint and code & Li et al.~\citep{Li2025ECGFounder}; official ECGFounder repository/model card & MIT License & Used only as a frozen ECG encoder for the PTB-XL audit; no ECGFounder checkpoint is redistributed in our artifact. \\
HMC Sleep Staging Database v1.1 & Alvarez-Estevez and Rijsman~\citep{AlvarezEstevez2022HMC}; PhysioNet & CC BY 4.0 & Used for the held-out controlled reference-shift evaluation of frozen EEG encoders (Appendix~\ref{app:post-hmc}). \\
BIOT and CBraMod checkpoints & Yang et al.~\citep{Yang2023}; Wang et al.~\citep{Wang2025CBraMod}; Braindecode implementations & Code: BSD 3-Clause; checkpoints: terms published with each release & Used only as frozen EEG encoders for the NMT, ERP/P300, and HMC evaluations; no checkpoint is redistributed. \\
MERL checkpoint and adapter & Liu et al.~\citep{Liu2024MERL}; official MERL release & Adapter code: MIT License (upstream notice retained); checkpoint: terms published with the release & Used only as a frozen ECG encoder for the PTB-XL site screen and calibrated LPI; no checkpoint is redistributed. \\
CLEF-Small checkpoint and code & Shu et al.~\citep{Shu2025CLEF}; upstream code repository and authors' Zenodo record & Code: BSD 3-Clause Clear License; checkpoint: terms of the Zenodo record & Used only as a frozen ECG encoder for the observational device audit (Appendix~\ref{app:clef}); no checkpoint is redistributed. \\
\bottomrule
\end{tabular}
\end{table}

\end{document}